%% file: ex_article.tex
\documentclass[
onefignum
,onetabnum
]{siamonline190516}

\input{ex_shared}

\ifpdf
\hypersetup{
  pdftitle={An Example Article},
  pdfauthor={A. Tsourtis, Y. Pantazis, and I. Tsamardinos}
}
\fi

\begin{document}

\maketitle

% REQUIRED
\begin{abstract}
  Inferring the driving equations of a dynamical system from population or time-course data is important in several scientific fields such as biochemistry, epidemiology, financial mathematics and many others.
  Despite the existence of algorithms that learn the dynamics from trajectorial measurements there are few attempts to infer the dynamical system straight from population data.
  In this work, we deduce and then computationally estimate the Fokker-Planck equation which describes the evolution of the population's probability density, based on stochastic differential equations.
  Then, following the USDL approach \cite{USDL_bioinformatics}, we project the Fokker-Planck equation to a proper set of test functions, transforming it into a linear system of equations.
  Finally, we apply sparse inference methods to solve the latter system and thus induce the driving forces of the dynamical system.
  Our approach is illustrated in both synthetic and real data including non-linear, multimodal stochastic differential equations,
  biochemical reaction networks as well as mass cytometry biological measurements.
\end{abstract}

% REQUIRED

% REQUIRED

\section{Introduction}
In many scientific fields ranging from biology 
%(cite mass cytometry 
\cite{Krishnaswamy_DREMI}, single cell RNA sequencing \cite{SCUBA_Marco_PNAS2014},\cite{Bspline_curve_fitting_on_timecourse_data_Liu2006}
%)
, ecology %(cite flocking, schooling 
\cite{Tereshko_reaction_difussion_bees_2000} and epidemiology \cite{Capasso_2017reactiondiffusion_epidemiology}, molecular motion in chemistry, traffic flows in transportation,
%(copy more field applications from others {\blue (finance)}, molecular motion in chemistry, traffic flows in transportation),
cross-sectional data over time are gathered.
Frequently, researchers can only collect and study cross-sectional data because the same subject cannot be measured again at different time points. For instance, biologists study protein signalling using mass cytometry technologies where each single cell is destroyed during measurement. Time-varying phenomena are typically modelled as deterministic or stochastic dynamical systems such as ordinary differential equations (ODEs), partial differential equations (PDEs), stochastic differential equations (SDEs) and variations. The discovery of the governing equations which drive the natural processes is crucial for the in-depth understanding of the complex interaction mechanisms, for forecasting the evolution of the studied phenomena and for making causal predictions on the effects of interventions. Unfortunately, and despite its usefulness, learning the dynamics of individual populations from cross-sectional data remains largely an open and challenging problem.

Discovering the governing equations from trajectorial data where the same subject is repeatedly measured at a series of time instances has been extensively studied during the last decade. Typically, a dynamics learning algorithm selects from a rich {\it dictionary} of non-linear functions (otherwise known as atoms or features or driving forces) which are candidates comprising the unknown system of equations. Given that sparsity is ubiquitous in many physical laws, sparse inference algorithms have been very successful in learning the driving dynamics both for ODEs  \cite{SINDy_2016,Hybrid_SINDy,ReactiveSINDy_Hoffman2018,SINDY_Bifurcation_plasmas_2017,Schaeffer_Integral_terms} as well as for PDEs  \cite{ScienceBrunton17,Schaeffer_2017,Schaeffer_PNAS13}, and SDEs \cite{Clementi_JCP18,PLOS_reactionet_Lasso}. Sparsity is critical for learning large systems from limited data and constitutes a form of complexity penalization and regularization \cite{Tibshirani96, Donoho_2006, Elad_book_sparse_representations}. Sparse optimization regression techniques aim to find the minimal subset of the dictionary that describes the data sufficiently well. Other approaches for dynamics learning based on Monte Carlo Bayesian sampling,  \cite{Friston_Dynamic_causal_modelling}and %Boltt 
%{\blue old work 
\cite{Bollt_Causal_network_inference_SIAM_2015}  also exist. %{\blue Brief mention on what they do?}
However, those approaches with the exception of the general framework in \cite{USDL_bioinformatics} are not transferable to cross-sectional data due to the absence of trajectories hence their inability to calculate the derivatives. Indeed, mean-field type approximations where the average trajectory is considered are not satisfactory due to the multi-modality in many cross-sectional datasets.

There exist recent works that approach the subject of system identification based on population data. At the equilibrium or steady-state regime the evolving pdf is not changing, thus data come from a single snapshot of the system. 
Weinreb et al. \cite{Weinreb_PopulationBalanceAnalysis_PNAS_17} used spectral graph theory with the Fokker-Planck formalism \cite{Jordan_FokkerPlanck_SIAM98} in order to predict cell state temporal ordering in single-cell data. Their algorithm output is a unique gene regulatory network, based on multiple assumptions and accompanied by a discussion on limits on ab-initio population data dynamics inference from an equilibrium distribution. 
Hashimoto et al. \cite{Hashimoto_population_data_RNN_2016} rigorously discuss recoverability assumptions of SDE inference both on the equilibrium and transient regime. Their approach is by the use of neural networks on cell differentiation cross-sectional data at few time points, under the assumption that there exists a potential related to the deterministic term in the unknown SDE. 
Again for dynamical cell differentiation inference, Marco et al. \cite{SCUBA_Marco_PNAS2014} use bifurcation analysis (for binary branching of the data) on evolving pdf's in order to determine the coefficients of a pre-determined potential form (and thus the drift term of the SDE's) along with network, provided that data have sufficient time resolution and low dimension. 
Krishnaswamy et al. \cite{Krishnaswamy_DREMI} carried out cellular network inference from mass cytometry dynamical population data, based on conditional density resampling statistical techniques.
During the final preparation of this paper, we came across the most relevant work to ours, which is under the light of operator approximation instead of differential equations inference. Taylor-King et al. \cite{Dynamic_Distribution_Decomposition_Claassen_2019} use as well, a weak formulation of the spatially evolving multimodal distributions. On the contrary, we extend the weak formulation to the temporal domain and our adopted Fokker-Planck formulation for SDE's is the analog to their Perron-Frobenius operator. Our algorithm infers the dynamics whereas theirs tracks the evolution of dynamically important states (autonomous states i.e. like a Markov chain). From all the aforementioned works on this field, Hashimoto et al. explicitly infer the unknown functional form of the drift term in the non-equilibrium regime.
In \cite{Hashimoto_population_data_RNN_2016} and Ma et al. \cite{gans_weakform_aggregatedata_Zhou_arxiv_2020} a neural network is used to describe the drift term of the FP equation, whereas in \cite{gans_weakform_aggregatedata_Zhou_arxiv_2020} a weak formulation is used in order to connect the F.P. equation with the neural network, while the diffusion term is assumed constant.

We propose a novel population dynamics learning (PDL) algorithm that infers the parameters of the equation that describes the evolution of the data over time. Under the assumption that the cross-sectional data have been generated by a system of stochastic differential equations, the evolution of the data probability density function (pdf) is described by the Fokker-Planck (FP) or forward Kolmogorov equation \cite{Gardiner_Book}. Inferring the coefficients of the FP equation from population data is essentially an inverse problem that requires solving the forward problem (i.e., compute the solution of the FP equation). However, this is not feasible due to the high-dimensionality of the FP equation which scales proportionally to the number of variables or species in biochemistry. Instead, we follow the Unified Sparse Dynamics Learning (USDL) framework \cite{USDL_bioinformatics} and considerably extend the theory behind the USDL algorithm. The main technical novelty of this work is the {\it weak space projection} of FP equation both in time and space, resulting in a transformed yet equivalent atemporal system of equations. This weak space is spanned by appropriately chosen spatio-temporal test functions that best capture the dynamical information of the population data. The integrals involved in the weak space projection are high-dimensional, however, we can statistically estimate them using Monte Carlo approximation. Indeed, an important property of the PDL algorithm is the use of sample averages for the computation of the high-dimensional integrals making the proposed algorithm highly efficient. The last step of PDL algorithm is solving the atemporal system of linear equations using sparse inference algorithms such as Orthogonal Matching Pursuit (OMP) \cite{Tropp_sparseOMP_2007} or Lasso \cite{Tibshirani96}.

We demonstrate the effectiveness of PDL algorithm with two synthetic and one real dataset examples. The first example is a 2D non-linear SDE system with quadruple-well potential where we show that PDL is capable of correctly identifying the underlying driving forces, deterministic and random terms. This example demonstrates that it successfully recovers a multimodal distribution from population data. The second example is a four variable biochemical reaction network that emulates a signaling pathway cascade of proteins. We show that PDL algorithm learns the biochemical reactions from population data generated from a single experiment (data set), correctly. Additionally, we explore the hyper-parameter space demonstrating that there is a relatively large regime for the hyper-parameters where results are stable. The final example is a set of mass cytometry data \cite{Krishnaswamy_DREMI} that contains protein signalling interactions. PDL algorithm was able to forecast the evolution of the protein populations.

 The manuscript is organized as follows. First, the SDE model over the dictionary and the deduced Fokker-Planck equation are presented along with assumptions. 
In \cref{sec:pdl:alg} the weak formulation of the FP equation is derived, with key points over the analytic calculations and numerical setup. The resulting system of integral equations is atemporal and linear with respect to the dictionary items. In \cref{sec:Linear} we show how this reformulation of the problem belongs to the sparse signal recovery field, where a wide range of theoretical tools and algorithms can be utilized to solve it and thus perform (dictionary atom) feature selection. At this point we present our PDL algorithm, where we chose OMP to obtain a sparse solution.
Finally in \cref{sec:experiments}, we demonstrate the efficacy of the PDL algorithm on synthetic and real temporal population data, followed by a discussion on advantages and limitations of our primal approach.

%\medskip
%\begin{figure}[H]
%    \centering
%    \includegraphics[width=0.75\paperwidth]{./figs/pdf_evolution_samples.eps}
%    \caption{Motivation example of two variable time-course (or population) data at different sampling times depicting a joint ``density evolution" over time. Note that at $t=1$, both variables' samples are concentrated at $\{0,0\}$, later divided among four regions that is two bimodal marginal distributions.
%    Naive averaging over the propagating distribution of each variable separately, would fail to provide the correct input data for existing algorithms based on trajectorial data.}
%    \label{fig:timecourse_data}
%\end{figure}
%========================================================
%========================================================
% \section{Model notation and Definitions}
\section{Preliminaries}
\label{sec:model}

\subsection{Population data}
The $p$-th sample of a time-course dataset (i.e., a non-repeated measurement) at time-point $t_k\in\{t_1, \dots,t_K\}$ is denoted as
\begin{equation}
\mathbf{x}_{k,p} := \mathbf{x}_{k,p}^0 + \int_{t_1}^{t_k} d {X}_t\,,\,\,\, \mathbf{x}_{k,p}\in\mathbb R^N
\end{equation}
where ${X}_t$ is the underlying $N$-dimensional stochastic process while $\mathbf{x}_{k,p}^0\in\mathbb R^N$ are the initial i.i.d. data (i.e., ${X}_{t_1}=\mathbf{x}_{k,p}^0$). The i.i.d. assumption for the initial data renders the fact that each sample at each time point corresponds to a separate object. For each time-point, $P$ samples that constitute a statistical approximation the probability distribution of the process are given. For the sake of simplicity, we do not consider extra measurement noise and the stochasticity of the data stems solely from the intrinsic randomness of the process.

% In real systems, no closed form expression for the drift term is known and the $N$-dimensional process $X_t$ can only be observed through simulation or measurement data $X(t_k)\in \mathbb{R}^N$ at discrete time-points $\{t_1,\dots, t_k, \dots,t_K \}$.
% Multiple observations (samples) per each $t_k$ constitute population data, that can form one or more clusters (clouds of samples).
% Temporal population data are given as input and we aim to reverse-engineer in order to construct the underlying system that is able to reproduce it.
% %==========================================================
% \cref{appendix:types_of_data} contains detailed information on the form of measurement data considered in this manuscript.

\subsection{SDE Modeling}

We assume that the stochastic process $X_t \in \mathbb{R}^N$ is described by a system of stochastic differential equations. An SDE constitutes of two parts: the deterministic part or drift and the stochastic part or the diffusion. We further assume that the drift can be written as a linear combination of potentially non-linear functions as well as the diffusion term has uncorrelated components. Under those assumptions, the mathematical formulation of the SDE is given by
\begin{equation}
\label{eq:SDE_general_reformulated}
dX_t = A \psi(X_t)dt + \Sigma dB_t
% &A\in \mathbb{R}^{N\times Q},    \quad\psi : \mathbb{R}^{N}  \rightarrow \mathbb{R}^{Q}  \nonumber 
\end{equation}
where $A\in \mathbb{R}^{N\times Q}$ 
%[{\red Are dimensions ok?}] 
is the unknown and typically sparse connectivity (or coefficient or parameter) matrix to be estimated while \emph{dictionary} $\psi(\cdot)$ is a (given) $Q$-dimensional vector-valued vector function $\psi: \mathbb{R}^{N}  \rightarrow \mathbb{R}^{Q}$ which contains all the pre-determined candidate functions that might drive the dynamics. Candidate functions are usually powers, cross-products, fractions, trigonometric, exponential or logarithmic functions of the state variables leading to non-linear dynamical systems. $B_t \in \mathbb{R}^{N}$ is an $N
$-dimensional standard Brownian motion while the diffusion matrix $\Sigma=\diag(\sigma_{1},\dots, \sigma_{N})$ is diagonal. See Appendix \ref{appendix:SDE} for detailed explanation.

% where the initial configuration of $X_t$ is $X_0\sim p_0(X_t):\mathbb{R}^N \rightarrow \mathbb{R}$ and Brownian motions $B_t$'s are iid.  
% The first term on the r.h.s. of \Cref{eq:SDE_general_reformulated} is the deterministic term or drift whereas the second term is the diffusion (random forcing). As explained in \cref{appendix:SDE}, we assume that the drift term is
% written as a linear combination of $Q$-element basis $\{\psi_q(x)\}_{q=1}^Q$ called {\it dictionary}, where the basis coefficients constituting matrix $A$ need to be determined.
% The analysis of the drift term by $Q$ dictionary atoms, in essence, is a {\it construction of new features} which 
% constitute the building blocks of the solution. $B_t \in \mathbb{R}^{N} \sim \mathcal{N}(0,I)$ and we assume uncorrelated constant diffusion coefficients $\Sigma= \text{diag}(\sigma_{1},\dots, \sigma_{N})$.

% % are $N$ continuous random variables (in other context termed dimensions, species, state variables etc) which are stochastic processes and can be described by stochastic differential equations (SDE's) as they evolve in time.
% In our setting, we make some basic assumptions on the general SDE form, which are described in detail, in \cref{appendix:SDE}. 

%========================================================
%========================================================
\subsection{Fokker-Planck Equation}
The Fokker-Planck (FP) equation describes the probability density {\it evolution} of a stochastic process that follows an SDE. 
%The SDE of the 1-dimensional $x_t$ is given by:
%
%\begin{align}
%    dx_t = &a(x_t, t)dt + b(x_t,t)dBt \\
%    x_t \in \mathbb{R}, \quad &a(x_t,t):\mathbb{R}\times \mathbb{R}^{+} %\rightarrow \mathbb{R} \quad (\text{DRIFT}) \nonumber \\
%     &b(x_t,t):\mathbb{R}\times \mathbb{R}^{+} \rightarrow \mathbb{R} \quad %(\text{DIFFUSION}) \nonumber \\
%     &B_t \in \mathbb{R} \text{is 1-dim std Brownian motion} \nonumber
%\end{align}
%\medskip
%deterministic dynamics $\longrightarrow$ drift term\\
%stochastic dynamics $\longrightarrow$ diffusion term
%
%\medskip
%Subject to initial condition: $x_0 \sim P_0(x):\mathbb{R}\rightarrow %\mathbb{R}$  which is the initial distribution of the first measurement.
Let $p(\mathbf{x},t)$ be the probability of observing the value ${\bf x}$ at time $t$ defined by
\begin{equation}
   % p(x,t)=\mathbb{P}(x_t=x|x_0 \sim p_0)
       p(\mathbf{x},t) = \mathbb{P}(X_t=\mathbf{x}|\mathbf{x_0}\sim p_0)
\end{equation}
%Then the F.-P. equation is a 2 dimensional PDE:
%\begin{equation}
%\boxed{ \partial_t p(x,t) = -\partial_x \{a(x,t)p(x,t)\} + %\frac{1}{2}\partial_{xx}\{b(x,t)b(x,t)p(x,t)\}}
%\end{equation}
%
%\underline{Generalization to $N$-dimensions}
%\begin{align}
%        dX_t = &a(X_t, t)dt + b(X_t,t)dBt \\
%  X_t \in \mathbb{R}^N, \quad &a(X_t,t):\mathbb{R}^N\times \mathbb{R}^{+} %\rightarrow \mathbb{R}^N \quad (a_1, a_2,\dots,a_N)  \nonumber \\
%     &b(X_t,t):\mathbb{R}^N\times \mathbb{R}^{+} \rightarrow \mathbb{R}^{N'} % \nonumber \\
%     &B_t \in \mathbb{R}^{N'} \text{is N'-dimensional  Brownian motion} %\nonumber
%\end{align}
%and the initial distribution is:
Then the FP equation for \Cref{eq:SDE_general_reformulated} is an $(N+1)$-dimensional parabolic PDE given by
\begin{align}\label{eq:FP_special}
 %\boxed{
 \partial_t p(\mathbf{x},t) = -\sum_{n=1}^N {\bf a_n}^T  \partial_{x_n} \{\psi(\mathbf{x})p(\mathbf{x},t)\} + \frac{1}{2}\sum_{n=1}^{N}\sigma^2_{n}\frac{\partial^2}{\partial_{x_{n}x_{n}}} p(\mathbf{x},t) %} 
\end{align}
where  $\mathbf{a_n}\in\mathbb{R}^Q$ is the $n$-th row of the connectivity matrix $A$.
%The general form of \cref{eq:FP_special} can be found in \cref{appendix:FP_general}.
Our goal is to learn the dynamical equations by solving \cref{eq:FP_special}, or in other words to 
determine the non-zero elements of matrices $A$ and $\Sigma$. 

% Note that in the case of
% $a(X_t,t)=X_t$, $b(X_t,t)=\text{const.}$ the FP is called Ornstein Uhlenbeck (OU) process where the analytic solution is an exponential.
% We remark that for continuous-time, continuous-space Markov Chains (as in time-course population data) the FP equation becomes Master Equation and the goal is to infer the transition rate matrix (here $A$). Deterministic equation systems ($\Sigma$ = {\bf 0}), continuous-time/discrete 
% space systems, discrete-time/continuous-space systems and the above are all special cases of the Kolmogorov equation.

%\underline{Assumption 4}: The diffusion coefficient
%\begin{equation}
%b(X_t,t)=\text{constant over time}=\begin{bmatrix}
%                  b_1 \\  b_2 \\ \vdots \\ b_N \end{bmatrix}  
%\end{equation}
%In practice, this means that unmeasured variables do not affect the system in %a different manner throughout the time measurements.
%
%\underline{Assumption 5}: $B_t\in\mathbb{R}^N$ is N-dimensional Brownian motion.
%

\section{Population Dynamics Learning Algorithm}
\label{sec:pdl:alg}

\subsection{Weak formulation of the FP equation}

We proceed with the weak space formulation, which constitutes the transformation of the dynamical system inference to an equivalent atemporal one. 
This is achieved through the utilization of appropriate test-functions and intuitively, the procedure can be understood as a projection operator that multiplies the system's equation by these test-functions and integrate over time and space.
Under this setup, we avoid using a numerical differentiation scheme (such as finite differences) for the crucial and sensitive time derivative computation on the 
left-hand side of \Cref{eq:FP_special}, like 
%the SINDy-like approaches do
\cite{PLOS_reactionet_Lasso,Hybrid_SINDy}
%[{\red reactionet uses ``empirical moment gradients" which is sliding window finite diffs...}]
.
Subsequently we avoid discretization errors due to long (and possibly uneven) data sampling time intervals, dependent on the sampling time resolution. Additionally, the FP \Cref{eq:FP_special} as a PDE is high-dimensional in principle, ($N+1$ dimensions in total) so even a small variable size $N$ renders the
integrals (when solving \Cref{eq:FP_special} for $p(\cdot,t)$) computationally intractable in practice for realistic systems. Thus weak formulation makes it computationally tracktable.

The weak equation partially reverses the derivation procedure to return an integral formulation, which is less strict than the PDE.
Essentially it is a projection of the system to a weak space spanned by test 
functions $\phi_m({\bf x},t):\mathbb{R}^N \times \mathbb{R}^{+} \mapsto \mathbb{R}, m\in\{1,2,\dots, M\}$, which will be 
subsequently differentiated \cite{Evans_PDES}.

The weak space FP formulation for \Cref{eq:FP_special} for $\Sigma=diag(\sigma_1, \dots, \sigma_N)$ is given by:
\begin{equation}\label{eq:weak_form_special}
\begin{split}
    \int_{0}^{T} \int_{\mathcal{D}} \phi_m({\bf x},t) &\partial_{t}p({\bf x},t)d{\bf x}dt = -\sum_{n=1}^{N} \int_{0}^{T} \int_{\mathcal{D}} \phi_m({\bf x},t)\partial_{x_n} \{a_n^T \psi({\bf x},t) p({\bf x},t) \} d{\bf x}dt \\
     &+ \frac{1}{2} \sum_{n=1}^{N} \int_{0}^{T} \int_{\mathcal{D}} \phi_m({\bf x},t)\partial_{x_{n}x_{n}} \{\sigma^{2}_{n}p({\bf x},t) \}d{\bf x}dt, \quad m=1,\dots,M 
\end{split}
\end{equation}
on the spatial domain $\mathcal{D} \subset \mathbb{R}^N$ and time domain $T\in \mathbb{R}$ for the $m$-th test function.
Thus we conclude with a set of $M$ integral equations. We choose test functions in a way that their form is 
able to capture the data heterogeneity, form and time-scales as we discuss later on.
As one can see, the integrals are $N$-dimensional as the variables $\{X^n\}_{n=1}^N$ are coupled in general.
In the following, we define these spatio-temporal functions as:
\begin{equation}\label{eq:test_funcs_main}
 \phi_m({\bf x},t) := \bar{\phi}_{m_1}({\bf x}) \tilde{\phi}_{m_2}(t)
\end{equation}
meaning that space $\bar{\phi}_{m_1}$ and time $\tilde{\phi}_{m_2}$ have different functional forms in general and $m_1\in\{1,\dots, M_1\}$, $m_2\in\{1,\dots, M_2\}$, $M_1 M_2=M$.
We proceed with analytic calculations in \cref{appendix:weak_from}.

At this point, we stress the fact that the weak space formulation {\bf does not affect the
constants} $a_{nq}$ that constitute the unknown matrix $A$ of the inference problem \Cref{eq:SDE_general_reformulated} if one chose to proceed differently, solving the minimization problem directly as in \cite{SINDy_2016}. In addition, the reader should not confuse the constants in \cref{eq:sup_assump} with the unknowns $a_{nq}$. 

In \cref{sec:Linear} we proceed with a demonstration of how these equations
form $N$
systems of linear equations of order $M$
each.  This means that we get $Q$ terms (one row) of the unknown matrix $A\in \mathbb{R}^{N\times Q}$ when solving for each dimension $n$.

%-------------------------------
\subsubsection{Test functions of the Weak formulation}
The role of test functions is vital for the inference problem since it is imperative that they accurately capture the information from the time domain to
the weak space spanned by the chosen test functions. 
Examples of test functions families are Fourier modes, B-splines, Legendre  polynomials, Hermitian polynomials etc. They are required to be smooth, bounded, easy to compute and not necessarily orthogonal. 
The optimal choice and number of test functions depend on the specific problem at hand and there is no family that can perform optimally for every problem.

\begin{figure}[h]
\centering
    \includegraphics[height=22em]{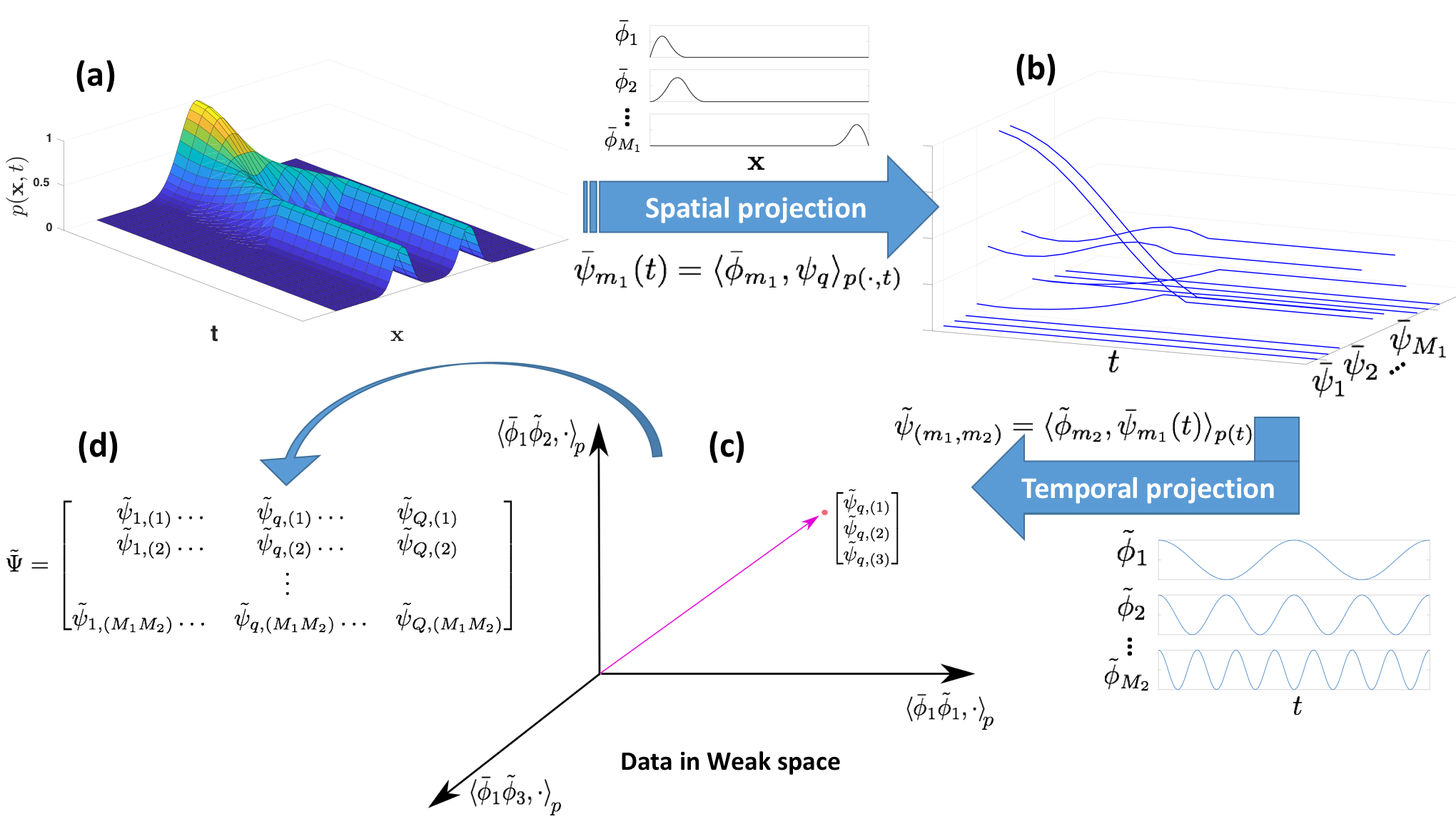}
\caption{Visual depiction of weak space projection over spatial and temporal test functions.
(a) Known one-dimensional evolving bimodal pdf $p(x,t)$ to be learned. We show the continuous-time pdf for better visualization, though PDL algorithm considers time-course measurements at discrete time points.
(b) Projection of $q$-th dictionary atom, based on data from $p(x,t)$, on $M_1=11$ spatial B-spline test functions throughout the temporal range $[0,T]$ (continuous-time visualization, $q$-th atom is chosen linear). %linear dictionary here for illustration purprose.
The initial distribution $p_0(x)$ gradually splitting, is mostly captured by test-functions $\bar{\phi}_6(x), \bar{\phi}_7(x)$, whereas the two equilibrium modes of $x$ are primarily captured by $\bar{\phi}_4(x), \bar{\phi}_5(x), \bar{\phi}_8(x), \bar{\phi}_9(x)$, because of their support. 
(c) Second projection of the data, formerly projected in space, over 3 temporal Fourier test functions of varying frequency ($M_2$ in total). Here only 3 are shown due to the high dimensionality of the $M_1 M_2$ combinations and one (out of $Q$) dictionary atoms. 
Eventually we get $M_1 M_2 = M$ such spatio-temporal projections $\langle \bar{\phi}_{m_1}\tilde{\phi}_{m_2}, \cdot \rangle$ as shown in \Cref{eq:weak_form_innerprod}.
(d) Resulting matrix of a system of $M$ linear equations in the weak space, for $Q$ dictionary atoms, as described in \Cref{eq:SLE_wrt_M} $(n=1)$. 
%Unknown vector $a\in \mathbb{R}^Q$ of $Q$ unknowns corresponding to the dictionary atoms, is sparse upon solving the SLE. 
Vector $Z$ is obtained by similar projection involving the temporal derivatives $\partial_t \tilde{\phi}(t)$. }
	\label{fig:DW_projection}
\end{figure}

B-splines of order $k$ are piece-wise polynomial functions of degree $k-1$. Their derivatives are lower order polynomials which renders them an attractive choice in the differentiation of the weak form in \Cref{eq:ibp_2}. In comparison to the Hermite and Legendre polynomials formerly used in ODE's \cite{USDL_bioinformatics}, the B-splines don't 
drive off to increasingly larger absolute values as their order increases, so
the computations contain significantly less numerical errors.
B-splines were used as spatial test functions ($\bar{\phi}_{m_1}(x)$) whereas for the time domain we used Fourier modes ($\tilde{\phi}_{m_2}(t)$). The latter constitute varying sinusoidal functions that capture spatial data density changes over time.
\cref{fig:DW_projection} shows the weak space projection of data $x(t)\sim p(x,t)$, first on spatial test functions $\bar{\psi}_{m_i}(t)$ \cref{fig:DW_projection}(b) and subsequently on temporal test functions $\hat{\psi}_{m_i,m_j}$ \cref{fig:DW_projection}(c).
Thus, population data are projected over all the combinations of the spatial and temporal test functions.
The resulting system of linear equations (d) contains the same unknown constants, as discussed in \Cref{sec:Linear}.

%-------------------------------
\subsubsection{Integral Computation with sample estimate}

In the weak formulation of PDE's every integral is an inner product (here in  $L^2$), as well as in \Cref{eq:weak_form_special}. %\Cref{eq:ibp_2}.
In our problem setup, integrals can be seen
as expectations $\mathbb{E}_{p}(\cdot)$ since the factor $p(x,t)$ is present in every one of them. Thus naive Monte Carlo
 multi-dimensional integration is feasible \cite{MC_integration}. Actually, sampling is the only option in high-dimensional integrals. Moreover, it eliminates the need
to approximate the probability density $p(x,t)$ by numerical integration. These attractive properties are very
important and allow for a fast and easy implementation. Details on the integral estimators are in \Cref{appendix:estimators}.
The variance of the estimator in \Cref{eq:naive_MC_integration} and consequently in \cref{eq:naive_MC_integration2} is $\mathcal{O}(\frac{1}{\sqrt{P}})$. Note that the weak formulation still contains derivative estimation of the test functions in 
\Cref{eq:ibp_2} though we are absolved of the finite differences derivative estimation error $\mathcal{O}(\delta t_{k})$ and variance $\mathcal{O}(p(x,t_{k+1})-p(x,t_{k}))$ associated with $\frac{\partial p(x,t)}{\partial_t}$. %[{\red Check}]. 
In addition, a typical finite differences approach would dictate a mesh discretization in $N$
dimensions for every $t$, which is numerically intractable even for small values of $N$.

%---------------
\begin{figure}[h]
\centering
    \includegraphics[height=18em]{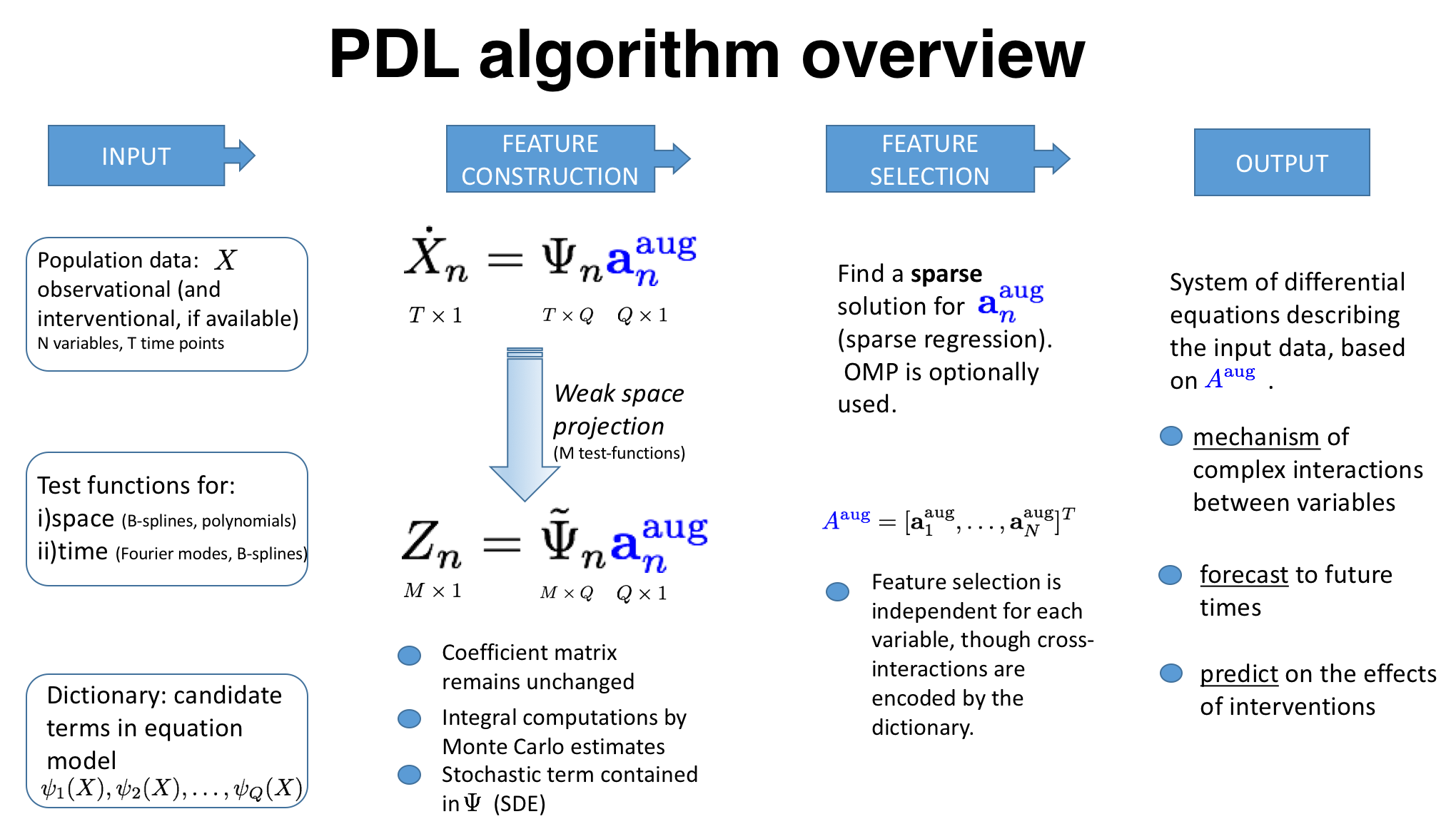}
\caption{ Overview of PDL algorithm for learning SDE's. Note that the weak space projection does not affect the coefficient matrix $A$, on which a sparse solution is required. The problem is projected to an atemporal equivalent, of dimensionality $M$. }
	\label{fig:PDL}
\end{figure}

%---------------------------------------------------------------
\subsection{Solving the Linear System}
\label{sec:Linear}
Upon computation of the integrals over time and space, \Cref{eq:weak_form_special} (specifically \Cref{eq:ibp_2}) is written as a system of linear equations:
\begin{equation}\label{eq:SLE_form}
\begin{split}
	Z_{m_1,m_2,n} = \sum^{Q}_{q=1} a_{nq} \tilde{\Psi}_{m_1,m_2,n,q} + \frac{\sigma^2_n}{2}W_{m_1,m_2,n}, \quad n=1,\dots, N
\end{split}	
\end{equation}
where the unknowns are a vector of coefficients ${\mathbf a_{n}}\in \mathbb{R}^Q$ plus diffusion coefficient $\sigma_n \in \mathbb{R}$. 
$Z_{m_1,m_2,n}$ contains the projected time derivatives of the density function $p(X,t)$ on the $m_1$-th spatial and $m_2$-th temporal test functions.
Respectively, $\tilde{\Psi}_{m_1,m_2,n}$ constitutes the data in dictionary space, $W_{m_1,m_2,n}$ the diffusion component, both subsequently projected onto weak space.
One can think of $\tilde{\Psi}$ as feature ``transformation" of constructed feature matrix $\Psi$.
For the $n$-th variable, the linear system is given by:
\begin{align}\label{eq:SLE_wrt_M}
    %Z_n &=  \sum^{Q}_{q=1} a_{nq} \tilde{\Psi}_{n,q} +\frac{\sigma^2_n}{2}W_{n} \nonumber \\
    Z_n &= \tilde{\Psi}_n {\mathbf a_n} + \frac{\sigma^2_n}{2}W_{n}= 
            \tilde{\Psi}^\text{aug}_n {\mathbf a_n}^\text{aug}
\end{align}
where $Z_n \in \mathbb{R}^M$, $\tilde{\Psi}_n \in \mathbb{R}^{M\times Q}$, $\mathbf{a}_n \in \mathbb{R}^Q$, $W_n \in \mathbb{R}^M$
and can be solved in the least-squares sense (see next section). {\bf Note} that due to \cref{eq:test_func_pairwise} there is dependence of $\tilde{\Psi}$ on $n$. Thus we independently solve $N$ linear systems of the form \cref{eq:SLE_wrt_M}.

The usual approach of other works concludes to a linear system of equations such as  \Cref{eq:SLE_form},
without the integrals, and having a separate system for each time point. %{\red Rephrase?}
In effect, the dimensionality of the problem increases with measurements size whereas in our proposed USDL methodology, time is captured in the atemporal \Cref{eq:SLE_form} of predefined size $M$, dependent on the input data smoothness. Hence there is no need for smoothing or subsampling (or block averaging) for computational cost reduction and convergence.

%[rewrite] The resulting system in compact form is:
%\begin{align}\label{eq:SLE_weak}
%    \mathbf{Z} &= \mathbf{\Psi}^{T}_\text{aug} A_\text{aug} \\
%    \mathbf{Z} &= [Z_1 |\dots |Z_N] \\
%    A_\text{aug} &= [A_1|A_2|\dots|A_Q| (\diag{\Sigma})^T] \\
%    \mathbf{\Psi}_\text{aug} &= [\tilde{\Psi}_1| %\tilde{\Psi}_2| \dots |\tilde{\Psi}_Q| W_1| \dots |W_N]
%\end{align}
%where $Z \in \mathbb{R}^{M\times N}$, $A_q \in \mathbb{R}^{N}$, $A_\text{aug} \in \mathbb{R}^{N\times(Q+1)}$, $\tilde{\Psi}_q \in \mathbb{R}^{M\times N}$, $\mathbf{\Psi}_\text{aug} \in \mathbb{R}^{(M\times (Q+1))\times N}$.

%---------------------------------------
\subsubsection{Minimization problem}\label{sec:min_prob}
%[{\red At least half should be moved to appendix.}]
Upon computation of matrices $Z_n$, $\tilde{\Psi}_n$ and $W_n$, the system \cref{eq:SLE_wrt_M} is written as:
\begin{equation}\label{eq:linearSLE}
    Z = \Psi a
\end{equation}
where we have appended the extra unknown $\sigma_n$ in vector ${\mathbf a}_n$ (also appended vector $W_n$ in array $\tilde{\Psi}_n$) and dropped the index $n$ and $\Tilde{}$ for simplicity of notation.

Solving \cref{eq:linearSLE} for $a$ lies in the category of regression problems.
The resulting systems are over-determined because we use a large number of test functions and a broad family of potentially useful dictionary atoms, and a least-squares type minimization of the form $\min ||Z-\Psi a||$ over $a$, does not have a sparse solution but a full (dense) one. 
In practice, a dense solution means that there is a high correlation of the available candidate features (columns of $\Psi$) as a result of noise in the data (ill-conditioned matrix $\Psi$) and sparse regression is preferable. To achieve this, a minimization problem with penalization is solved instead:
\begin{equation}\label{eq:SSR_L0}
    \underset{a}{\min} ||a||_0 \quad \text{subject to} \quad {||Z-\Psi a ||_2 }\leq \epsilon, 
\end{equation}
which uses the $L^0$ norm over the minimization, where $||a||_0$ is the number of non-zero elements in $a$ and $\epsilon$ is the regression error. 

In this work, learning is performed by determining a small subset of important features from an over-determined set of possible features ($Q$ in total), using a non-convex sparse regression greedy algorithm termed Orthogonal matching pursuit (OMP), that approximately solves problem \cref{eq:SSR_L0}. 
Apart from being fast, OMP has the important property that its hyper-parameter is easy to interpret and approximate from the input data. Another convenient attribute of OMP is that of adding prior knowledge, over the set of features describing the unknown driving forces, in a straightforward way (see \Cref{sec:double_well}). This is important in applications where
the user has a partial knowledge on the equation terms or performs interventions in a systematic way (see \cref{sec:intervention}). %[{\red mention anything on the choice of LASSO?}]

Details on the minimization problem can be found in \Cref{appendix:minimization_problem}.
%-------------------------------------------------------
\subsection{PDL Pseudocode}\label{sec:alg}
The PDL algorithm is summarized in \Cref{alg:PDL} as well as in \cref{fig:PDL}. In case that the input consists of one time-series (trajectory), the former version of USDL \cite{USDL_bioinformatics} is applicable.

\begin{algorithm}
\caption{PDL (Population Dynamics Learning)}
\label{alg:PDL}
\hspace*{\algorithmicindent} \textbf{Input:} population data: $P$ observations of $N$ variables over $K$ time points:
$S=\Big\{x_{1}^{p}(t),\dots, x_{N}^{p}(t)|t=t_1,\dots,t_K \Big\}^P_{p=1}$,
dictionary $\psi_{1:Q}(x)$, set of test functions in time and space $\big\{\phi_m\big\}^M_{m=1}$. When data from $(R)$ interventions are present, $S=\{S^{1},\dots, S^{R}\}$  \\
\hspace*{\algorithmicindent} \textbf{Output:} Inferred $\hat{A}$ s.t. $Z = \tilde{\Psi} \hat{A}$\\
\begin{algorithmic}[1]
\State $\Psi$ = {\tt comp\_psi}($S$) \Comment{data on dictionary (basis) $\in \mathbb{R}^{T\times Q\times N}$}
\State // Compute Weak Space projections $Z, \tilde{\Psi}, W$ %for derivatives, drift and noise}
\For{$n=1,\dots,N$}
    \For{r=1,\dots, R}  %\Comment{assume $\tilde{\phi}_{m_1}({\bf x})=\prod_{n}\tilde{\phi}_{m_1}(x_n),\quad {m_1}\in 1:M_1$}
    \For{m=1,\dots, M} %\Comment{$\phi_m(x,t)=\bar{\phi}_{m_1}(x)\tilde{\phi}_{m_2}(t),\quad M_1 M_2=M,\quad m_2\in 1:M_2$}
   % \State $z_{n,m}^{bound,(r)} \gets \mathbb{E}_{p(x^{(r)}_n;t_K)}[\phi_m(x^{(r)}_n(t_K),t_K)] 
 %   -\mathbb{E}_{p(x^{(r)}_n;t_1)}[\phi_m(x^{(r)}_n(t_1),t_1)]$
    
        %\For{$t=t_1,\dots, t_K$}
            %\State $z^{(r)}_{n,m} \gets \int_{D_n} \partial_t\phi_m(x_n,t)p(x_n(t),t)dx_n  $
            \State $z^{(r)}_{n,m} \gets \mathbb{E}_{p(x^{(r)}_n;t)}[\partial_t \phi_m(x^{(r)}_n(t),t)]$ 
            \Comment{sample estimate for integral using (\cref{eq:naive_MC_integration2})}
            \For{q=1,\dots, Q}
                \State $\tilde{\Psi}^{q,(r)}_{n,m} \gets \mathbb{E}_{p(x^{(r)}_n;t)}[\partial_{x_n} \phi_m(x^{(r)}_n(t),t) \psi_q(\mathbf{x}^{(r)})] $
                \Comment{$\psi_q$: $q$-th column of $\Psi(,,n)$}
            \EndFor
            \State $w^{(r)}_{n,m} \gets \mathbb{E}_{p(x^{(r)}_n;t)}[\partial_{x_nx_n} \phi_m(x^{(r)}_n(t),t) ] $
       %\EndFor

        \EndFor
    %\State $z^{(r)}_n \gets [z^{(r)}_{n,1:M}]^T$
    %\State $w^{(r)}_n \gets [w^{(r)}_{n,1:M}]^T$
    %\State $\tilde{\Psi}^{(r)}_n \gets [\Psi^{(r)}_{n,1:M,1:Q}]$
    \EndFor
\State $Z_{n} \gets [z^{(1)}_{n,1:M}, \dots, z^{(R)}_{n,1:M}]$ \Comment{$\in M R$}
\State $W_{n} \gets [w^{(1)}_{n,1:M}, \dots, w^{(R)}_{n,1:M}]$
\State $\tilde{\Psi}_{n} \gets [\tilde{\Psi}^{(1)}_{n,1:M}, \dots, \tilde{\Psi}^{(R)}_{n,1:M}]$
\Comment{$\in M \times Q$}
\State $\mathbf{a_n}$ = {\tt OMP}($Z_n, \tilde{\Psi}_n$,$W_{n}$) \Comment{Solve sparse regression using OMP}
\EndFor

\State Estimate relative error\\ %L2 distance or  ERC, MIP 
%\State $\hat{A} = [a_1, \dots, a_N]^T$
\Return $\hat{A}= [\mathbf{a_1}, \dots, \mathbf{a_N}]^T$
\Comment{$\in N \times Q$}

%\Return $\hat{A}$%, metrics MIP and ERC

\end{algorithmic}
\end{algorithm}
%============================================

\section{Experimental results}
\label{sec:experiments}
In this section we demonstrate the performance of our proposed PDL algorithm \footnote{The code is available online at: https://github.com/mensxmachina/...
} on three different example paradigms; 
from synthetic multi-dimensional problems with known solutions (\cref{sec:double_well,sec:protein_net}) to a real data problem (\cref{sec:mass_cyto}). We seek a parsimonious dynamical model with as few (non-linear) terms as possible, representing the available data.

\subsection{Learning the dynamics of a multimodal (joint) distribution}\label{sec:double_well}
Our first system under study is based on multimodal SDE's of two variables of different time scales and intrinsic stochastic noise. 
The drift term of each variable consists of a double-well potential (non-linear, fourth order)
and both terms are of the same functional form with different coefficients resulting to low and high equilibration times of the corresponding stochastic processes $x_1,x_2$. 
We chose this well-studied paradigm because it is used to
model a wide range of physical, financial and other phenomena, having the intrinsic difficulty of sample concentration on two modes over time. 
This data peculiarity, renders existing approaches of: a) averaging-out samples per time point towards a single trajectory or by
b) averaging multiple trajectories into fewer in order to reduce computational cost and convergence issues, not applicable.
The two-variable system of SDE's is given by:
\begin{subequations}\label{eq:SDEs_N2}
\begin{align}
	dx_{1} &=  - (x_{1}^3 - x_{1})dt \quad + \sigma_1 dW_1 \\
	dx_{2} &=  - (x_{2}^3 - 0.5^2 x_{2})dt + \sigma_2 dW_2
\end{align}
\end{subequations}
\begin{figure}[h]
\centering
    \includegraphics[width=0.71\paperwidth]{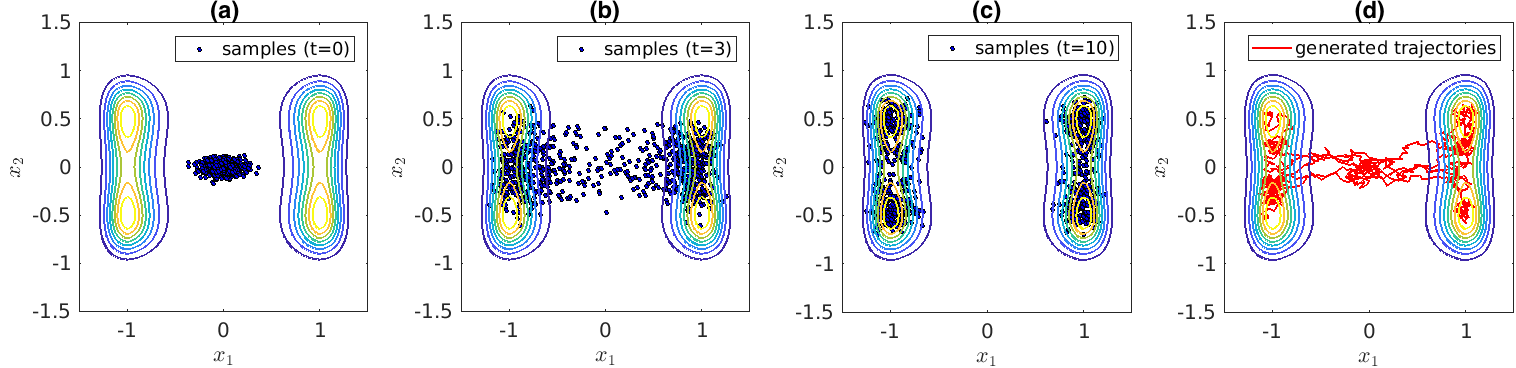}
\vspace{-3mm}
	\caption{$\mathbf{(a)-(c)}$ Time-course (population) data evolution, of bimodal stochastic processes $x_{1}, x_{2}$ of \Cref{eq:SDEs_N2}, measured at $\mathbf{(a)}$ initial,  $\mathbf{(b)}$ transient and $\mathbf{(c)}$ steady-state times (blue dots). Contours in each plot correspond to the steady-state pdf. 
	%$\mathbf{(a)}$ Contour plot of the level sets of the steady state joint pdf which is multimodal. 
	The quadruple-well potential of this system restricts samples in four minima centered at: $\{-1,-0.5\}, \{-1,0.5\}, \{1,-0.5\}, \{1,0.5\}$, so simple averaging of the data along each variable would result to samples around $\{0,0\}$ and inference is bound to fail.
	%$\mathbf{(b)},\mathbf{(c)}$ Separately shown, the density evolution of bimodal stochastic processes $x_{1}, x_{2}$ of \Cref{eq:SDEs_N2} over time (blue dots).
	%$p(x_{1},t)$ starting from $P(x_1,t=0)=p_{0}(x_1)\sim \mathcal{N}(0,\sigma^{2}_{0}=0.1^2)$ 
	%and $P(x_2,t=0)=p_{0}(x_2)\sim \mathcal{N}(0,\sigma^{2}_{0}=0.1^2)$.
	The noise coefficients are $\{\sigma_1=0.2, \sigma_2=0.1\}$, resulting to larger variance of samples for process $x_1$.
	$\mathbf{(d)}$ Upon inference of the system \cref{eq:SDEs_N2} based on time-course data, we generate stochastic trajectories (red solid lines) starting from the initial distribution centered at $\{0,0\}$. This qualitative visualization shows agreement with input data, as the trajectories converge to the steady-state (or equilibrium) minima. In \cref{fig:DW_pre_rec_err_R_1to6} we show quantitative results.  
%	For comparison, we used SINDy \cite{SINDy_2016} coupled with a cubic polynomial dictionary and input based on an average trajectory of the population data over each measurement timepoint. As expected, SINDy is not able to capture the bi-modality of population data, nor it is designed to be able to infer the stochastic noise coefficients $\sigma$. None of the terms in \Cref{eq:SDEs_N2} is correctly identified. 
	%The two modes are $\pm 1$ corresponding to the double well potential attaining minima at \{+1,-1\} for $x_1$ and \{+0.5,-0.5\} for $x_2$ respectively.
	} 
	\label{fig:two_sp_FP} 
\end{figure}
The deterministic drift term $a(x_1,t)$ consists of a double-well potential $U$ whose force is $a(\mathbf{x}) =-\nabla U(\mathbf{x}) =-\nabla(\frac{x_1^4}{4}-\frac{x_1^2}{2})=-(x_1^3-x_1)$. \Cref{fig:two_sp_FP} shows how the joint population data distribution of $x_1$ and $x_2$ evolves in time, approaching four potential minima. 
The importance of this demonstration lies in that our proposed PDL algorithm can handle multimodal population data, meaning that the initial distribution of samples splits into two sample regions as time progresses (see \Cref{fig:two_sp_FP}).
The known nonlinear analytic \Cref{eq:SDEs_N2} is used to generate noisy data of different stochastic noise level $\sigma_n$, given as input to our PDL algorithm.

%mention analytic solution as for the OU process in the Bioinformatics paper
%The stationary distribution $\mu(x)$ of \Cref{eq:SDEs_N2} is given by
%\begin{equation}
%    \mu(x)\propto e^{-\frac{1}{\sigma^2}x^T \Sigma^{-1}x}
%\end{equation}{}
%The inverse of the covariance matrix $\Sigma$ is not necessarily sparse even when $A$ is sparse. This implies that if measurements are obtained as i.i.d. 
%samples from the stationary distribution then it is impossible to infer the causal relationships between the variables
%since they have been lost. On the contrary, using richer information that is contained in the dynamics, primarily time correlations (in either one long or multiple experiments),
%the true connectivity matrix $A$ is estimated and the causal relations can be correctly inferred.
%[maybe shift to Appendix...]

A polynomial dictionary including at least up to cubic terms is sufficient for this example, though we experimented with higher order dictionaries as well (see supplementary). 
In principle, richer dictionaries express the dynamics better while on the same time sparsity penalizes possible overfitting, though higher order terms can compensate for lower order ones and in the presence of highly noisy data, falsely provide a different solution due to unidentifiability. % \cite{Schaeffer_2017} last example on solitons...
For this two-dimensional case, the dictionary consists of 10 variable-combination terms in total; $\psi(x)= \{1, x_{1}, x_{2}, x_{1}x_{2},\dots, x^{3}_{2}\}$. The inference problem \cref{eq:SLE_wrt_M} lies
in defining the multiplicative constants of these terms $\mathbf{a_n}=[a_{n,1},\cdots, a_{n,10}]$, for each variable, plus the diffusion constants $\sigma_1,\sigma_2$. Hence structure (feature selection of active terms in the dynamics) and parameter estimation (coefficients of active terms) is achieved. 
The approximated matrix $\hat{A}$ having rows consisting of these multiplicative coefficients for each variable, is termed connectivity (or parameter or coefficient) matrix.  

A lower value in the linear term constant in \cref{eq:SLE_wrt_M} (shallower double-well)
results to slower convergence of the corresponding process, so for a successful inference across variables, the data should capture the dynamics
of all the variables. 
According to the available data, an appropriate choice on the number of test functions (in space and time) has to be made; less test 
functions (space: bigger discretization $dx$ intervals, time: low frequency modes) result to a {\it coarser zoomed-out overview of the dynamical information} whereas utilization of a large number of test functions
(very fine $dx$ mesh, additional high frequencies) results to poor sampling of the integrals in the weak formulation, given the fact that the number of time points and samples per time point of the dataset remain fixed.

The test functions used for this example are $M_1$ quadratic B-splines in space and $M_2$ Fourier modes in time. We determined that $ M_1=16$ and $ M_2=31$ suffice and thorough experimentation for setting hyper-parameters can be found in the supplementary material. The support of the B-splines used here, is defined by the range of each variable and by the variance of each cloud. In other words, the interval
length under the non-zero area of a B-spline should be comparable to the average variance of the clouds of samples, in order to {\bf decipher their
displacement}. In this example case, the average cloud width of variable $x_1$ is $0.4$ whereas for $x_2$ is $0.2$, meaning that the support of each B-spline should be in this range. By employing $M_1=16$ equally spaced B-splines over $[-2:2]$, their support is $0.25$, resulting to the
model with the smallest relative error (discussed in more detail in the next section and in the Supplementary Material).
%Alternatively, data driven spatial test functions
%do not require this parameter tuning step [ongoing work], though it is easily incorporated as a preprocessing stage. [future work]

\begin{figure}[h]
\centering
	\includegraphics[width=0.28\paperwidth]{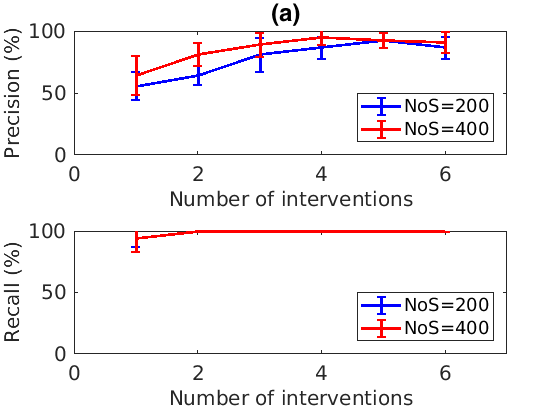}
	\includegraphics[width=0.28\paperwidth]{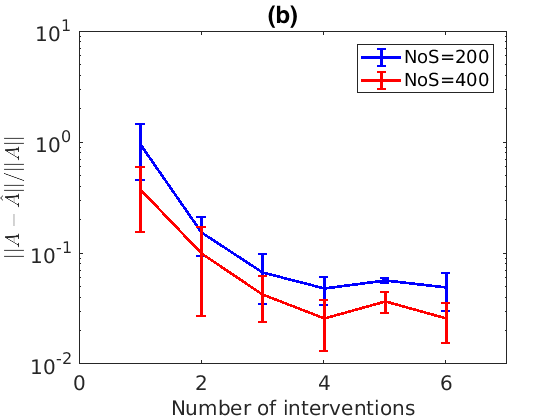}
	\caption{Quantitative estimation of quality of inference for the quadruple-well example. {\bf{(a)}} Precision and recall curves for increasing number of interventions having 200 and 400 samples per measurement time point (cloud of samples). Interventions are activations with randomly picked initial distribution $p_0$, providing rich dynamical information, improving the identifiability of the unknown system.
	 Matrix $A$ contains the coefficient of each constructed feature which are candidate terms of the unknown system of differential equations (see \cref{fig:PDL}).
	 Precision and recall indicate that the structure of inferred coefficient matrix $\hat{A}$ is recovered as we employ more activations. In addition, the relative error {\bf{(b)}} measures the distance of inferred $\hat{A}$ coefficients from the true $A$.
	Error bars are calculated over 6 (IID) iterations (each including randomly picked activations from a set of initial distributions $p_0({\bf x})$). $M_1=16$ B-splines spatial test functions and $M_2=31$ Fourier modes are used (more information on the setting of these parameters can be found in the Supplementary). 
	%[Uneven intrinsic noise in the SDE's is $\mathbf{\sigma}=\{\sigma_1=0.2, \sigma_2=0.1\}$].
	}
\label{fig:DW_pre_rec_err_R_1to6}
\end{figure}

% explanation of the experimental results
%---------------------------------------------------------
For this example system, we generate NoS trajectories of \Cref{eq:SDEs_N2} per intervention, via the Euler-Maruyama numerical integration scheme. A key assumption of population data is that samples are destroyed upon measurement (on measurement time points), so every sample comes from a different realization of dynamics up to that time point. 
Interventions are randomly picked activations (as discussed in \cref{sec:intervention}) from a pool of initial configuration distributions $\{p^{1}_0(\mathbf{x}), p^{2}_0(\mathbf{x}), \dots\}$ (see Supplementary).  
From a dynamical systems point of view, it is not guaranteed that spawned trajectories from a single small subset $p_0$ of the spatial domain can explore the system phase space thoroughly. In our case this translates in a roughly evenly spread amount of trajectories between the two meta-stable regions of the double well over time. 
For this reason, we employ additional activations (additional data sets) in order to correctly identify the terms involved in the underlying equation, that is, to include all the dominant terms (100\% recall) and exclude additional terms (100\% precision). 
\cref{fig:DW_pre_rec_err_R_1to6} shows the performance of the PDL algorithm over multiple interventions.
As the number of samples increases (red line), the accuracy of the algorithm improves. The same holds when additional interventional data in the form of activations are included. The ``relative error" a posteriori metric
(defined as $rr=\frac{||\hat{A}- A||_2}{||A||_2}$) dictates improved inference as well.
\iffalse
[For this example, a backward phase of hard thresholding small values ($\leq 0.01$) of $\hat{A}$ after OMP is performed.
We could have used a heuristic method such as soft iterative thresholding accompanied with a stopping criterion as in \cite{Clementi_JCP18}, though after inspection and new data generation, error terms were of $\mathcal{O}(10)^{-2}$. 
Their presence in the model of generating new data did not affect similarity against the training data (no overfitting). Here, prior knowledge is added for the existence of diffusion terms by starting OMP with a non-empty index set and corresponding initial residual error.]
\fi

%---------------------------------------------------------
\subsection{Inducing the dynamics of a synthetic protein reaction network: Cascade}\label{sec:protein_net}

\begin{figure}[h]
\centering
    \includegraphics[width=0.6\paperwidth]{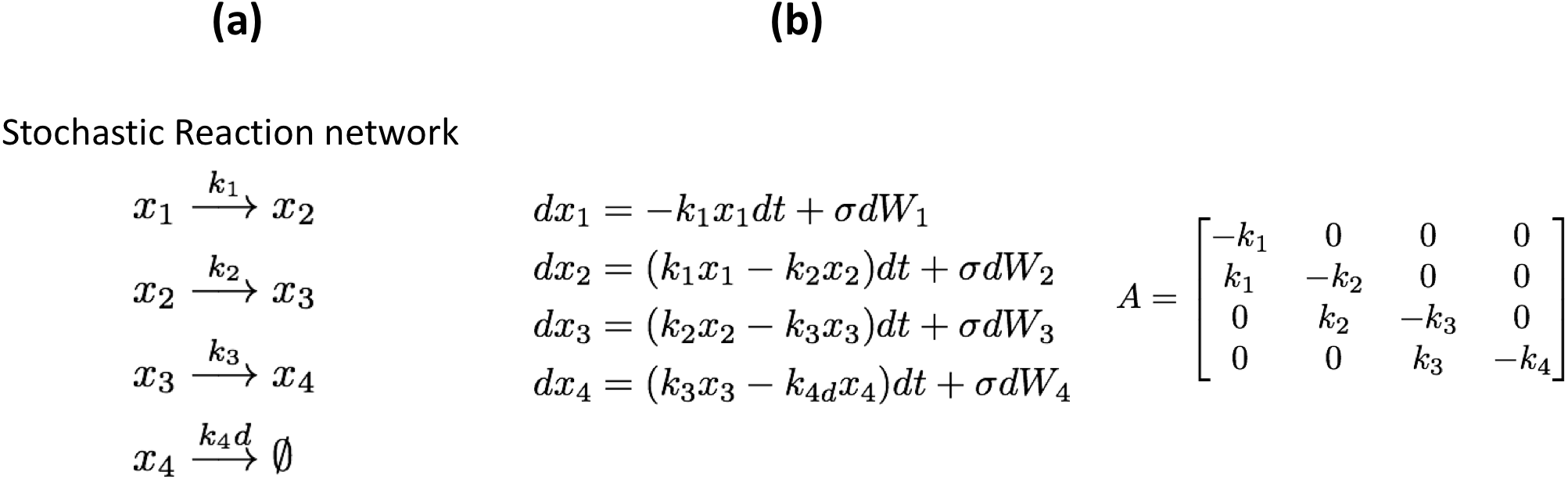}
	\caption{${\bf(a)}$ System of four Stochastic Differential equations resembling a reactions network cascade of four proteins (variables). $x_1$ is converted to $x_2$ at rate $k_1$, $x_2$ to $x_3$ at rate $k_2$, $x_3$ to $x_4$ at rate $k_3$ and $x_4$ is depleted at rate $k_{4d}$. ${\bf(b)}$ Linear SDE system which corresponds to (a), compactly written as $\dot{X}=AX+\Sigma W$. 
	The last term in each equation is the diffusion, $W_n$ is a Brownian motion and constant $\sigma$ diffusion coefficients as in \Cref{sec:double_well}.
	Training data are generated by simulating multiple trajectories of (b), each starting from an initial distribution $p_0$. Then samples per measurement time point constitute a cloud and collectively clouds are the population data, given as input to the PDL algorithm.
	Ultimately we infer the approximated connectivity matrix $\hat{A}$ consisting of the reaction rates $k_q$ and thus, from the structure of $\hat{A}$, recover the variable network in (a). }
	\label{fig:cascade_4sp}
\end{figure}

We proceed with another synthetic data example system based on mass-action kinetics and carry out network inference from the recovered differential equations.
We construct population data from a four-variable stochastic reaction system (\cref{fig:cascade_4sp}(b)) by generating multiple trajectories starting from an initial
distribution $p_0\sim \mathcal{N}(0,\Sigma_0)$.
The motivation is the thorough understanding of the PDL inference capabilities and limitations, in a complex system that can be tuned in every aspect.
Generalizations of this kind of dynamical systems (reaction networks) are applicable to various scientific/industrial fields and in \Cref{sec:mass_cyto} we proceed with real mass-cytometry data resembling this paradigm.
Depending on the choice of dictionary one can, 
in principle account for unary, binary (i.e. $x_1\rightarrow x_2x_3$) or higher order species (variable) interactions, though we restrict to the linear case (\Cref{fig:cascade_4sp}).
%In this synthetic biological example case, we construct population data from a deterministic reaction system, simulated with standard ODE solver [{\red cite ISAP?}], which is in turn
%``contaminated" by additive measurement (Gaussian) noise in order to transform it to a more realistic system and demonstrate robustness with extrinsic noise. {\blue Remark: This kind of noise is not modeled by the diffusion term (intrinsic noise), nevertheless we demonstrate the generalization of the algorithm to time-course deterministic data. In this case of zero diffusion, \Cref{eq:FP_special} without the last term is known as the Louville equation which is a special case of the FP.}

In biochemical kinetics applications, each variable might be a set of protein or gene abundances $x_1,x_2,\dots, x_N$ involved in $N$ reactions.
Each reaction $n$ is characterized by a propensity vector containing the reaction rates $k_1,\dots, k_Q$ plus the diffusion coefficient $\sigma_n$, which are unknown and 
constitute the $n$-th row of matrix $A$. Our primary goal is to infer the sparse matrix $A$ and in effect, from its structure recover the ground truth reaction network of interacting variables \cref{fig:cascade_4sp}(a).
{\it Matrix $A$ encodes the direct causal interactions within the set of variables, so if element $a_{nq}$ is zero, then no direct causal interaction from $x_n$ to $x_q$ exists.}

\begin{figure}[h]
%\centering
    %\includegraphics[width=0.8\paperwidth]{./figs/cascade_raw_vs_generated_dt0_5.png}
    %\includegraphics[height=17em]{./figs/Cascade4_training_and_forecasting_plus_SINDYinBlackcrop.png}
    \includegraphics[width=0.7\paperwidth]{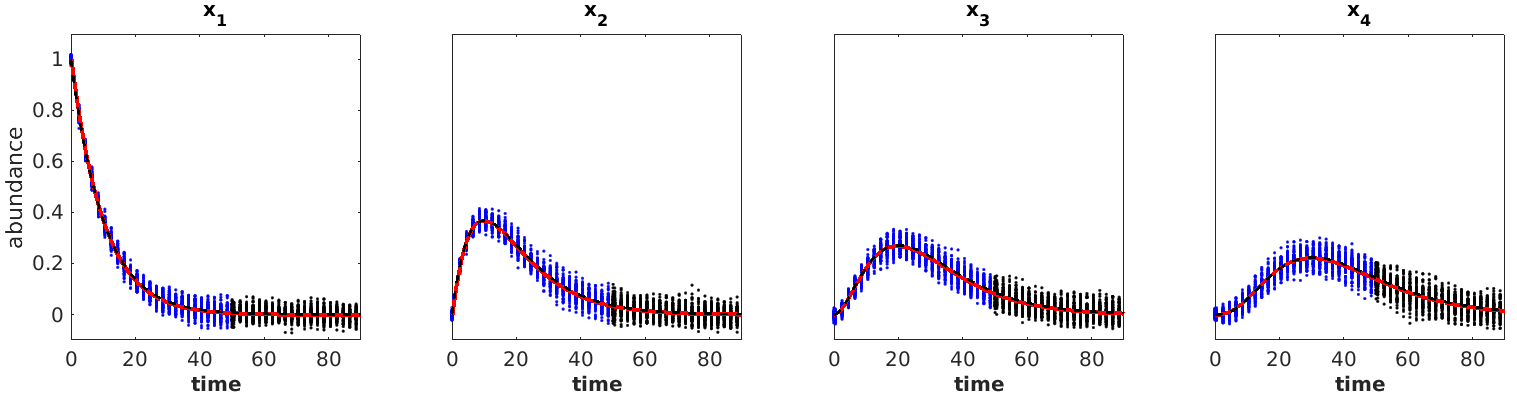}
	\caption{Population data (dots) for the four-variable system of SDE's (synthetic protein Cascade) of \cref{fig:cascade_4sp}. % starting from Gaussian $p_0$.
	%\sim \mathcal{N}([1,0,0,0],[0.01,0.01,0.01,0.01]^2)$. 
	The training data up to time $t= 50$ are shown in blue, whereas the test data (black) lie in the time interval $[50:90]$.
	%Each propagating ``cloud" of data consists of $NoS=400$ samples. 
	The training data are given as input to the PDL algorithm and inference of the unknown connectivity matrix $\hat{A}$ represents a system of SDE's similar to \cref{fig:cascade_4sp} (b).
	Next, we simulate the SDE dynamics based on $\hat{A}$ using a numerical scheme, and generate a trajectory, here shown in red solid lines and estimate the goodness of fit in the training set (for clarity, we set the $\sigma=0$ although it is accurately recovered).
	In addition, the test set provides a test bed for forecasting of the identified system, where other methods such as extrapolation (eg using Bezier curve) would fail to capture the curvature of the dynamics.
	Since variables $x_1,x_2,x_3,x_4$ depend sequentially, each one's intrinsic stochastic noise $\sigma_n dW_n$ propagates along time in the form of increasing variance until reaching a steady state where it remains constant (here $\sigma_n=0.01$ for all four variables). 
	For comparison, we used SINDy with a linear dictionary and as input, one average time series based on $80\%$ of randomly peaked samples of each cloud. The generated trajectories from $\hat{A}$ by SINDy are shown in black dotted lines and the inferred $\hat{A}$ is almost correct, although SINDy is not designed to recover the stochastic noise coefficient of the SDE's.
	Both algorithms attain a relative error of $0.02$ with respect to the ground truth connectivity matrix $A$, so the network in \cref{fig:cascade_4sp} is recovered. We conclude that PDL attains good forecasting for various stochastic noise levels, with the added benefit of noise coefficient inference over SINDy.  
	%Population data sample clouds difference is $dt=0.5$, though for illustration purposes we plot every $dt=2$ time units.
	 %[Mention SINDy in main text?]
	 }
	\label{fig:cascade_4sp_results}
\end{figure}

\cref{fig:cascade_4sp_results} shows the population data distribution evolving in time along with generated trajectories upon inference of the dynamical system with the PDL algorithm (red solid lines). 
Success of inference is not affected for this one-intervention (activation) data example, as long as the  time difference between clouds $dt$ in regions of steep dynamics is relatively small. 
%[This is not a limitation of the method but of the data, not carrying enough temporal information to...] 
For instance, we set measurement time difference $dt=0.5$ time units between clouds, because the steepest derivative (fast dynamics) in the interval $[0:20]$ is that of variable $x_1$. 
In real biochemical reaction sampling, measurement time points are very scarce, two orders of magnitude less than the measured samples per time point and pairwise further away as time progresses.  
The sequential association (pairwise coupling of a cascade) between variables in this specific demonstration implies, that unsuccessful inference of the first variable, affects correct prediction of trajectories of the whole system.
The inference breaks down for this example for $dt\geq0.8$, where the rate of change in abundance of $x_1$ is very steep to be accurately captured. For this reason, inference is equally satisfactory (wrt relative error) if we set $dt\geq0.8$ in the interval $[20:50]$ and while maintaining $dt=0.5$ in $[0:20]$. 
In practice, despite the scarcity of measurement times, protein abundances smoothly increase and subsequently decrease as shown variables $x_2,x_3,x_4$ so variable $x_1$ is an exaggerated example. On the other hand, the noise $\sigma$ (associated with the variance of each cloud) should not be very high with respect to the range of the propagating mean protein abundance, as very high signal-to-noise (SNR) ratio can interfere with the recovery of a meaningful solution $\hat{A}$
(by meaningful we imply that generating new data based on inferred $\hat{A}$, the trajectory neither diverges nor underfits the training data). This is an unidentifiability problem because many SDE systems can be candidate solutions. As demonstrated in the quadruple-well example, additional data from interventions improve the learnt system and can compensate for high SNR.
In \Cref{fig:cascade_4sp_results} we have separated the data in training (blue) and test (black) subsets, corresponding to different time intervals. In this way we assess the success of the generated model in forecasting. 
An extrapolation method would fail to capture the curvature, whereas by learning the governing equations, projection in time is accurate.

\begin{figure}[h]
    \includegraphics[width=0.17\paperwidth]{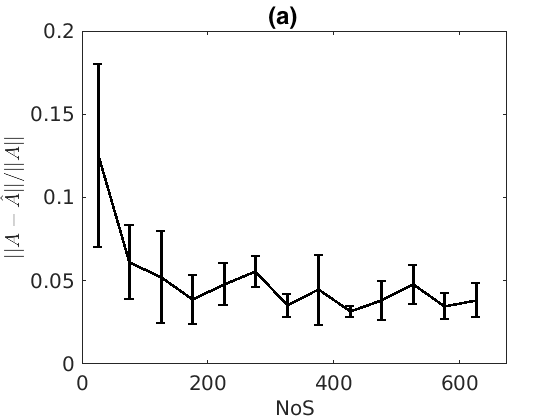}
    \includegraphics[width=0.17\paperwidth]{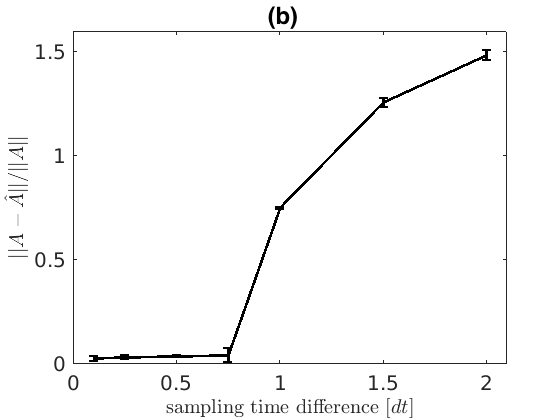}
    \includegraphics[width=0.17\paperwidth]{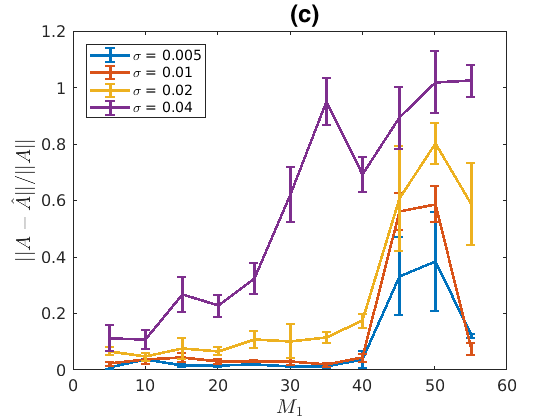}
    \includegraphics[width=0.17\paperwidth]{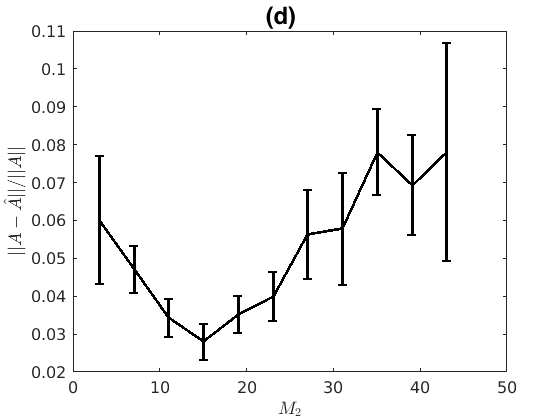} 
    \caption{PDL hyper-parameter tuning for the four protein Cascade system.
    Relative error quantification over: ${\bf(a)}$ increasing number of samples per time point, ${\bf(b)}$ increasing sampling time difference $[dt]$ between clouds, ${\bf(c)}$ increasing number of spatial test-functions $M_1$, ${\bf(d)}$ increasing number of temporal test-functions $M_2$ for different variance of clouds of samples (i.e. different stochastic noise coefficient $\sigma$).
    The optimal parameters that are set constant while varying the others as shown in ${\bf(a)}$-${\bf(d)}$ are $NoS=400$, $dt=0.5$, $M_1=25$, $M_2=15$ for this example of stochastic noise coefficient $\sigma=0.01$ and errorbars are over 6 i.i.d runs.
    We deduce that more samples per sampling time point improve the integral estimations of \Cref{eq:weak_form_special},  resulting in lower relative error as shown in ${\bf(a)}$.
    In ${\bf(b)}$ we see that as the sampling time difference between the clouds of data is increased more than $0.75$ time units for this system, successful recovery of $\hat{A}$ is not feasible.
    ${\bf(c)}$ demonstrates that by employing more B-splines, their support is decreased and as a consequence the relative error increases. The inferred solution $\hat{A}$ collapses (relative error sharply approaches $1$) when the support of the B-splines is much smaller (over 4 times smaller) than the variance of each propagating cloud. Increasing values of stochastic noise $\sigma$, resulting to wider clouds, further verify this result.
    Temporal test functions parameter shown in ${\bf(d)}$ is less sensitive wrt the relative error. The  optimal value of Fourier modes to be used is 15 for this example clouds sampling time difference set at $dt=0.5$ time units.}
	\label{fig:cascade_4sp_RelErr}
\end{figure}

%--------------------------------------------------------
\subsubsection{Setting algorithm hyper-parameters}
%[{\red \small Mention limitations of PDL and when proper choice on dictionary, dt, M1, M2 not properly met it breaks down.}]
Next we quantify the impact of the data parameters of: number of samples $NoS$, sampling interval $dt$ and algorithm parameters: $M_1, M_2$, that need to be tuned for the successful inference of the unknown system. 
The ground truth $A$ is known, so a metric such as the $L_2$ distance or ``relative error" is used here. Otherwise we use the $L_2$ average distance between the generated trajectories based on $\hat{A}$ and input data averages, as a metric (\cref{sec:mass_cyto}). 
\Cref{fig:cascade_4sp_RelErr}(a) shows the expected reduction in the relative error with increasing number of samples per cloud $NoS$, by improving the estimation of the integrals in \Cref{eq:weak_form_special} thus reducing statistical noise.
Even for $NoS=100$, the principal terms (the terms present in every $\hat{A}$ estimate over randomly chosen test data sets) in the recovered solution $\hat{A}$ are very close to those of the ground truth $A$ and a plateau is reached over $NoS=350$. 
We note that the relative error metric based on the inferred matrix $\hat{A}$ and ground truth $A$ is not an unbiased estimator, meaning that we cannot get arbitrarily close to zero by using more samples.

In \Cref{fig:cascade_4sp_RelErr}(b) we see the effect of sparser sampling times (higher $dt$) on the relative error. As intuitively expected, when the 
clouds are further away over the same time horizon, the dynamical information is less and the system becomes less identifiable as discussed in the beginning of \cref{sec:protein_net}.
Sparser sampling times when fast dynamics occur, negatively affect the projected derivative estimation on the left hand side of \Cref{eq:SLE_wrt_M} and consequently the relative error.
For this system we observe a sharp increase in the relative error for $dt\geq0.8$ time units. 

In \Cref{fig:cascade_4sp_RelErr}(c), as the number of spatial test functions (B-splines) $M_1$ increases, the support of each B-spline is narrower since the range of the spatial domain remains fixed at $[-0.2:1.2]$ as defined by the training data (abundance range). In effect, the sampling of
each integral of the weak form in \Cref{eq:ibp_2} is based on less samples thus providing a worse estimate. 
Moreover, the support (or width) of each B-spline should be equal or greater than the variance of each cloud in order for the algorithm to track their evolution over time. In other words, a cloud captured by only one B-spline at a given time point once, does not carry enough information that can be subsequently projected to temporal test functions and constitute a useful projected pdf. 
In this example, where the variance of the steady state clouds of samples is around $0.1$ ($\sigma=0.01$), the relative error spikes to $1$ (cyan line) when $40$ B-splines are used with $0.035$ spatial units support. Increasing values of stochastic noise $\sigma$, resulting to wider clouds, require wider B-splines hence smaller $M_1$.

The number of temporal test functions $M_2$ (Fourier modes) used, is related to encoding  spatially projected information over time and depends on the curvature of the clouds over time and the placement of nodes of spatial test functions. Faster changes in time (steep curvature) require higher frequency sinusoidal functions.
In \Cref{fig:cascade_4sp_RelErr}(d), by employing three temporal test functions $M_2$, only the low frequency changes are identified. On the other hand, higher frequency sinusoidal functions encode noise thus deteriorating the relative error. 
This limitation of constantly improving the relative error by using more sinusoidal over fixed sampling rate [rate is determined directly by $dt$ and its variability by $M_1$], is related to the Nyquist–Shannon sampling theorem in conjunction with the sampling time difference used in \Cref{fig:cascade_4sp_RelErr}(b).
Although this hyper-parameter is less sensitive with respect to the others,
we conclude that the determination of $M_2$ is problem specific and tuned 
in accordance to $M_1$, with respect to the metric chosen.
More detailed experiments indicating the inter-dependence of these parameters are provided in the supplementary material (S.M.5).

%---------------------------------------------------------
\subsection{Biochemical protein reaction network inference using Mass cytometry data}\label{sec:mass_cyto}
We proceed and evaluate our proposed
methodology on real publicly available mass cytometry data \cite{Krishnaswamy_DREMI}.
Mass cytometry cell analysis techniques are important for understanding cellular responses (stimuli from other cells, signals etc) by measuring tens of interacting proteins in each cell simultaneously (predefined times), over millions of cells.
Given the high resolution it is expected to become a standard technique in medical sciences in the near future. 
Reconstructing the pathway upon activation (ordered relation and intensity between proteins) is a non-trivial task because only a few proteins inside the cells are measured and on top of that many interfering mechanisms with different
rate are also occurring. 
%Both result in a large number of latent confounding factors and in our point of view this current stochastic USDL counterpart is suitable to tackle this problem. 
Considering the above, it is very hard to reconstruct directly the complete system of interactions.
However, network reconstruction would be more successful if restricted to subnetworks. 
The abundance of each protein can be described as a stochastic process affected by the abundance of other proteins, so the underlying model is chosen to be an SDE for each one.% [{\blue Restate wrt the last paragraph of this section}]
 Using our Fokker Planck formulation, the propagating density of abundances in the form of population data, is a novel approach to this statistically limited data regime.
Next, we focus on a subnetwork of four and eight proteins (supplementary).
Our findings 
are compared qualitatively against other studies \cite{Krishnaswamy_DREMI,USDL_bioinformatics} and from the KEGG database, since the exact interactions mechanism are unknown.
%We opted to model the protein interactions with a dictionary composed by linear terms (higher order dictionary results can be found in the Supplementary Material). The protein interactions found
%in literature are summarized in [refer to Tcell network graph in Supplementary] and constitute the basis on what is considered as true positive (TP).

Experimental data belong to the population of naive $CD4^{+}$ regulatory T-cells and the activation cocktail which stimulate the receptors CD3/CD28 were applied. We focus on the subnetwork
containing proteins pCD3z, pSlp76, pErk and pS6, downstreaming the signal in this order (cascade). The mass-cytometry measurement times are closer at earlier times, where dynamics prevail though scarce  as one can see in \cref{fig:Dremi_data_inferred_X1_colloc} (upper left).

In the data set under study and mass cytometry data sets in general, sampling times are scarce resulting to poor dynamical information. Hence the algorithm would not able to deduce the underlying interactions correctly without more frequent temporal measurements (as shown in the supplementary material), as there exist multiple dynamics that could give rise to the measured distributions or in other words, multiple possibilities for the underlying mechanisms (see \cite{Weinreb_PopulationBalanceAnalysis_PNAS_17} for a thorough explanation on inference limitations on single-cell data). 
The current framework assumes Gaussian stochastic noise over each cloud, which does not hold upon inspection of the histograms.
Variable variance as time progresses is another peculiarity of this data-set, as our current
framework assumes constant $\sigma$. 
Last, smoothing out the dynamics of the propagating distributions is desirable 
in this context of: very limited measurement times, no additional interventions and high measurement noise of unknown number of unmeasured variables.
For these reasons, we propose re-simulation of the input population data, by the collocation method (details in the supplementary and previous our publication \cite{USDL_bioinformatics,Ramsay_Book,Ramsay_2007_parameter_estimation}) and all the analyses in this subsection are based on re-simulated data. 
The collocation method, is a regularized least-squares minimization, taking account of the variance of each cloud of samples, along with their relative distance in time and thus provides a well-informed time-series (or multiple time-sereies in case of multimodality) on which population data are simulated.
Re-simulation is vital for obtaining a meaningful solution under this particular regime, taking account of the advantages and limitations of PDL as discussed in \Cref{sec:double_well} and \Cref{sec:protein_net}. A more advanced method for curve fitting time-course gene expression data, along with a review on this subject can be found in \cite{Bspline_curve_fitting_on_timecourse_data_Liu2006}, though the collocation method is satisfactory for our demonstrations.

We remark that by construction, there is no need of random term to be included in the inference, thus $\Sigma=0$ in \Cref{eq:FP_special}. Nevertheless each cloud of samples is based on a Gaussian distribution resembling measurement noise and the theoretical deductions from \Cref{eq:FP_special} up to \Cref{eq:ibp_2} hold, known under the name of Liouville equation.

%---------------------------------------------------------
%\subsubsection{Protein pathway}
%We aim to infer the four protein pathway CD3z $\rightarrow$ Slp76 $\rightarrow$ Erk $\rightarrow$ S6.
Despite the fact that the complete biochemical network is probably nonlinear with respect to 
the variables, the assumed model for inference is linear $\frac{dX}{dt}=AX$, similar to \cref{sec:protein_net} (for $\Sigma=0$). The linear model is sufficient for this low dimensional\textbackslash single maximum paradigm and at the same time, the associated connectivity matrix $A$ encodes only the direct causal interactions between the variables.
Another inherent difficulty on the inference of this example is: a) the measurement error associated with machine errors, assumed to be additive and b) uncertainty error due to the 
fact that each measurement comes from a different cell and each cell has different concentrations of the measured quantities.

%----------------------------------------------
\subsubsection{External forcing as prior knowledge: Four-protein pathway}
A major modelling assumption of the formalism is that the modelled system is closed. 
Intuitively this means that only the variables considered, account for changes on each other over time (where non-modeled or stochastic forces are on top of these changes, casting the distributional clouds).
On the contrary, this does not hold true for a real biochemical pathway where the proteins are sequentially converted to the next and in which we merely focus on a fraction of the variables. 
Considering all the aforementioned, we add driving protein CD3z to the inference, infer CD3z, use its data
though never use it to generate trajectories for that particular particular variable. 
%CD3z is involved in the numerical integration schemes only. 
In this way we extract information from this
driving variable, improve the inference of the other three, at the cost of not having a meaningful SDE describing CD3z (see demonstration in supplementary). 
PDL returns an SDE for CD3z that is good fit, though not meaningful as based on the others. 
The reason for the problematic inference for this variable arises because a combination of the linear dictionary terms,
starting from an initial abundance distribution $p_0(X)$, is used to describe the curvature of the sole driving variable, {\it in a closed system}. 
Hence additional interactions of the modelled variables are devised to account for the {\it non-monotonic} behavior of CD3z, which in principle is regulated by unmeasured non-modeled variables. %[rephrase?]
Using but not inferring CD3z consists prior knowledge to the system because from a biological point of view, the CD3z protein lies in the cell surface being one of the gatekeepers of T-cell activation (signal transduction) \cite{Tcell_CD3}.

\begin{figure}[H]
%\centering
    \includegraphics[width=0.17\paperwidth]{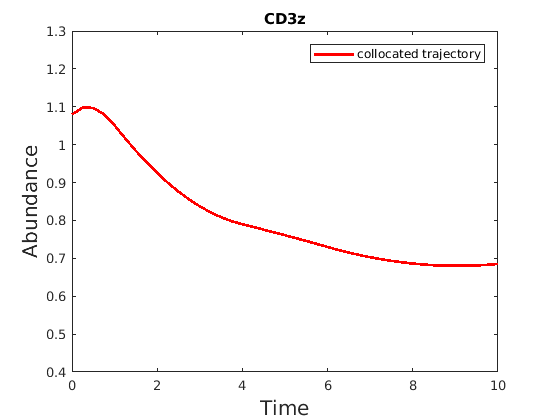}
    \includegraphics[width=0.17\paperwidth]{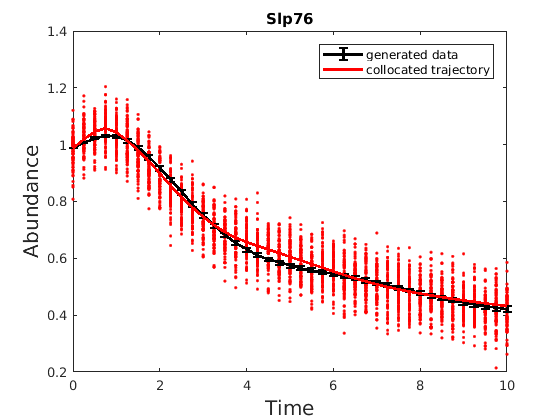}
    \includegraphics[width=0.17\paperwidth]{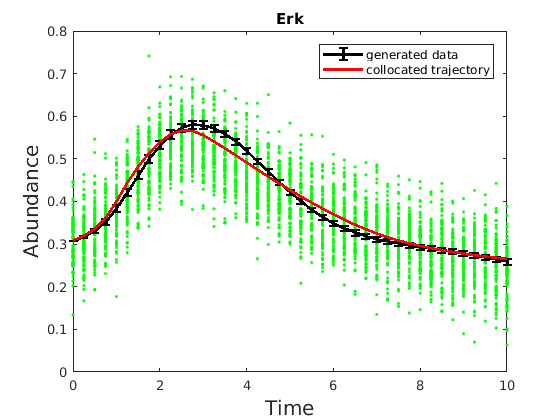}
    \includegraphics[width=0.17\paperwidth]{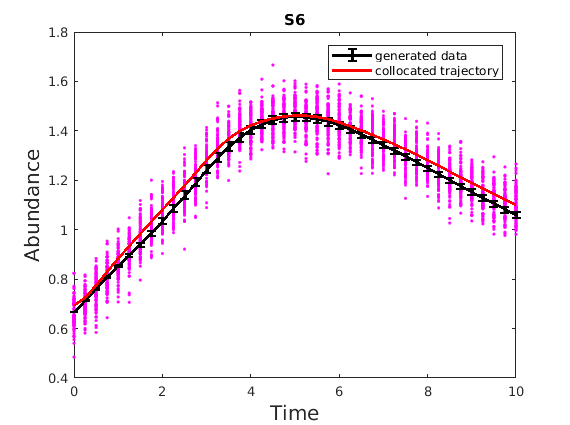}
    \includegraphics[width=0.01\paperwidth]{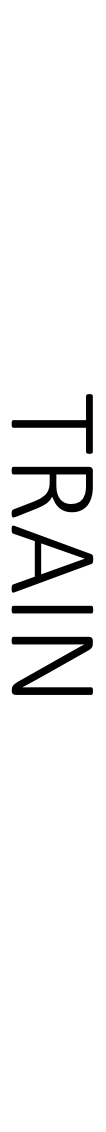}\\
    \includegraphics[width=0.17\paperwidth]{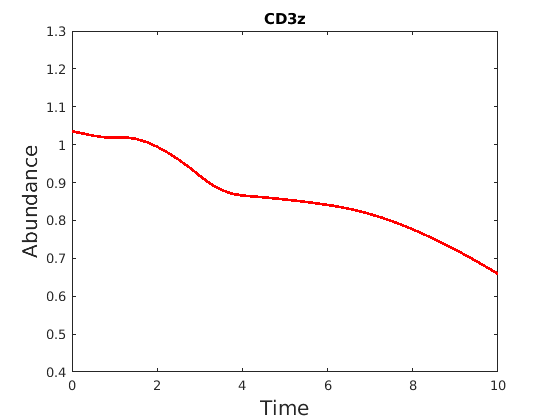}
    \includegraphics[width=0.17\paperwidth]{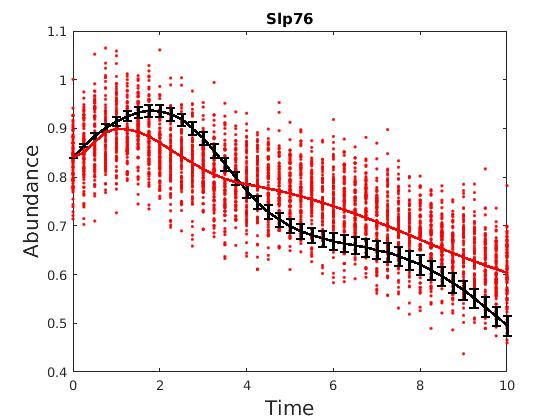}
    \includegraphics[width=0.17\paperwidth]{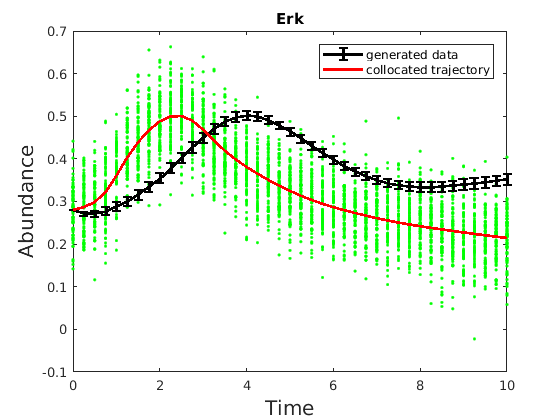}
    \includegraphics[width=0.17\paperwidth]{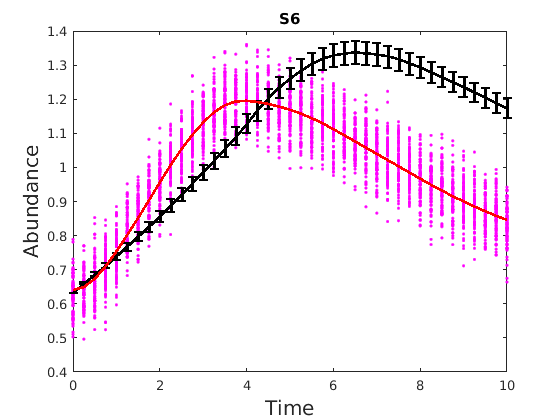}
    \includegraphics[width=0.01\paperwidth]{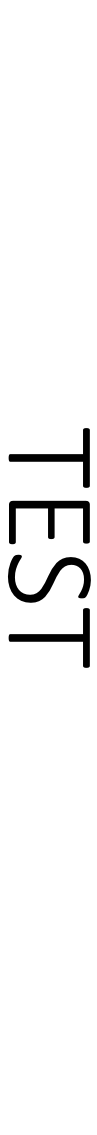}\\
    \includegraphics[width=0.17\paperwidth]{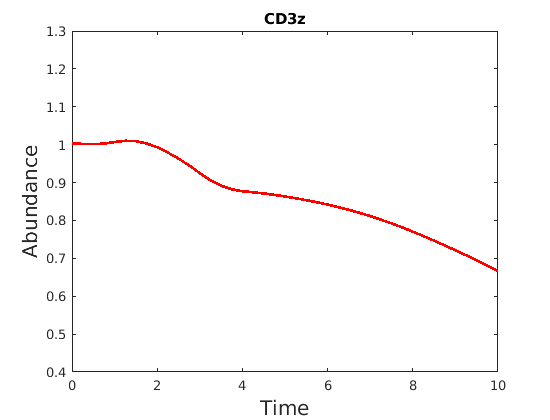}
    \includegraphics[width=0.17\paperwidth]{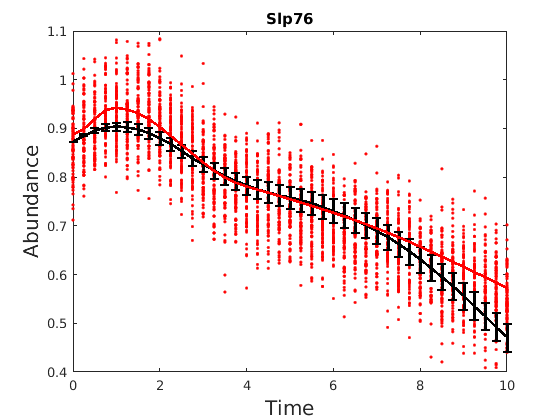}
    \includegraphics[width=0.17\paperwidth]{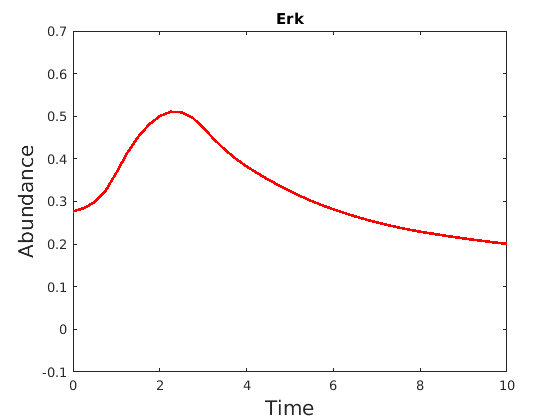}
    \includegraphics[width=0.17\paperwidth]{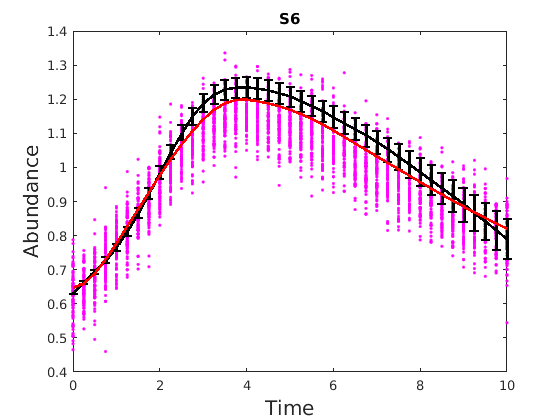}
    \includegraphics[width=0.01\paperwidth]{./figs/test_label.png}\\
    \includegraphics[width=0.19\paperwidth]{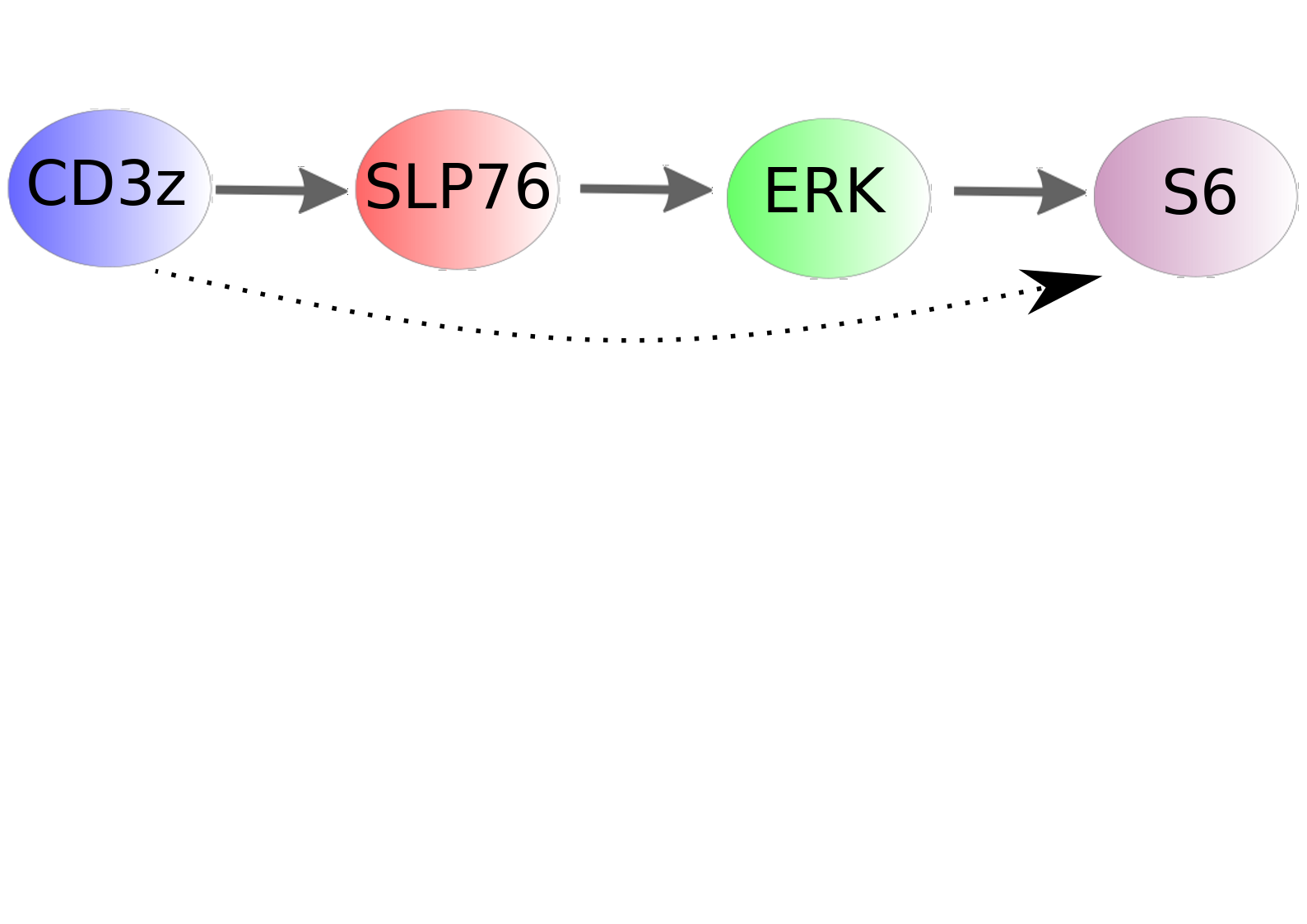}
    \tiny
    \begin{tabular}{|c ||c |c |c |c |}
    \hline
    protein & CD3z & SLp76 & Erk & S6 \\
    \hline
     $||X_{coll}-X_{gen}||_2$ (train) &  {\blue 0} & 0.026 & 0.047 &  0.029    \\
    \hline
     $||X_{coll}-X_{gen}||_2$ (test) &  {\blue 0} &0.066 & 0.33 & 0.28 \\   
     \hline
     $||X_{coll}-X_{gen}||_2$ (test) &  {\blue 0} &0.035 & {\blue 0} & 0.072 \\   
     \hline
    \end{tabular}
    \caption{%(upper left panel)
	%Mass cytometry population data (dots) of four protein CD3z, SLp76, Erk and S6 abundances, along with collocated trajectories (solid lines), which are constrained averages. The collocated trajectories characterize the re-simulated data. 
	Population data (dots) of four protein CD3z, SLp76, Erk and S6 abundances, based on re-simulated data as explained in \cref{sec:mass_cyto}.
	(first row panels) The average generated trajectories based on inferred $\hat{A}$ matrices of 25 training sets are shown in black solid lines with error-bars, and their  
	average $L_2$ distance from the collocated trajectories (red solid lines) is shown in the table. The driving protein CD3z is given as prior knowledge as mentioned in the main text. We used a quadratic dictionary which results to a more accurate $\hat{A}$ in comparison with the linear, based on the $L_2$ distance, although the pairwise associations between variables remain (graph). 
    The difficulty of this example is finding a model that is minimally complex (low order terms) and at the same time capture the dynamics (curvature) and explain the interactions (qualitatively as graph and quantitatively as equations). 
	The graph shows the inferred relations based on the structure of $\hat{A}$ and the dashed edge CD3z$\rightarrow$S6 is a false positive that is not found in the bibliography, possibly an artifact of limited temporal measurements. We conclude that the inferred models $\hat{A}$ predict the behaviour of the input data satisfactorily. 
	(second row panels) Use of $\hat{A}$ matrices from training set on test set of another activation-dosage (laboratory test), in order to estimate performance. Our assumption is that the underlying mechanisms remain the same over this protein subnetwork. We see that there is partial transferability of the inferred model, probably due to the difference of the driving protein $CD3z$ affected by unmeasured proteins. Nevertheless, none of the trajectory deviates and qualitatively we claim that the underlying mechanism is the same in both data sets. 
	%[A similar goodness of fit as in the training phase, is attained when swapping test and training sets].
	(third row panels) Extra prior knowledge on Erk further improving the test set fitting, indicating the accuracy of our models and the limited knowledge of unmeasured proteins.
	Further experiments are included in the supplementary material SM1, SM7.
	}
	\label{fig:Dremi_data_inferred_X1_colloc}
\end{figure}

%In this way, we demonstrate that the resulting model over the three subsequent proteins, captures the curvature of their data. Of course, 
As we already mentioned, higher order dictionaries can be more precise though their use in interpreting the interactions is more complex.
In fact, low sample sizes or infrequent sampling, coupled with high order dictionaries run the risk of overfitting. In order to be able to achieve more precise fitting over the linear dictionary (shown in supplementary) and retain interpretability with respect to interactions, we added quadratic terms of the form $x^{2}_{1},x^{2}_{2},x^{2}_{3},x^{2}_{4}$ to the dictionary \cref{fig:Dremi_data_inferred_X1_colloc}.  
%The $n$-th row of $A$ indicates the involved terms in that reaction. The inferred coefficients (rates) in $\hat{A}$ for variables $x_m$ and $x_{m}^2$ cancel out (approximately) and the remaining coefficients indicate by their sign, up-regulation or down-regulation of variable $x_n$. Taking account of the variables that constitute the added dictionary terms, the resulting network of interacting variables is the same as if we used linear terms only [rewrite?]. 
Depending on the application, there is a trade-off between model complexity and interpretability.

\cref{fig:Dremi_data_inferred_X1_colloc} shows the four-protein pathway, where the driving variable CD3z data is used for the inference of the 
rest; proteins Slp76, Erk and S6. 
In this way we use the past of the CD3z for the future trajectory of the SLp76, Erk and S6.
Random subsampling (of size 800 samples per time point) of the mass-cytometry data, result to slightly different inferred dynamical systems or connectivity matrices $\hat{A}$. 
Based on those inferred matrices, we generate data trajectories using a numerical integration scheme ($4$-th order Runge-Kutta), starting from an initial configuration $p_0$ deduced from the population data at $t=0$.
As it is evident from the \Cref{fig:Dremi_data_inferred_X1_colloc} (first row panels), the generated trajectories based on the inferred systems are in accordance with the input population data, capturing the minima correctly.
Examples of these, sequentially dependent systems, are sensitive to lagged or imprecise derivative estimations, because the errors are accumulated down the pathway. 
Next we try to assess the inferred matrices $\hat{A}$ on experimental data (test set) coming from another activation, corresponding to a different dosage. We do not know in advance if we can capture the dynamics correctly, although we expect that the underlying mechanisms should remain the same, 
with a different initial distribution $p_0$. \Cref{fig:Dremi_data_inferred_X1_colloc} (second row panels) shows a satisfactory (non-divergent) fit, not being able to capture the maxima exactly. This might hold for a number of reasons. 
First and foremost, we focus on the modelling of a subnetwork and there might be other, unmeasured mechanisms due to the different dosage. Secondly, we chose a quadratic polynomial that best fits the input data, though it might be less transferable for this dataset. Third, the accumulated error in this cascade, builds on the last protein S6. 
We further add protein Erk as prior knowledge and conclude to an improved estimation for the dynamics of the other proteins of the test set (\cref{fig:Dremi_data_inferred_X1_colloc} last row panels). This implies that there exist hidden mechanisms, in the form of unmeasured proteins, affecting this four protein cascade and one of the assumptions we make is that there is no confounding (\cref{appendix:SDE}). A similar conclusion for mass cytometry data on the differentiation of fibroplast cell line beyond a time point, is discussed in \cite{Dynamic_Distribution_Decomposition_Claassen_2019}.

We show in the supplementary material that a linear dictionary is less accurate than a quadratic dictionary but slightly more transferable. Results on a bigger protein network are included.

%---------------------------------------------------------
\section{Discussion}
Our work lies in the identification of the underlying system of differential equations based on spatio-temporal data.
We have presented a new robust algorithm, PDL, for the inference of stochastic dynamical systems based on population (or time-course) data.
This work builds on our previous framework \cite{USDL_bioinformatics},  extending  from trajectorial (or time-series) data to propagating distributions, with the deduction of the Fokker-Planck formalization.
To the best of our knowledge, 
this is the first approach in deriving both the unknown drift and diffusion equation terms from population data, without transforming it into trajectories instead.
The latter pdf evolution can be multimodal in principle, capturing meta-stability of the underlying unknown system and this is our major contribution to the field.
In the synthetic example cases, perfect identification of the stochastic dynamics was achieved. %, even in the presence of additive noise.
In addition to other methods that capture the deterministic parts of SDE's (terms comprising the gradient of a potential) provided that the dynamics have reached a steady-state, we were able to recover the diffusion coefficient of SDE's (noise term).
%Outliers did not pose any problems in the inference either.
A limitation of the current SDE model, not the PDL framework, is the assumptions of 
i) the diffusion coefficient in the noise term of the SDE being constant over time, 
ii) the Brownian motion random term being normally distributed. Nevertheless, these assumptions can be 
overcome, as shown in the main text, by re-simulation of the data with an equivalent data set. Moreover, deduction of an extended form of the FP equation for multiplicative Brownian motions can be derived analytically for these cases.

The cornerstone of our work, is the weak space formulation of the problem, transforming it to an atemporal one while preserving the unknown coefficients. In this way, we gain robustness against measurement noise, scalability to higher dimensions and longer times provided that an appropriate choice of test-functions is made. 
Along with test-functions, an appropriate choice of dictionary is crucial for solving this inverse problem. A dictionary lacking major terms (low expressive power) of the unknown equation cannot always be compensated by higher order terms, whereas a very rich dictionary (especially in the low sample or high measurement noise regimes) can induce spurious correlations and make sparse regression harder. 
There is no universal solution to this challenge, as it requires prior intuition on the dynamics of the data set in order to provide the algorithm with explanatory features.

The PDL algorithm, correctly identified the non-linear double-well system, having the intrinsic difficulties of bi-modality, different diffusion coefficients and different equilibration times. 
The algorithm is designed such that, additional data in the form of different initial conditions or interventions on variables, can be incorporated (in a straightforward manner on top of previous computations) and bolster its inference capabilities. Towards the same direction is the direct supply of prior knowledge,
in case we have strong evidence of some dictionary terms being present. 
The majority of the algorithms of ODE's or PDE's are based on the approximation of temporal or spatial derivatives by a numerical scheme like finite differences, which are very sensitive to noise of 
``unclean" data. This fact requires smoothing techniques in order to produce reasonable approximations, though 
PDL shifts this burden to the (smooth) test-functions via the weak formulation. 
%In the case of limited temporal data for the algorithm to perform satisfactorily, we augment by additional data in intermediate time points through a simple preprocessing procedure, without altering dynamical information.

In this current form, the algorithm requires the propagating distributions (clouds) to be be measured frequently, in  order to be
captured by the spatial test functions and subsequently, by the temporal test functions. This might prove to be a limitation in specific applications of the algorithm, nevertheless we proposed the collocation method as a practical solution. Although smoothing in time is avoided, we still need temporal
information (either in the form of frequent measurement or with additional interventional data)  in order to disambiguate the alternative dynamics, depending on the smoothness of the evolving clouds. We strongly believe that this limitation can be alleviated with a more sophisticated choice of data-driven test functions, which is under investigation. 
PDL scales well with increasing number of samples (used in the integral estimators), although the current implementation is restrained by the variable with the biggest range, which might prove slow for systems with unbalanced relative magnitudes of  variables.

Application to mass-cytometry data, upon inspection making realistic assumptions on the interactions, concluded to results supported by the literature. Preprocessing by re-simulation of the population data
was mandatory due to uneven, distant measurement time-points along with non-Gaussian skewed pdf's swamped in measurement noise, which violated the algorithm assumptions. The inferred subnetworks of interacting proteins included some additional interactions, though it was expected for this
low-informative data set and our dictionary choice. %almost unidentifiable paradigm.

%So far we tune the algorithm hyper-parameters with respect to the input data, though this procedure can be straightforwardly automated in a preparatory stage. 
%[The sparse regression algorithm OMP [{\blue ?}] stopping criterion threshold is chosen with respect to the BIC scoring. OMP is chosen as the feature selection algorithm for this work, though not mandatory, provided a slight variation of the minimization problem is performed.
%Extension to data driven spatial test-functions eliminating some hyper-parameter tuning, is under preparation.]

%[Additionally, the transformed structure learning problem can be considered not only
%as an SSR problem but also as a feature selection problem [cite Guyon], a subfield of machine learning and statistics.
%The extensive body of work on feature selection could be also employed and, therefore, boost the accuracy of
%the overall inference.]
%What we believe should follow in this field....

%==============================================
%==============================================
\appendix

\section{Model, Notation and Assumptions}\label{appendix:SDE}

The general form of a system of SDE's reads:
\begin{align}\label{eq:SDE_general}
        dX_t = &a(X_t, t)dt + b(X_t,t)dB_t \\
  X_t \in \mathbb{R}^N, \quad &a(X_t,t): \mathbb{R}^N\times \mathbb{R}^{+} \mapsto \mathbb{R}^N   \nonumber \\
     &b(X_t,t): \mathbb{R}^N\times \mathbb{R}^{+} \mapsto \mathbb{R}^{N'}  \nonumber \\
     &B_t \in \mathbb{R}^{N'} \text{is $N'$-dimensional  Brownian motion} \nonumber
\end{align}
where the first term on the r.h.s. of \Cref{eq:SDE_general} is the deterministic term whereas the second term is the diffusion (random forcing). 
$a(X_t,t)$ is a vector field called {\it drift} and $b(X_t,t)$ is a matrix field called the {\it diffusion} coefficient of the process $X_t$. 
$B_t$ constitutes an $N'$-dimensional Brownian motion, so their product acts as a random force on top of the deterministic counterpart, which is normally distributed, 
thus not making the deterministic dynamics diverge in later times. Intuitively, the derivative of a Brownian motion $dB_t$ can be understood as a continuous-time zero-mean white noise with variance one.\\
\medskip \\
\underline{\bf Assumption 1}: The random term describes all the {\bf latent variables} (unknown) acting on the variable being modelled. 
In case where no latent variables are present, the model reduces to an ODE (or PDE)
where the deterministic term $a(\cdot,t)$ takes over.\\

\medskip
We choose an appropriately rich dictionary of functions (in other context termed dictionary atoms) $\psi_q(X(t)): \mathbb{R}^N \mapsto \mathbb{R} ,q=\{1, \dots, Q\}$ which serve as basis functions for the drift term.
\medskip
\underline{\bf Assumption 2}: 
The $n$-th deterministic drift term $a(X^{n}_{t},t)$ can be written {\bf exactly} as a linear combination of the dictionary atoms by:
\begin{equation}
    a(X^{n}_{t},t) = \sum_{q=1}^Q a_{n,q}\psi_q(X(t))
\end{equation}
Note that in general, $\psi_q=\psi_q(X(t))$ meaning that we include non-linear functions such as $X_n^2(t),cos(X_n(t)), X_nX_{n'}$ etc as well. 
The constants $a_{n,q}$ make up the connectivity matrix $A\in\mathbb{R}^{N\times Q}$ of unknown coefficients to be determined later and determines the interactions between variables $X$.\\
\underline{\bf Assumption 3}: Time-invariant diffusion coefficient of each process $X^{n}_t$
\begin{equation}
b(X^{n}_t,t)=\text{constant over time}=\begin{bmatrix}
                  \sigma_1 \\  \sigma_2 \\ \vdots \\ \sigma_N \end{bmatrix}  
\end{equation}
In practice, time invariance means that unmeasured variables do not affect the process $X^{n}_t$ in a different manner throughout the time measurements. More complex forms of diffusion could better reflect reality though this simplification is sufficient for our modelling.

\medskip
The covariance matrix of the diffusion of \Cref{eq:SDE_general} accounts for  interactions between the unknown variables and is given by: 
\begin{equation}
    \Sigma({\mathbf x}) = {\bf \sigma}^T({\bf x}){\bf \sigma}({\bf x}), \quad {\bf x}\in \mathbb{R}^N, \Sigma \in \mathbb{R}^{N\times N}
\end{equation}
\\
\underline{\bf Assumption 5}: The covariance matrix is diagonal:
\begin{align}%{figure}
    \Sigma= \begin{bmatrix}
    \sigma_{1}^{2}& 0 &\dots\\
    0 & \sigma_{2}^{2}& 0\\
    \vdots\\
     & &\dots& \sigma_{N}^{2}
    \end{bmatrix}
\end{align}%{figure}
This implies that there are {\bf no latent confounders} in the dynamics. Only latent variables affecting {\bf one} system variable $X^{n}_t$.
%==========================================================
\section{Measurement Data}\label{appendix:types_of_data}
\subsection{1-Dimensional SDE}
%[{\red CORRECT or REMOVE these two subsections}]
We define $L$ objects $\{ O\}_{l=1}^{L}$ that propagate in time, where $O_{l}$ is a specific item
i.e. protein concentration and $L >> 1$.
%\begin{figure}[h]
%    \centering
%    \includegraphics[height=14em]{./figs/object.png}
%    \caption{Objects propagating in time.}
%    \label{fig:objects}
%\end{figure}
We further assume that time is continuous and each object value at time $t$: $O_l(t)$ 
is given by a stochastic process whose SDE is given by:
\begin{equation}
    d(O_l)_t = \mu((O_l)_t,t)dt + \sigma(O_l,t)dW_t
\end{equation}
where the first term on the r.h.s. is the (deterministic) drift term whereas the second term is the diffusion (random forcing), as in \Cref{eq:SDE_general}.%, see figure (\ref{fig:objects}).\\
%\underline{\bf Assumption 1}: The random term describes all the {\bf latent variables} (unknown) acting on the variable being modelled. 
%In case where no latent variables are present, the model reduces to an ODE
%where the deterministic term $\mu(\cdot,t)$ takes over.
\\

\subsection{Mass Cytometry measurements}

Instead of the 1-Dimensional value each object $O_l(t)$ acquires in time, we get $P$ {\it measured} values, at predefined measurement times $t_k \in \{t_1, t_2 \dots, t_K\}$, each 
contaminated with measurement error. Let variable $X$ measure the objects and $S_t$ be the set of those measurements at time $t_k$
\begin{align}
    S_{t_k} &= \{x^{p},\quad p=1,\dots,P_{t_k} \} \\
    x^{p}_{k} &= X(O_l(t_k)) = O_l(t_k)+\text{msnt\_err}
\end{align}
\underline{\bf Assumption 6:} {msnt\_err} $\sim \mathcal{N}(0,\sigma_{msnt})$, which is a reasonable assumption for mass cytometry data and measurement data in general provided no bias along measurements.
In the following synthetic examples, $\sigma_{msnt}$ can be chosen appropriately small, so that noise doesn't flatten (dominate over) the distribution of samples per time-point. As we later explain, this error is incorporated in the latent variables, though {\bf from this point on, we consider as data, only the measurements} $x^{p}_{n,k}$ ($n$ indicating variable index).

Note: We randomly pick $P$ out of the $L$ in total, on every non-repeated measurement time-point $t_k$. Each measurement destroys the object, so there cannot be $X(O_l(t_{k+1}))$ after $X(O_l(t_{k}))$.
Instead we might measure $X(O_{l+7}(t_{k+1}))$ for instance, where $O_{l+7}$ has not been measured in $\{t_i\}_{i=1}^{k}$. Formally
\begin{align*}
    \{O_{t'}\}\cap \{O_{t''}\}=\emptyset, t'\neq t''
\end{align*}
This does not pose a problem, as we are interested in the distribution of the iid measured objects as a set (distribution at specific time point) and
not in each individual object.
The data we will be considering from this point on, can be written in vector form as:
\begin{equation}
    S = [ x^{(1)}(t_1),\dots, x^{(1)}(t_K), \dots, x^{(P)}(t_1),\dots, x^{(P)}(t_K)]
\end{equation}

\underline{\bf Assumption 7}: The measured objects $x^{(p)}$ are described by stochastic processes, although there is no one-to-one correspondence  
along measurement times. We will use this modelling assumption to formulate the time evolution of 
$S_{t_k}=\{ x^{(p)}(t_k),\dots, x^{(p)}(t_k)\}$ later on and the generalization to $N$ dimensions $S^{N}_{t_k}$.

We can now think of $X$ as a stochastic variable, having a density comprised of $P$ samples, on every time point $t_k$, where these samples come from $P$ {\it realizations} (termed trajectories) of the same underlying SDE. %[rewrite ?].

The are two major categories of temporal data depending whether the same object (variable) is repeatedly measured or not; time series and
time-course (or population) data. For the case of time series data, each object is measured sequentially over
time at predefined sampling time-points. On the contrary, non-repeated measurements or
time-course data, measure a different object at each time instant. This might be the case when
the object is destroyed along the measurement process, as for instance, in mass cytometry
(see discussion above), requiring re-initialization for data acquisition at later times. However,
it is assumed that (under the same experimental conditions) all measured objects are drawn
 from the same unknown distribution.

%==================================================
 \section{Fokker Planck: general form}\label{appendix:FP_general}
 
 The FP equation, for the general stochastic process \Cref{eq:SDE_general}, is a $N+1$ dimensional parabolic PDE given by:
\begin{align}
 \partial_t p(\mathbf{x},t) &= -\sum_{n=1}^N \partial_{x_n} \{a_n(\mathbf{x},t)p(\mathbf{x},t)\} + \frac{1}{2}\sum_{n_1,n_2=1}^{N}\frac{\partial^2}{\partial_{x_{n1}x_{n2}}}\{D_{n_1 n_2}(\mathbf{x},t)p(\mathbf{x},t)\} \\
 &D_{n_1 n_2}(\mathbf{x},t)=\frac{1}{2}\sum_{k=1}^{N'}
 b_{n_1,k}(\mathbf{x},t)b_{n_2,k}(\mathbf{x},t)
\end{align}
with initial condition $p(x,0)=p_0(x)$.
Upon the assumptions on the drift and noise terms (\cref{appendix:SDE}), the special case considered in this manuscript is given by \cref{eq:FP_special}. 
 %==================================================
\section{Weak formulation}\label{appendix:weak_from}

\Cref{eq:weak_form_special} can be written as
\begin{equation}\label{eq:weak_form_innerprod}
\begin{split}
    \Big\langle \phi_m({\bf x},t), \partial_{t}p({\bf x},t) \Big\rangle = -\sum_{n=1}^{N} \Big\langle \phi_m({\bf x},t), \partial_{x_n} \{a_n^T \psi({\bf x},t) p({\bf x},t) \} \Big\rangle &\\
     + \frac{1}{2} \sum_{n=1}^{N} \Big\langle \phi_m({\bf x},t), \partial_{x_{n}x_{n}} \{\sigma^{2}_{n}p({\bf x},t) \} \Big\rangle, \quad m=1,\dots,M &
\end{split}
\end{equation}
with the $\langle f,g \rangle=\int^{T}_{0}f(t)g(t)dt$ denoting the inner product between functions $f$ and $g$ in the $L^2(\mathbb{R}^{N+1})$ function space.

We proceed with integration by parts and after straightforward calculations: 
\begin{equation}\label{eq:ibp_1}
\begin{split}
	&\int_{\mathcal{D}} \phi_m({\bf x},T)p({\bf x},T) d{\bf x} 
	- \int_{\mathcal{D}} \phi_m({\bf x},0)p({\bf x},0)d{\bf x} 
	- \int_{0}^{T}\int_{\mathcal{D}} \partial_t\{\phi_m({\bf x},t)\} p({\bf x},t) d{\bf x}dt= \\
	\sum_{n=1}^{N} \sum_{q=1}^{Q} & a_{nq} \int_{0}^{T} \int_{\mathcal{D}} \partial x_n \{\phi_m({\bf x},t)\} \psi_q({\bf x}) p({\bf x},t) d{\bf x}dt 
	+ \frac{1}{2} \sum_{n=1}^{N} \int_{0}^{T} \int_{\mathcal{D}} \sigma_{n}^2 \partial_{x_{n} x_{n}} \{\phi_m({\bf x},t)\} p({\bf x},t) d{\bf x}dt
\end{split}	
\end{equation}
where we assume that the distribution is zero on the boundary $p({\bf x},t)|_{\partial D}=0$ for all $t$.
It is apparent that by applying the weak formulation, the derivatives are ''shifted`` to the test functions  $\phi_m$, which is one major advantage of our modeling approach.

As one can see, the integrals are $N-$dimensional as the variables $\{X^n\}_{n=1}^N$ are coupled in general.
In the following, we define these spatio-temporal functions as:
\begin{equation}\label{eq:test_funcs}
 \phi_m({\bf x},t) := \bar{\phi}_{m_1}({\bf x}) \tilde{\phi}_{m_2}(t)
\end{equation}
meaning that space $\bar{\phi}_{m_1}$ and time $\tilde{\phi}_{m_2}$ have different functional forms and $m_1\in\{1,\dots, M_1\}$, $m_2\in\{1,\dots, M_2\}$, $M_1 M_2=M$. 
We choose test functions in a way that their form is 
able to capture the data heterogeneity, form and time-scales as we discuss later on. Moreover this is why their functional form are
different in principle, though we could use the same family of functionals.

On top of the choice of \cref{eq:test_funcs}, we proceed with the simplification regarding the (de)coupling of the spatial variables:
\begin{equation}\label{eq:test_func_pairwise}
 \bar{\phi}_{m_1}({\bf x}) \approx \bar{\phi}_{m_1}(x_1)\dots \bar{\phi}_{m_1}(x_N)
\end{equation}
though the variable {\bf interactions are incorporated in the dictionary atoms} because the drift term ${\mathbf a_n} \Psi(X)$ term constitutes of 
$Q \geq N$ components (for a linear dictionary $\Psi(X(t))=X(t)$ and $Q=N$).\\
\underline{\bf Superposition assumption:} 
We assume that the unknown complex distribution $p({\mathbf x}, t)$ can be written as a linear combination of simple, ``appropriately chosen", spatio-temporal test functions:
\begin{equation}\label{eq:sup_assump}
    p({\mathbf x}, t) = \sum_{m_1} \sum_{m_2}  d_{m_1,m_2} \tilde{\phi}_{m_2}(t) \prod_n  \bar{\phi}_{m_1}(x_n) 
\end{equation}

\medskip
We proceed with \Cref{eq:ibp_1}, for  $n\in \{1,\dots, N\}$ variables in total, so for the $n$-th component and $m_1$-th, $m_2$-th test functions we get:
\begin{equation}\label{eq:ibp_2}
\begin{split}
	&\int_{\mathcal{D}_n} \bar{\phi}_{m_1}(x_n)\tilde{\phi}_{m_2}(T)p(x_n,T) dx_n 
	- \int_{\mathcal{D}_n} \bar{\phi}_{m_1}(x_n)\tilde{\phi}_{m_2}(0)p({x_n},0)dx_n 
	- \int_{0}^{T}\int_{\mathcal{D}_n} \bar{\phi}_{m_1}(x_n)\partial_t\tilde{\phi}_{m_2}(t) p(x_n,t) dx_n dt \\
	&= \sum_{q=1}^{Q} a_{nq} \int_{0}^{T} \int_{\mathcal{D}} \partial x_n \big(\bar{\phi}_{m_1}(x_n)\big)\tilde{\phi}_{m_2}(t) \psi_q({\bf x}) p({\bf x},t) d{\bf x} dt 
	+ \frac{\sigma_{n}^2}{2}  \int_{0}^{T} \int_{\mathcal{D}_n}  \partial_{x_{n} x_{n}} \big(\bar{\phi}_{m_1}(x_n,t)\big)\tilde{\phi}_{m_2}(t) p(x_n,t) dx_n dt
\end{split}	
\end{equation}
where we have $N\times M$ integral equations in total. Note that the first two integrals in the l.h.s. depend on the population data of $X(t_1)$ and $X(t_K)$ so we consider them as constants $C_{m_1, m_2,n}$ computed once.
 %-----------------------------------------------------------
 \section{Integral Estimators}\label{appendix:estimators}
 Suppose that we fix time $t=\tau$ and let $x_i$ be drawn from $p(x,\tau)$, as is the case with
population data. Then for any function $f(x):\mathbb{R}^N \mapsto \mathbb{R}$ we approximate:
\begin{equation}\label{eq:naive_MC_integration}
    \mathbb{E}_{p(x;\tau)}[f(x)]=\int_{\mathcal{D}} f(x)p(x;\tau)dx \simeq \frac{1}{P} \sum_{i=1}^{P} f(x_i)
\end{equation}
where $P$  is the number of samples considered. Also, we approximate the time integral on the discretized time domain $\{t_1<t_2<\dots\leq t_K=T\}$ using the standard numerical analysis trapezoidal rule on $f(x,t): \mathbb{R}^N\times\mathbb{R}^+ \mapsto \mathbb{R}$:
\begin{align}\label{eq:naive_MC_integration2}
 \mathbb{E}_{p(x,t)}[f(x,t)] &= \int_{0}^{T} \int_{\mathcal{D}} f(x,t)p(x,t)dxdt \nonumber \\
 &\simeq \frac{1}{2}\sum_{k=1}^{K-1}
 \Big(\int_{\mathcal{D}} f(x,t_{k+1})p(x,t_{k+1})dx + \int_{\mathcal{D}} f(x,t_k)p(x,t_k)dx \Big)(t_{k+1}-t_k)
\end{align}

 %-------------------------------------------------------
 \section{Minimization problem}\label{appendix:minimization_problem}
 In the ideal case where the number of rows (test functions) $M$ is equal to the number of columns (dictionary atoms) $Q$ in matrix $\Psi$, one could possibly solve the system directly and recover the correct coefficients provided that the noise in the data has not affected the condition of matrix $\Psi$. 
In principle though, the resulting systems are over-determined because we use a large number of test functions and a broad family of potentially useful dictionary atoms, and a least-squares type minimization of the form $\min ||Z-\Psi a||$ over $a$, does not have a sparse solution but a full (dense) one. Formally
\begin{equation}\label{eq:ord_LS}
    \underset{a}{\min}|| Z-\Psi a||^{2}_{2},\quad \text{with solution}  \quad \hat{a}=(\Psi^T \Psi)^{-1}\Psi^TZ  %\raggedright{(LS)}
\end{equation}
In practice, a dense solution means that there is a high correlation of the available candidate features (columns of $\Psi$) as a result of noise in the data (ill-conditioned matrix $\Psi$) and sparse regression is preferable. To achieve this, a minimization problem with penalization is solved instead:
\begin{equation}\label{eq:SSR_L0appendix}
    \underset{a}{\min} ||a||_0 \quad \text{subject to} \quad {||Z-\Psi a ||_2 }\leq \epsilon, 
\end{equation}
which uses the $L^0$ norm over the minimization, where $||a||_0$ is the number of non-zero elements in $a$ and $\epsilon$ is the regression error. This is a non-linear, non-convex optimization problem (NP-hard) and the solution space grows exponentially with the size of $a$ making it computationally intractable. A usual approach is to relax the norm in the objective function and solve the convex $L^1$ least-squares minimization problem or 
the $L^1-$regularized form:
\begin{equation}\label{eq:SSR_L1reg}
    \underset{a}{\min} (\frac{1}{2}||Z-\Psi a ||^{2}_{2} +   \lambda||a||_1) , 
\end{equation}
where sparsity is enforced by the $L^1$ norm and controlled by the positive Lagrange multiplier $\lambda$ \cite{Schaeffer_2017}. In other words, $\lambda$ balances complexity and sparsity of the solution and an approximation algorithm solving \cref{eq:SSR_L1reg} is LASSO \cite{Tibshirani96}. 
See \cite{PLOS_reactionet_Lasso} for a two step modified Lasso suited for high dimensional, multiscale stochastic (unary/binary) reaction networks using time series data having intrinsic uneven noise.
Our modelling assumption is that the underlying dynamics are governed by a few terms, which in mathematical terms translates in sparsity of the solution of \cref{eq:linearSLE}.

We note that despite its computational advantages, the least-squares approach in system identification usually produces a ``full" solution where each component of %$\mathbf{a},\mathbf{\alpha}$ $\bf{a,\alpha}$ 
$a$ is nonzero despite the true structure being sparse, making inference algorithms sensitive to noise and under-sampling. A computationally demanding alternative though robust against noise and outliers (avoiding over-sparsity) is the Entropic Regression system identification \cite{ER_Bollt_18}. We demonstrate robustness of our proposed algorithm against outliers in the supplementary material provided.
One can combine the least-squares formulation with thresholding (either iterative \cite{SINDy_2016,Clementi_JCP18} or not, hard \cite{SINDy_2016,Hybrid_SINDy} or soft \cite{Clementi_JCP18}) and conclude to a desirable, sparse solution.

Hyper-parameter tuning is important, because an excessively sparse solution will under-fit the data whereas a nearly full solution will over-fit. The chosen criterion for OMP parameter tuning is the Bayesian Information Criterion (BIC) penalizing model complexity (more terms in the solution) and specifics can be found in the Supplementary.
We remark that the sparse optimization problem penalizes the magnitude of the coefficients directly, so in order to avoid inclusion/exclusion of a feature whose vector magnitude differs substantially from the others, we normalize each column $\psi_q$ of $\Psi$ by its $L^2-$norm. 
%[{\blue Mention hard thresholding?? }]

 %-----------------------------------------------------------
\section{System interventions: additional datasets from variable perturbations}
\label{sec:intervention}
It is usually the case that multiple datasets of the {\it same underlying mechanism} are given as input. 
A significant attribute of the USDL methodology is the framework of handling additional data, of the same set of variables and sampling times.
Let us assume for illustration purposes that we simulate a known differential equation multiple times while perturbing its variables and generate datasets, each containing many trajectories.
These datasets can be categorized in two types: i) \underline{activation:} starting the simulation from a different initial configuration, eventually evolving to a steady or equilibrium state, and
ii) \underline{inhibition}: neutralization of the effect of a specific variable over time.

Exploration of the phase space, through additional data, results in rich dynamical information as input to the inference problem, narrowing down the possible solutions to the minimization problem discussed in \cref{sec:min_prob}. 
In other words, additional data make the unknown system more identifiable.
On the contrary, as the number of state variables increases, very large sample sizes are required for successful inference,
which for biological laboratory experiment measurements cannot be available, though experimental perturbations can be performed instead. These type of variable perturbations can be incorporated through interventions \cite{USDL_bioinformatics}.
Collectively, i) and ii) constitute interventions on the underlying system.
Each different activation intervention $r$, results to a new SLE of the form of \Cref{eq:linearSLE} (remark that we omit species index $n$) such that:
\begin{equation}
    Z^{(r)} = \Psi^{(r)} a^{(r)}
\end{equation}
where the unknown $a^{(r)}$ vector should have been determined independently. On the contrary, we merge the weak space projected data into a larger SLE, so in matrix form we end up with \cref{eq:Matrix_Activations}:
\begin{figure}[h]\label{eq:Matrix_Activations}
\centering
   $\begin{bmatrix} 
            Z^{(1)}   \\
            Z^{(2)}     \\
            \vdots  \\
            Z^{(R)} 
    \end{bmatrix}$=
    $\begin{bmatrix} 
            \Psi^{(1)}   \\
            \Psi^{(2)}     \\
            \vdots  \\
            \Psi^{(R)} 
    \end{bmatrix}$
    $a$
    \caption{$R$ activation interventions in matrix form (for the $n$-th variable)}
\end{figure}
where $a$ is determined over all activations with a single call of the feature selection algorithm.
In the examples in \Cref{sec:experiments} we used activations, though we refer the reader to the supplementary material for the derivation of inhibitions for the SDE setup, similar to the ODE anzatz \cite{USDL_bioinformatics}.

%---------------------------------------------
\section{Algorithm limitations}
1. Might be slow for variables having wide range being captured by spatial test functions, though current matlab version is parallel and scales well. \\
2. Partial parameter setting based on fitting to average trajectories of the input data. Data driven test functions and automatic algorithm hyper-parameter estimation ($M_1, M_2$, feature selection stopping criteria) is under development. \\
3. Requires relatively short sampling times (w.r.t. the fastest dynamics) for adequate time-derivative component estimation [not particularly short, as short as a finite difference scheme would require]. Tackled with data resimulation (constrained smoothing or collocation method). \\
4. Requires Gaussian constant-variance noise. Tackled with resimulation of the data. \\
5. Depending on the identifiability or complexity of the system, it might require multiple interventions (activations/inhibitions) in order to gain insight from richer dynamics.

%==================================================

 \section{Notation}
\begin{table}[h]
        table containing common acronyms, terms and mathematical notation \\
    \centering
    \begin{tabular}{l l} % indent both cols on the left
        \hline
        FP & Fokker Planck equation \\
        SDE & Stochastic Differential Equation \\
        PDL & Population Dynamics Learning\\
        $X^n_t$ & $n$-th variable which is a stochastic process\\
        $X_t$ & vector of variables (stochastic processes) which follow SDE's \\
        $X(t)$ & set of data points at time $t$ of vector of variables $X_t$\\
        $\mathbb{P}(X_t,t)$ & unknown probability density function describing evolution of variables $X_t$  \\
        $p({\mathbf x},t)$ & value of the pdf for all data ${\mathbf x}$ at time $t$ that is $X(t)$\\
        $P$ & number of data points measured per time point \\
        $S$ &  set of all data points over all time points (possibly over multiple interventions)\\
        $Q$ & number of dictionary atoms \\
        $\psi(x)$ & vector of symbolic functions of dictionary items \\
        $\psi_q(x)$ &  $q$-th dictionary item \\
        $\Psi$ & data matrix over dictionary items (all timepoints) \\
         $\phi_m(x,t)$ & $m$-th test function (constructed feature)\\
        $\bar{\phi}(x)$ & spatial test function \\
        $\tilde{\phi}(t)$ & temporal test function \\
        $M$ & number of test functions in total\\
        $Z$ & matrix containing time derivatives projected to Weak Space\\
        $\tilde{\Psi}$ & matrix containing projected dictionary items to Weak Space \\
        $\tilde{\Psi}_n$ & $n$-th column of matrix $\Psi$ for $n$-th variable \\
        $W$ & matrix containing projected diffusion term to Weak Space \\
        $A$ & unknown matrix of constants defining coefficients of the solution \\
        $\bf a_n$ & vector of $n$-th row of $A$ defining dictionary coefficients of that variable\\
        $\Sigma$ & (diagonal) covariance matrix of diffusion coefficients of the solution \\
        $\mathcal{D}$ & subspace of $\mathbb{R}^N$ where data are measured \\
        OMP & Orthogonal Matching Pursuit learning algorithm \\
%        $\theta$ & OMP threshold parameter stopping criterion (imposing sparsity)\\
%        $\Theta$ & Set of candidate threshold parameters\\
%        $K$ & max features OMP stopping criterion\\
        $rr$ & relative error a posteriori metric\\
        \hline
    \end{tabular}
    \caption{ }
    \label{tab:notation}
\end{table}

\section*{Acknowledgments}
We would like to thank Georgios Papoutsoglou for his helpful comments. This  work  was  funded  by  the  European  Research  Council  under  the  European  Union’s Seventh Framework Programme (FP/2007-2013) / ERC Grant Agreement n.  617393.

\bibliographystyle{siamplain}
\bibliography{references}

\end{document}

\maketitle

\section{Learning the dynamics of an 8 protein network.}

\begin{figure}[ht]
\centering
	\includegraphics[width=0.17\paperwidth]{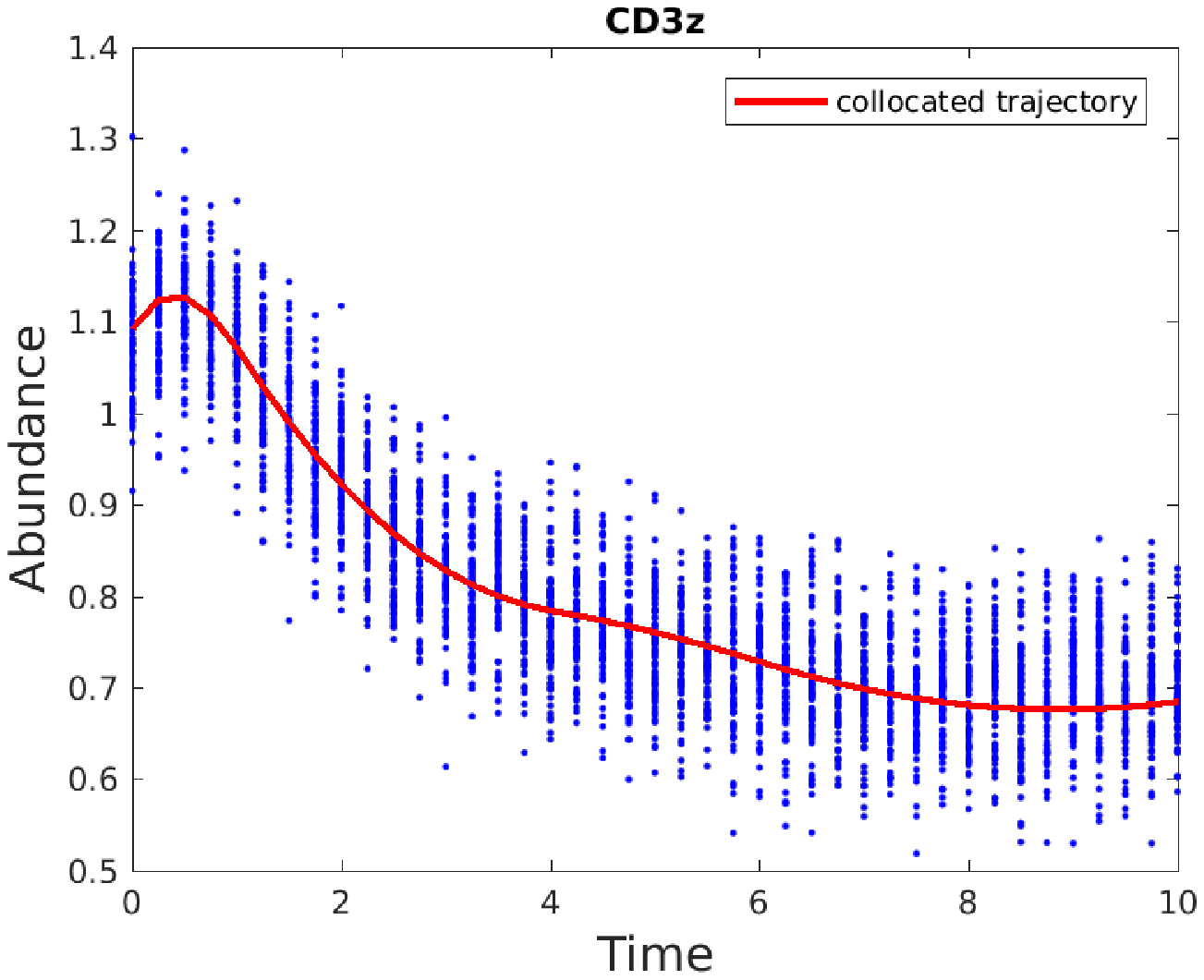}
    \includegraphics[width=0.17\paperwidth]{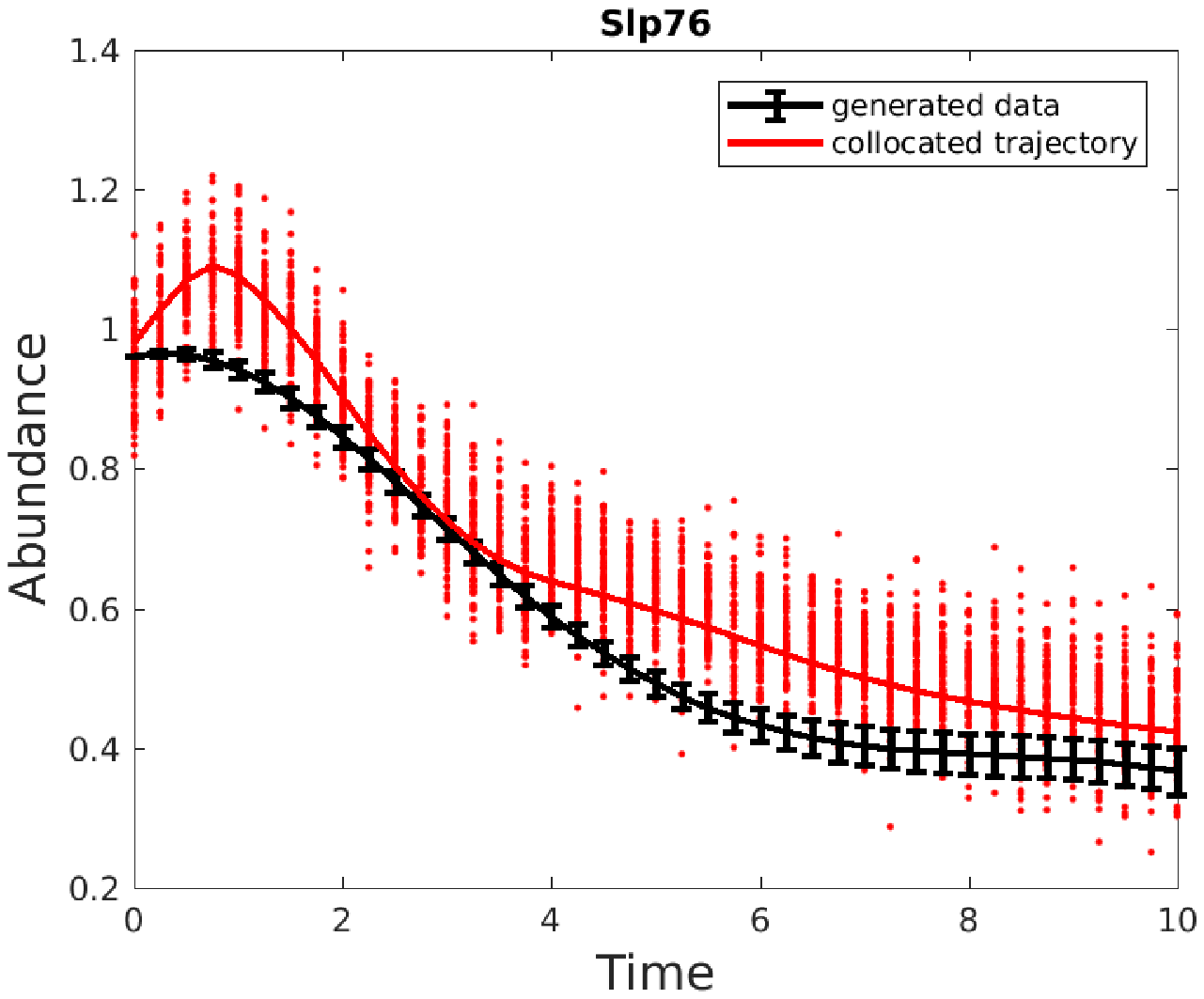}
    \includegraphics[width=0.17\paperwidth]{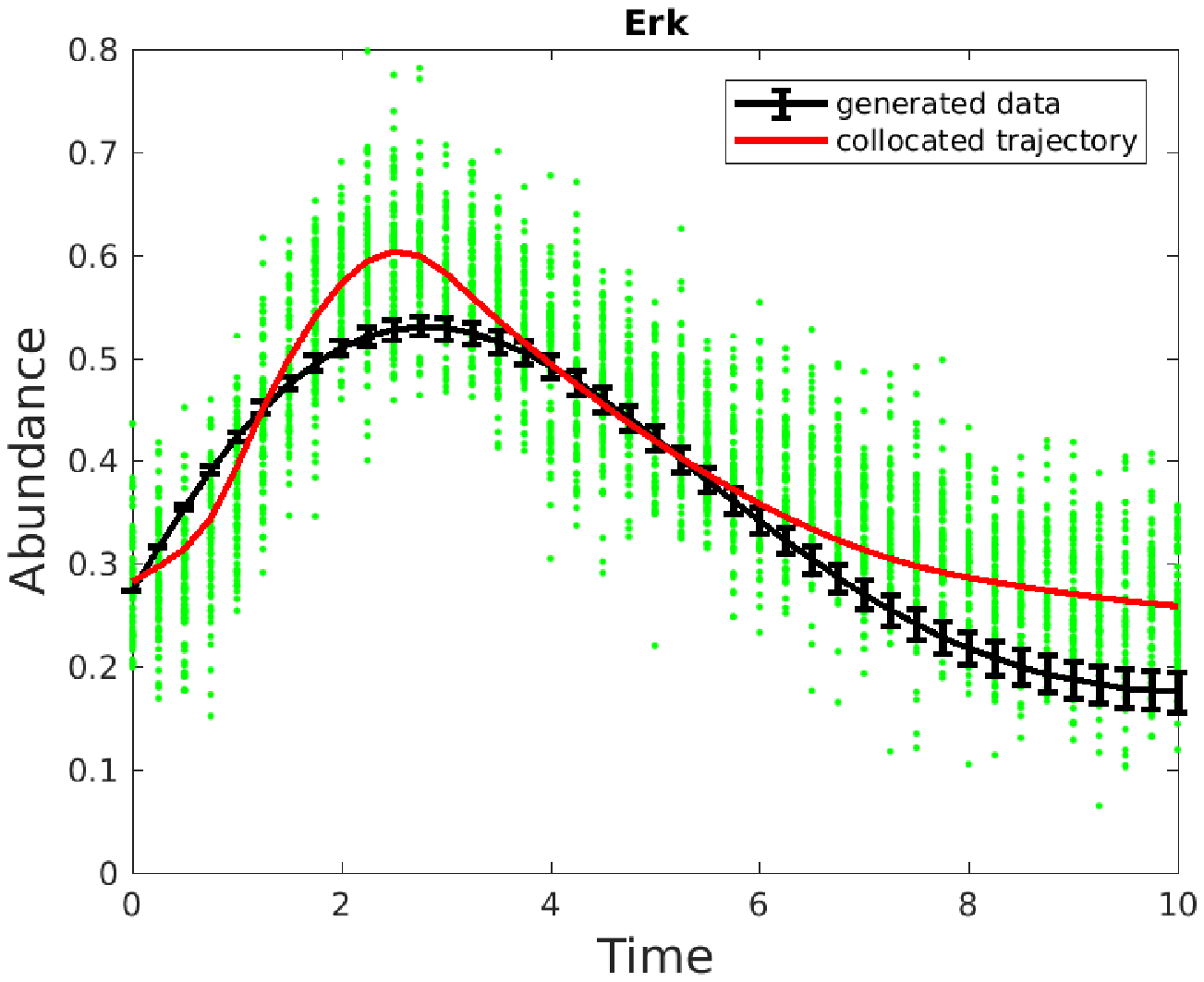}
    \includegraphics[width=0.17\paperwidth]{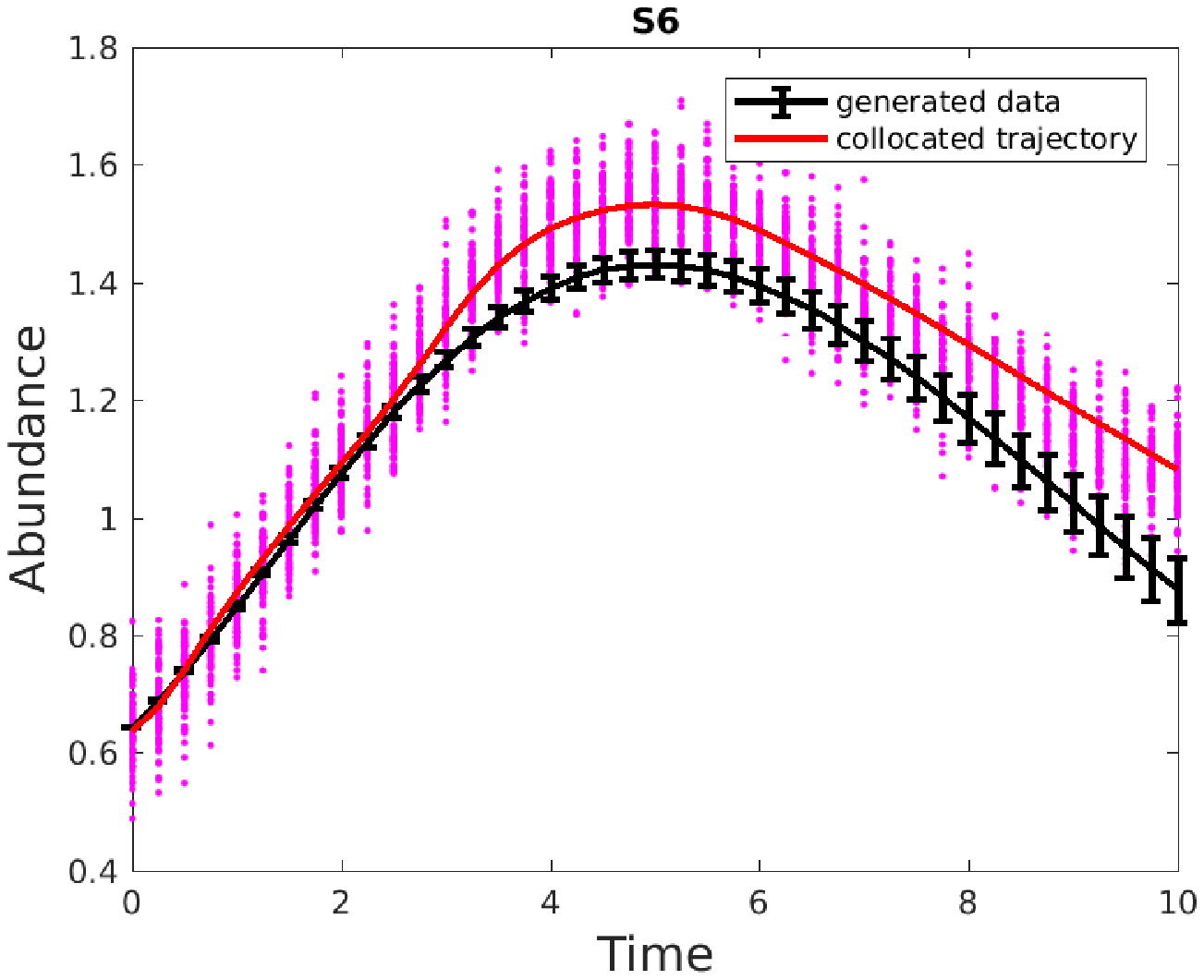}
    \includegraphics[width=0.17\paperwidth]{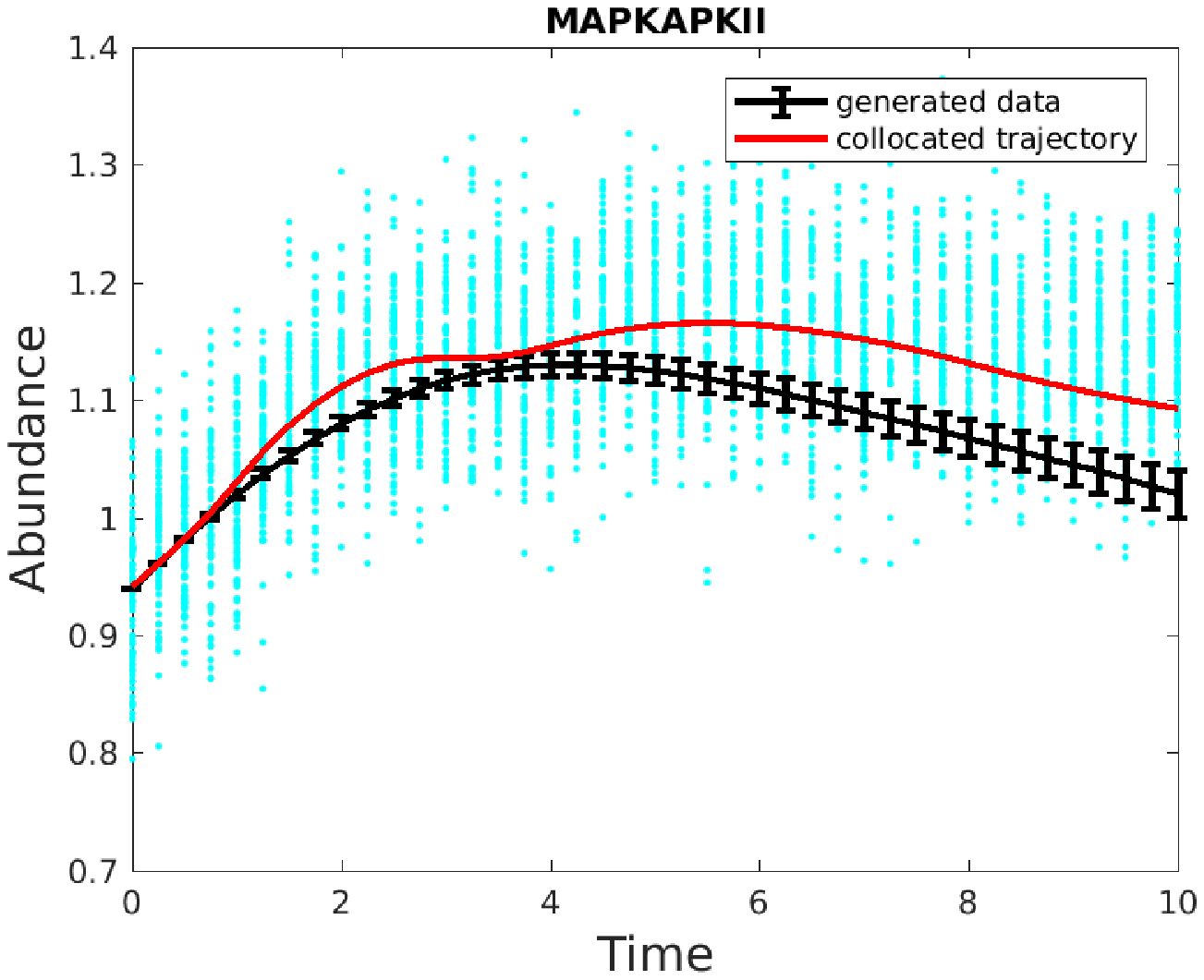}
    \includegraphics[width=0.17\paperwidth]{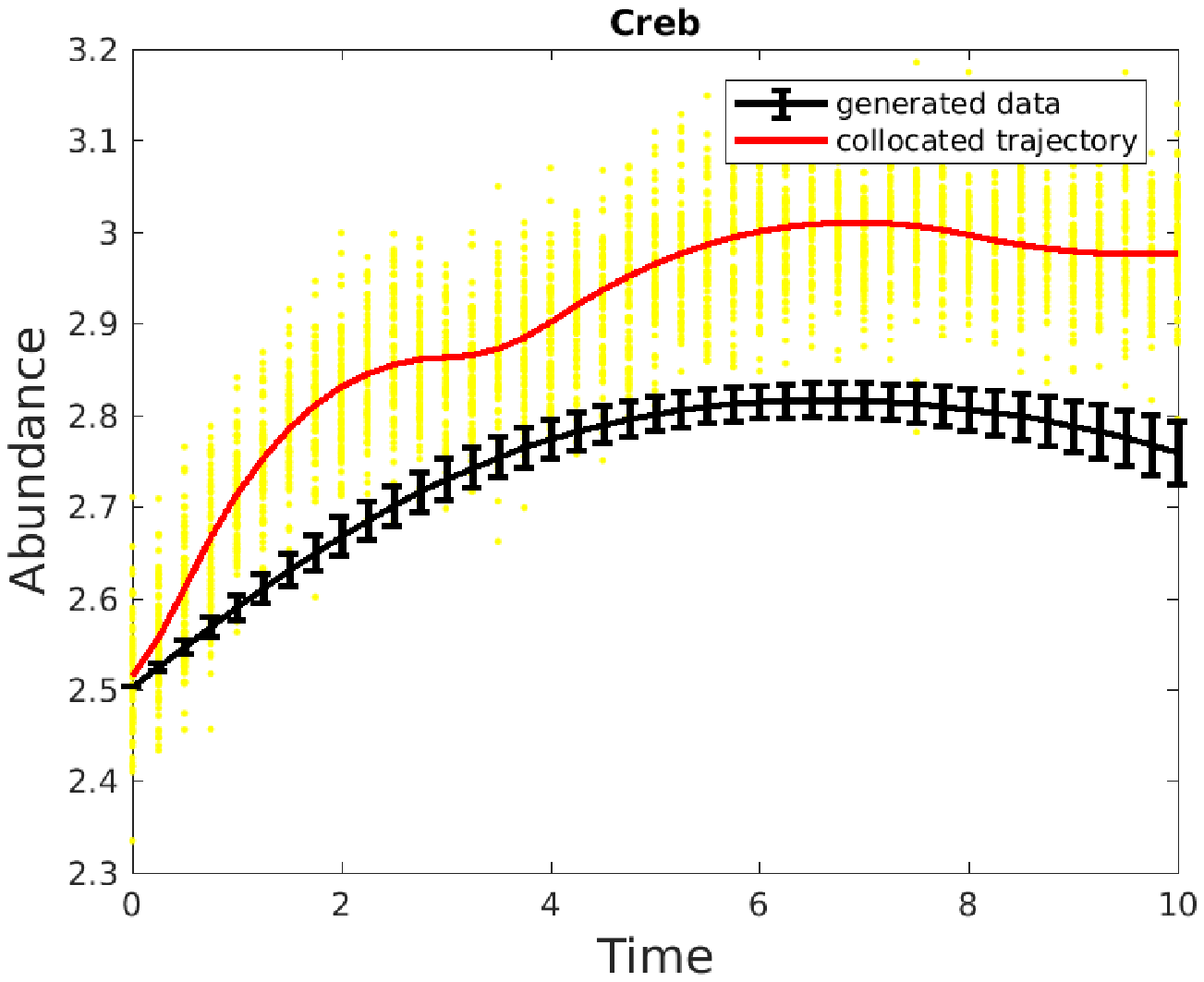}
    \includegraphics[width=0.17\paperwidth]{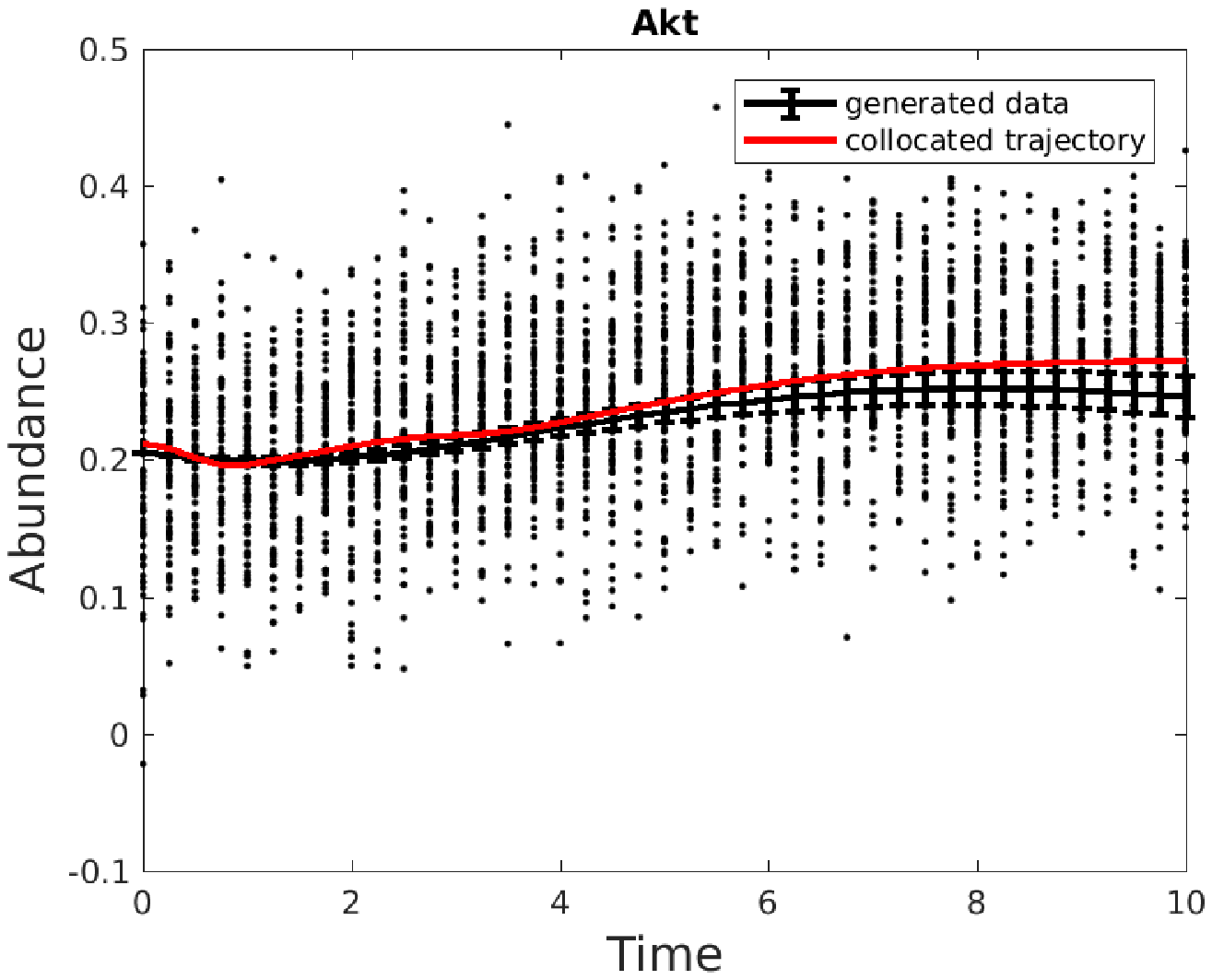}
    \includegraphics[width=0.17\paperwidth]{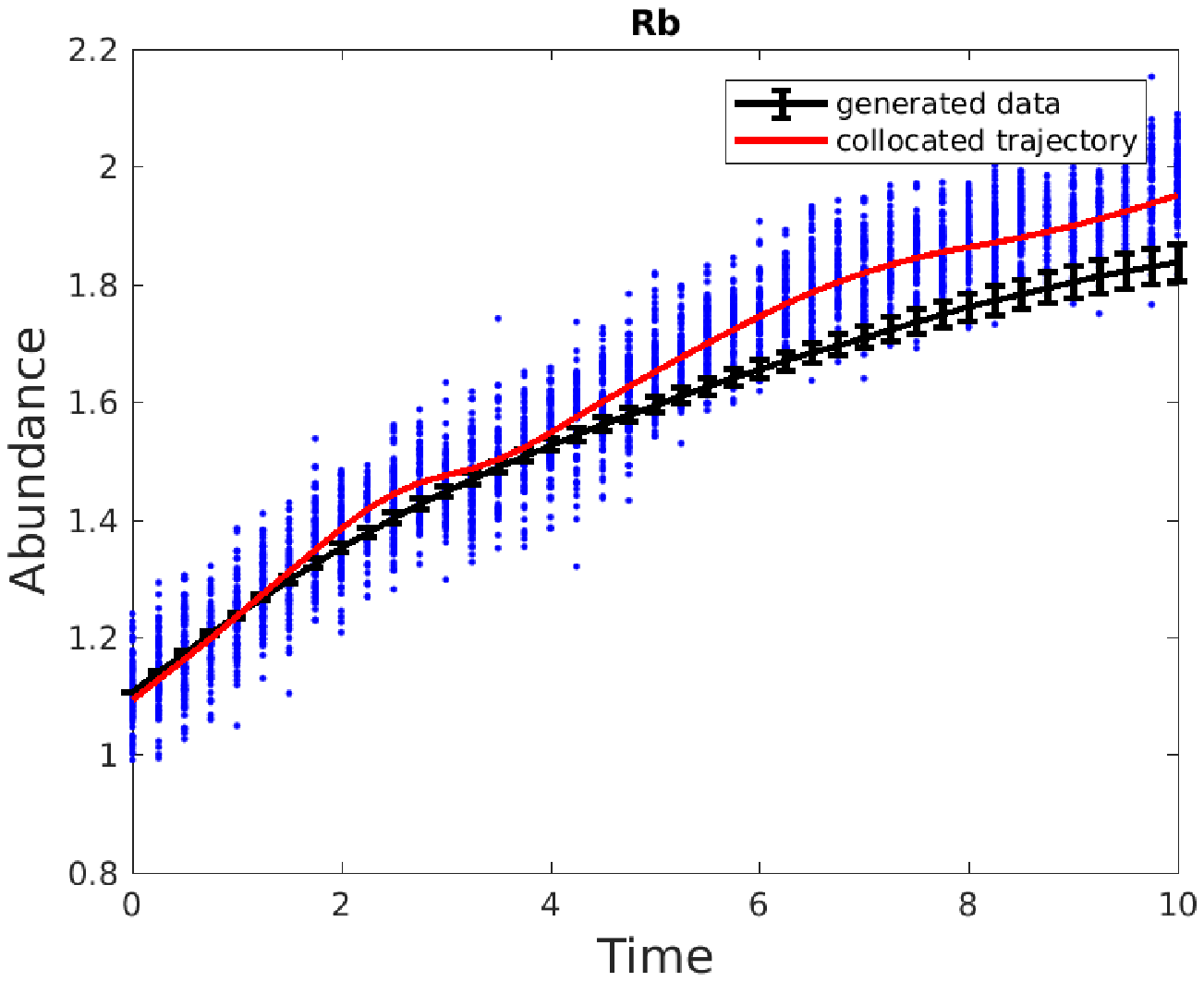}
    \begin{tabular}{|c ||c | c | c | c |c | c | c | c |}
\hline
protein & CD3z &  SLp76 &  Erk & S6 & MAPKAPKII &  Creb &  Akt & Rb \\
\hline
$||X_{coll}-X_{gen}||_2$ & {\blue 0}  & 0.13 &  0.13 & 0.09 & 0.04   & 0.05 &  0.07 & 0.04     \\
\hline
\end{tabular}
    \caption{Protein network of eight interacting proteins. % (CD4+, $CD3\_CD28$).
    Population data (dots), collocated trajectories (red). Upon inference of $\hat{A}$ over 25 iid runs on subsampling of the mass cytometry data, we generated trajectories with errorbars  (black).
    The first protein trajectory CD3z is given as prior knowledge, as explained in the main text. 
    A linear dictionary is used for this train set % [B-spline supp=0.25 ($M_1=16$), cloud variance =0.24, $M_2$=7, threshold=0.001, NoI=25 iid runs] 
    and inference over this bigger network is successful. The $L_2$ distance from the collocated trajectories is shown in the table for each protein. Compare with KEGG Tcell graph}
	\label{fig:DREMI_8}
\end{figure}

\begin{figure}[ht]
\centering
	\includegraphics[width=0.17\paperwidth]{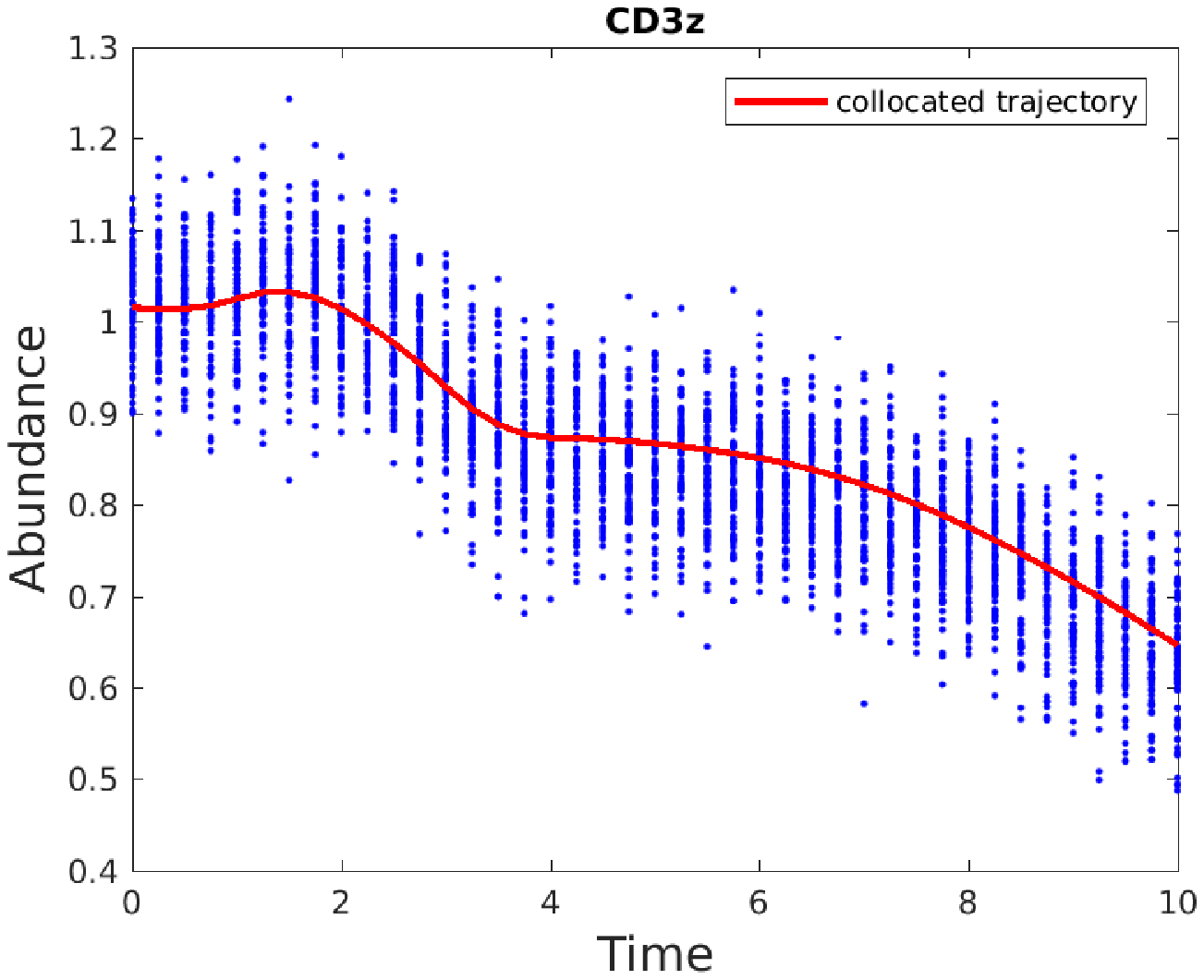}
    \includegraphics[width=0.17\paperwidth]{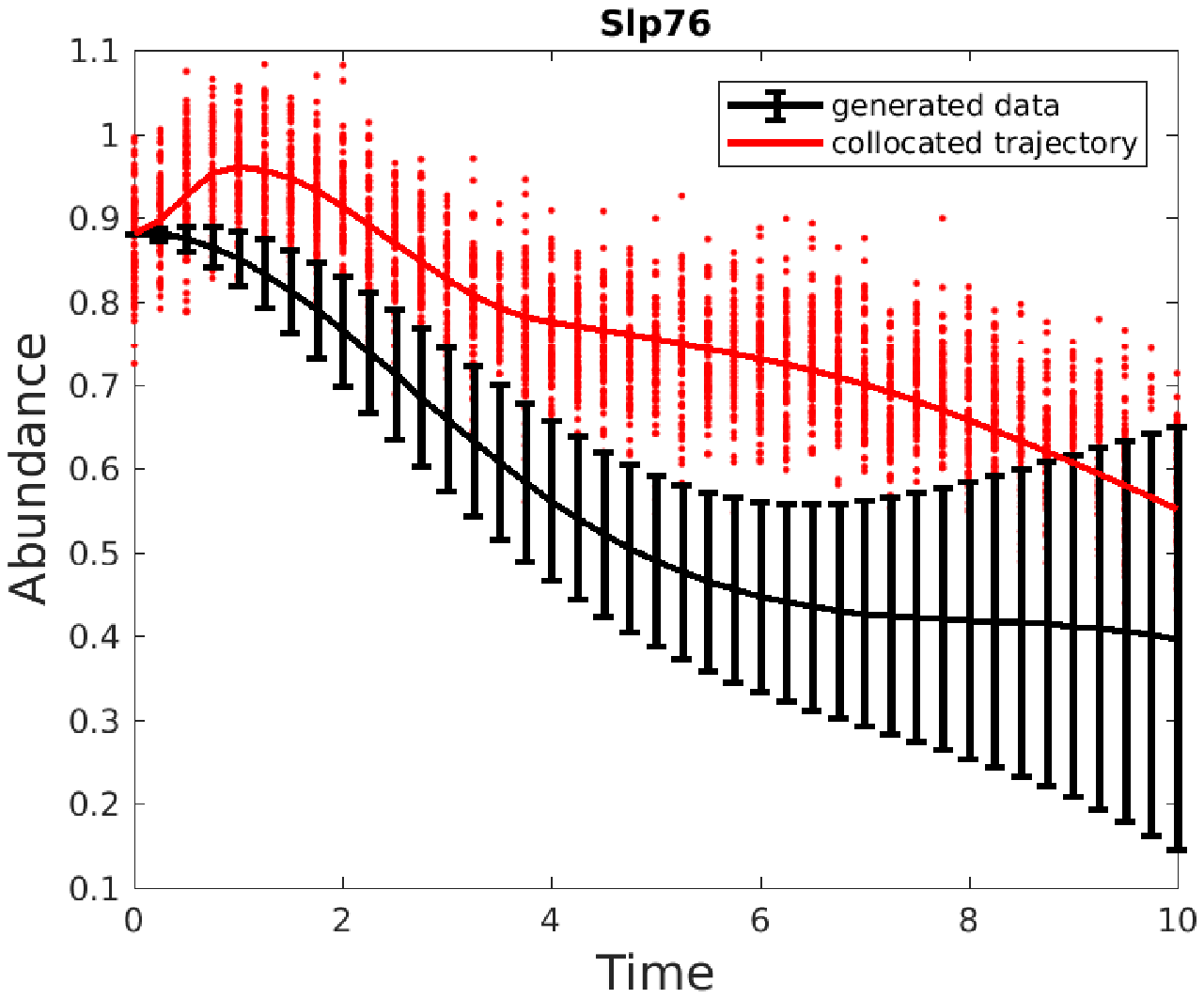}
    \includegraphics[width=0.17\paperwidth]{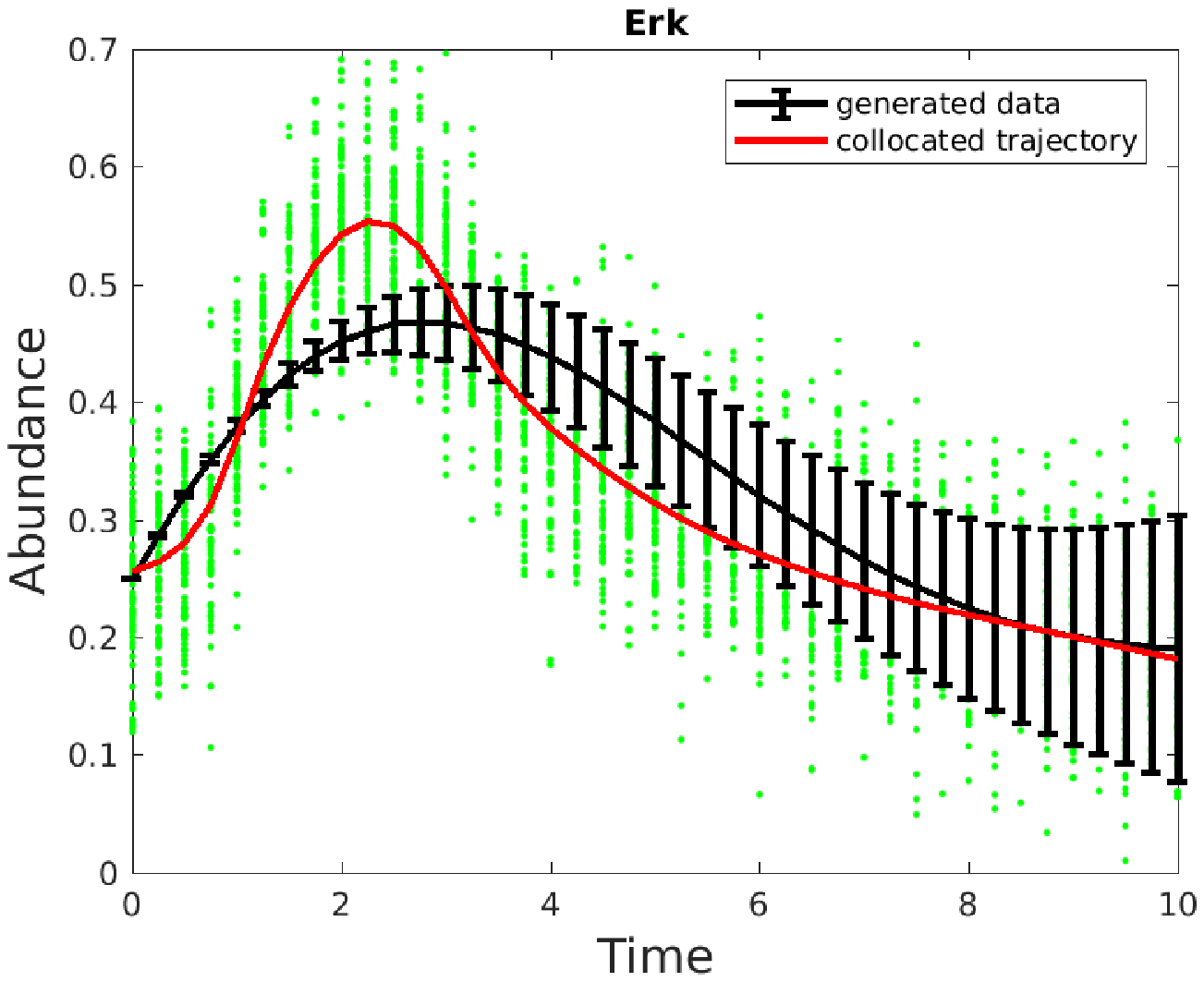}
    \includegraphics[width=0.17\paperwidth]{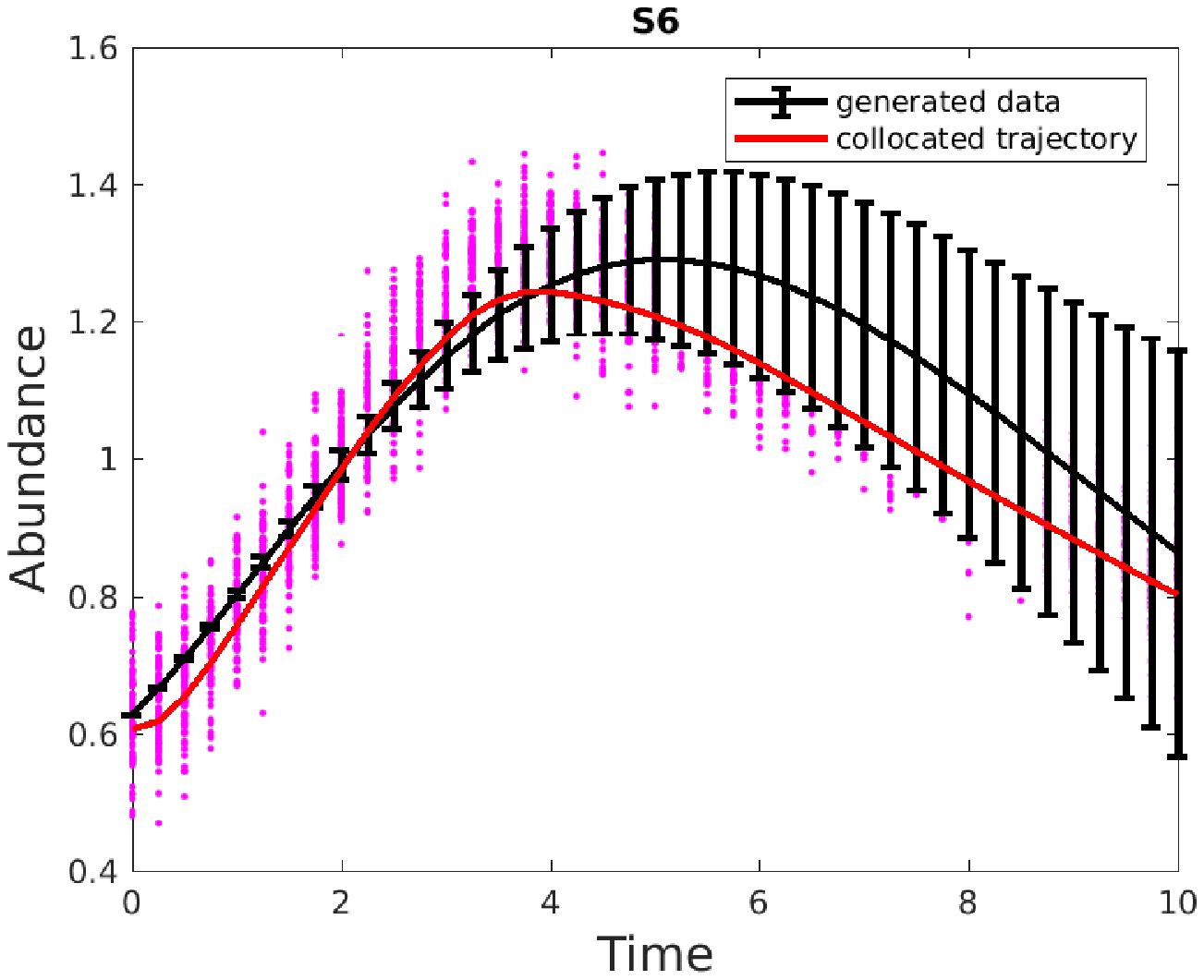}
    \includegraphics[width=0.17\paperwidth]{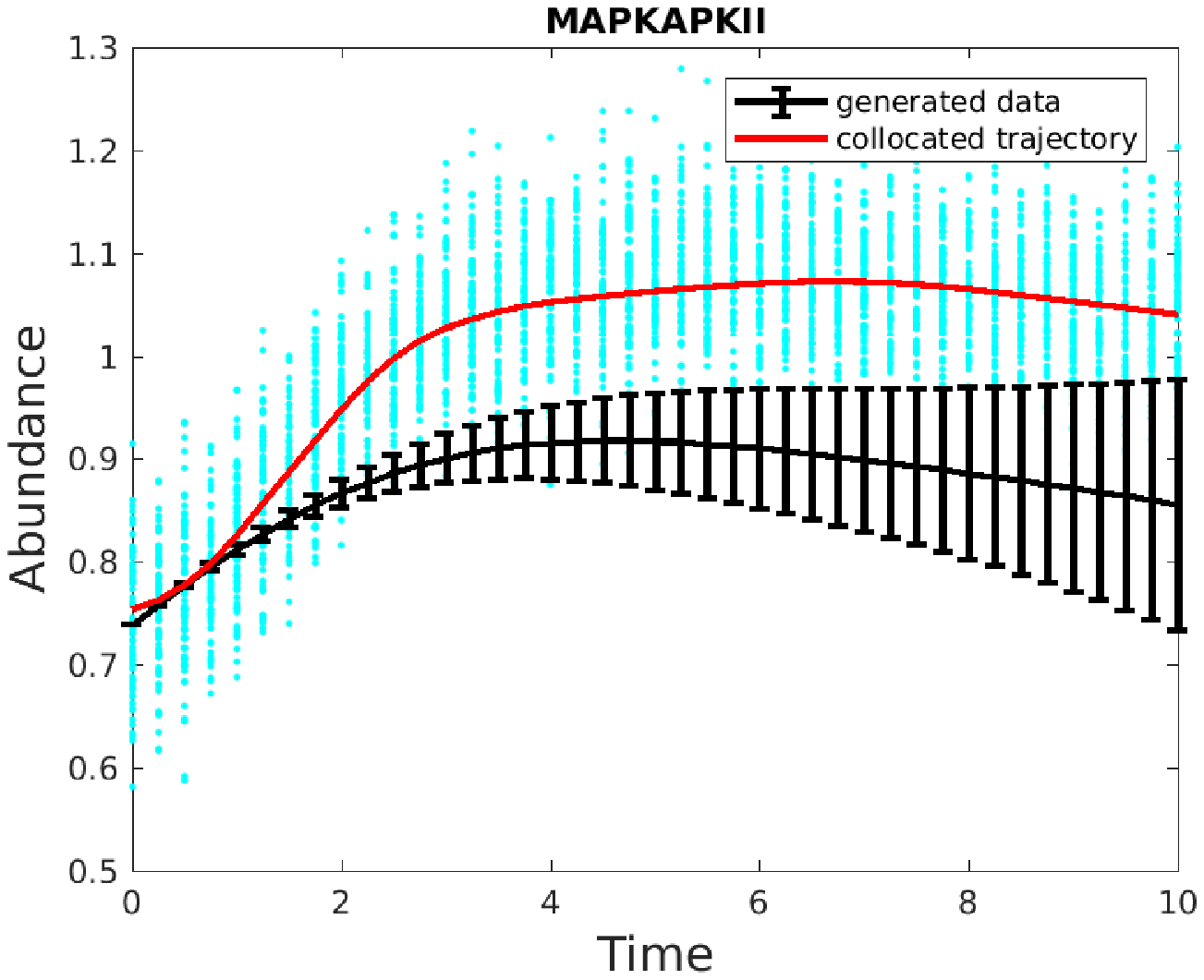}
    \includegraphics[width=0.17\paperwidth]{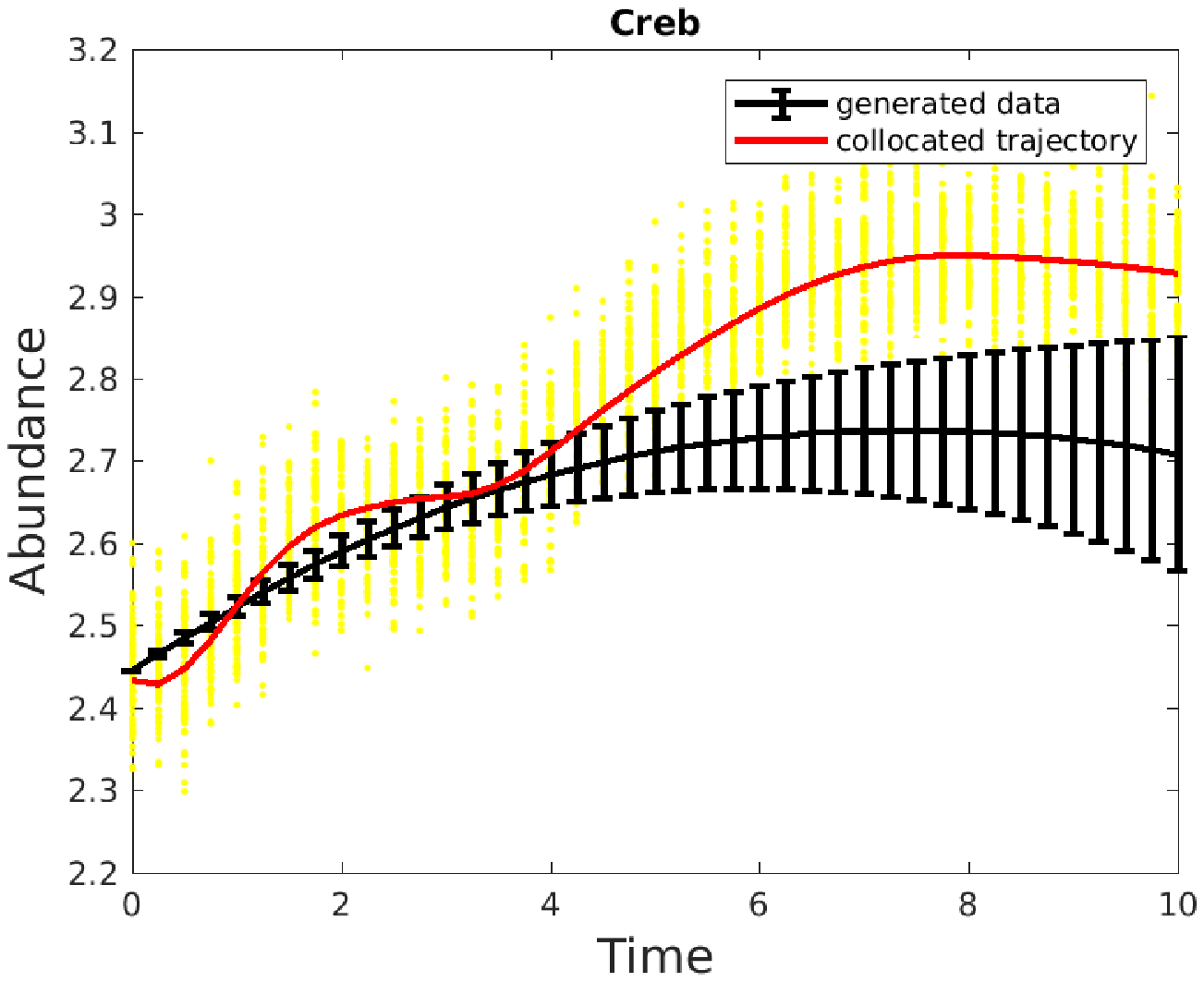}
    \includegraphics[width=0.17\paperwidth]{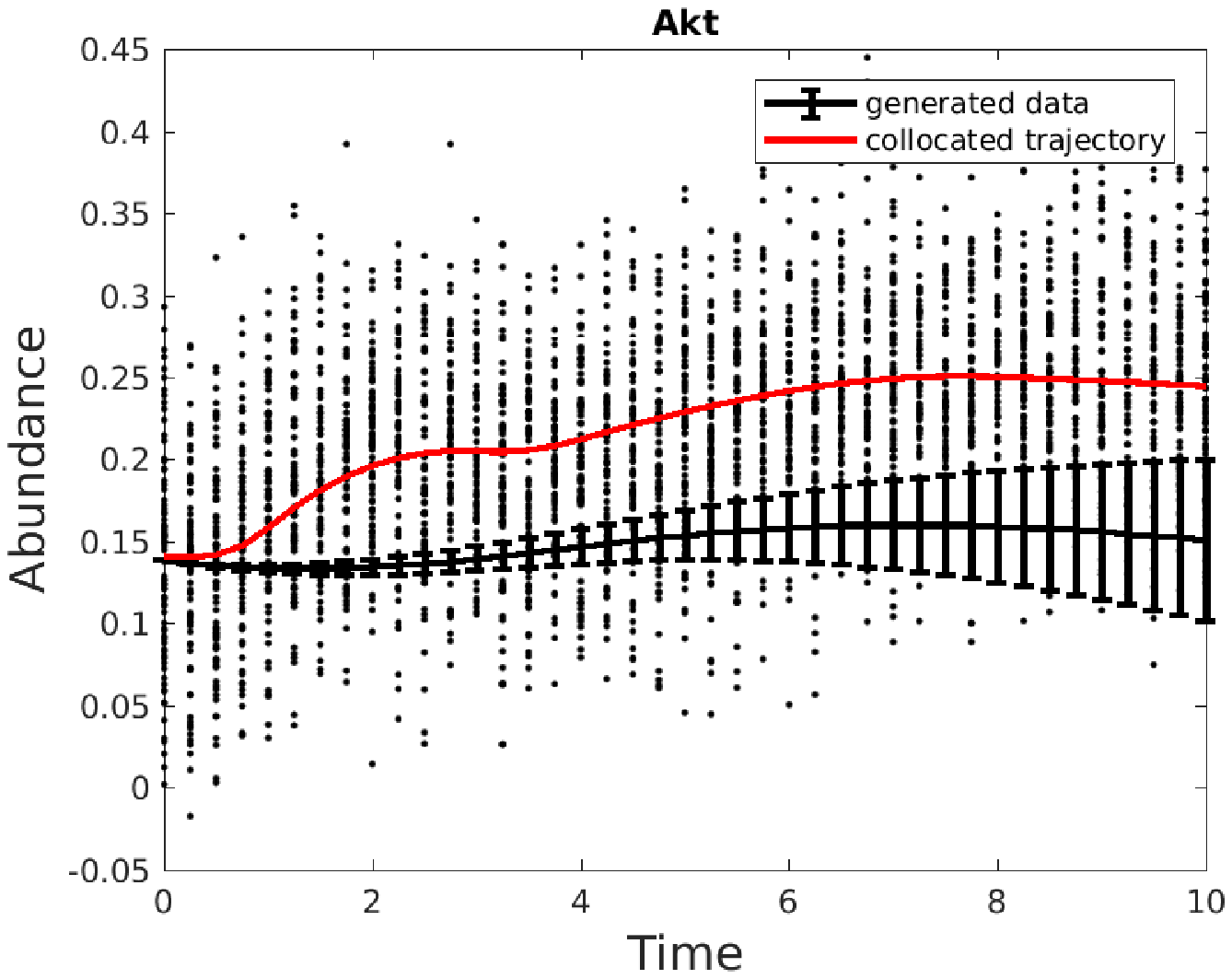}
    \includegraphics[width=0.17\paperwidth]{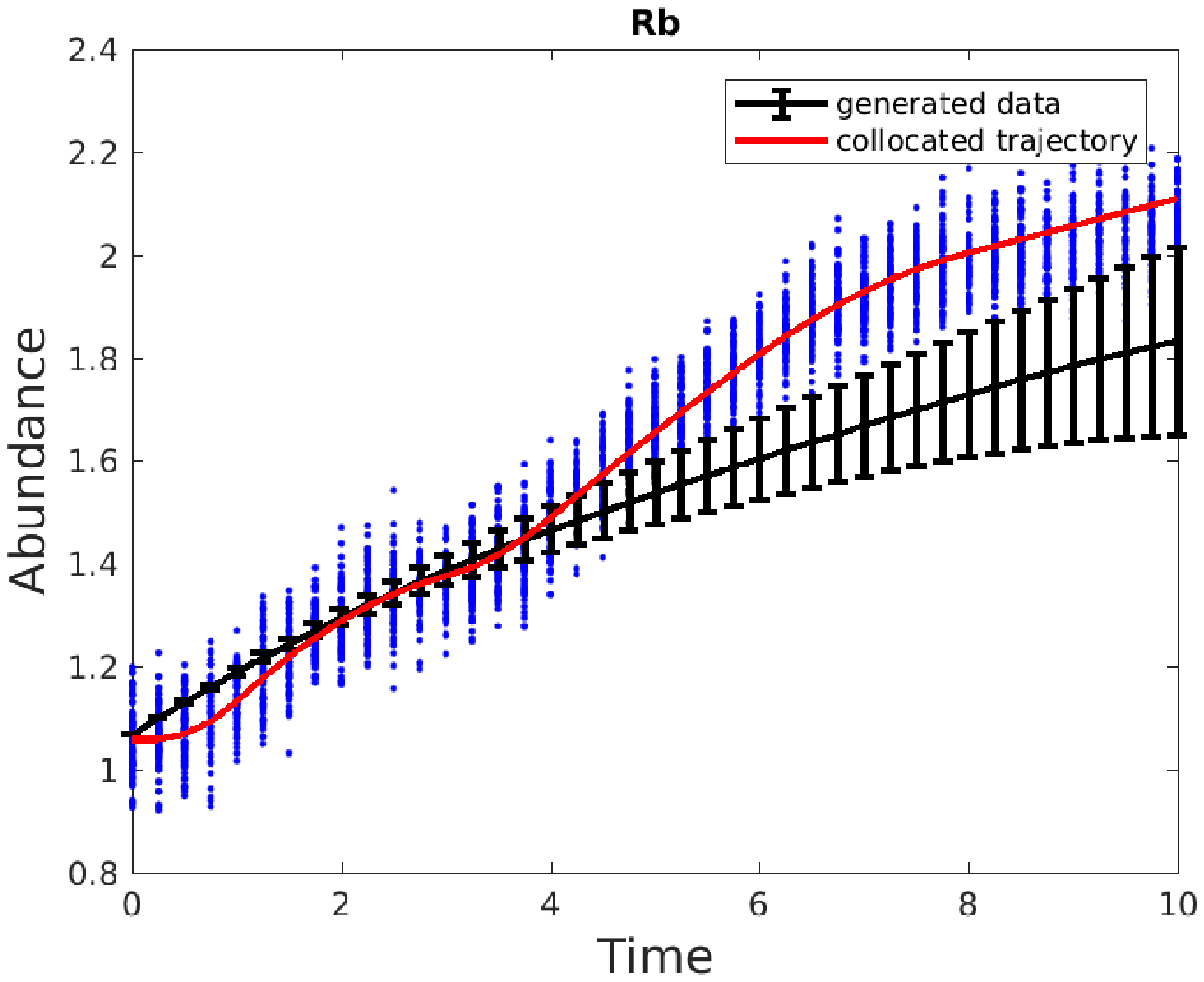}
    \begin{tabular}{|c ||c | c | c | c |c | c | c | c |}
\hline
protein & CD3z &  SLp76 &  Erk & S6 & MAPKAPKII &  Creb &  Akt & Rb \\
\hline
$||X_{coll}-X_{gen}||_2$ & {\blue 0}  & 0.32 &  0.32 & 0.279 & 0.158   & 0.06 &  0.355 & 0.133     \\
\hline
\end{tabular}
    \caption{Train data set for the 8 protein network of interacting proteins shown in \cref{fig:DREMI_8}. %(CD4+, $CD3\_CD28$). 
    $\hat{A}$ from the training data is used on this data, in order to estimate goodness of fit of the inferred model. 
    Comparison is based on the average $L_2$ distance and we can deduce that there is partial transferability of the identified model due to unmeasured variables, as explained in the main text. Further experimentation using a quadratic dictionary instead, concludes to a better fit in the training set and slightly wider error bars in the test set, which indicates that depending on the application we can opt for different description of the dynamics.}
	\label{fig:DREMI_8testing}
\end{figure}

\begin{figure}[ht]
\centering
    \includegraphics[height=14em]{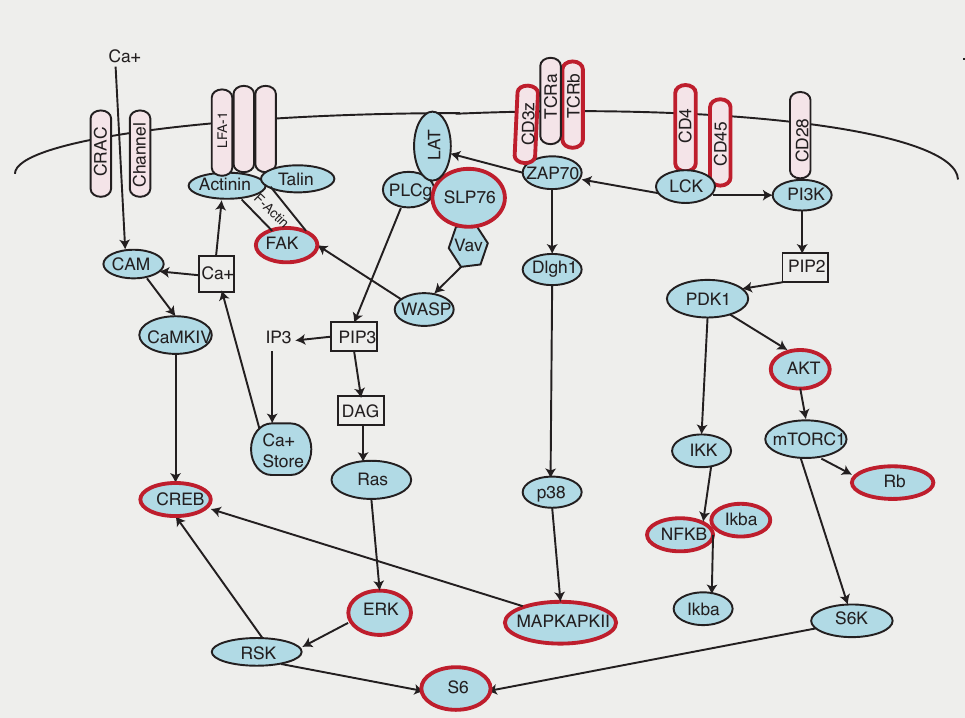}
    \caption{Experimental signaling pathways inside a T-cell. (figure taken from \cite{Krishnaswamy_DREMI} S.I.)}
    \label{fig:DREMI_net_suppl}
\end{figure}

We increased the number of variables to a sub-network of eight, which are four additional over the  {CD3z, Slp76, Erk, S6} namely: {MAPKII, Creb, Akt, Rb}.
\cref{fig:DREMI_8} shows the re-simulated data in order to match the algorithm assumptions and identifiability as discussed in the main text.
We seek for interactions already known by lab experiments as shown in \cref{fig:DREMI_net_suppl}.
Overall, we get a satisfactory fit based on the mean trajectory (black line with errorbars) by using a linear dictionary. The goodness of fit can be quantified by the average $L_2$ of the distance from the collocated trajectory, on which the re-simulated data is based. 
[variance of  each cloud is circa 0.24 and the B-spline support is 0.25. We get similar results for a narrower cloud provided that the B-spline support is reduced accordingly.]
Next in \cref{fig:DREMI_8testing}, we proceed with testing our models on unseen test data set, based on another activation (laboratory experiment with different dosage).
This is a harder inference problem than that of the four proteins only, as more unmeasured proteins might affect the behaviour of the added proteins.
The linear dictionary exhibits satisfactory fit and the errors accumulate over time, meaning that the variability of the generated trajectories increases. A quadratic dictionary concluded to a better fit for the training set though is less transferable to the test set.  

%------------------------------------------
\section{Quadruple-Well example: further experimental design}

\begin{figure}[h]
\centering
    \includegraphics[width=0.24\paperwidth]{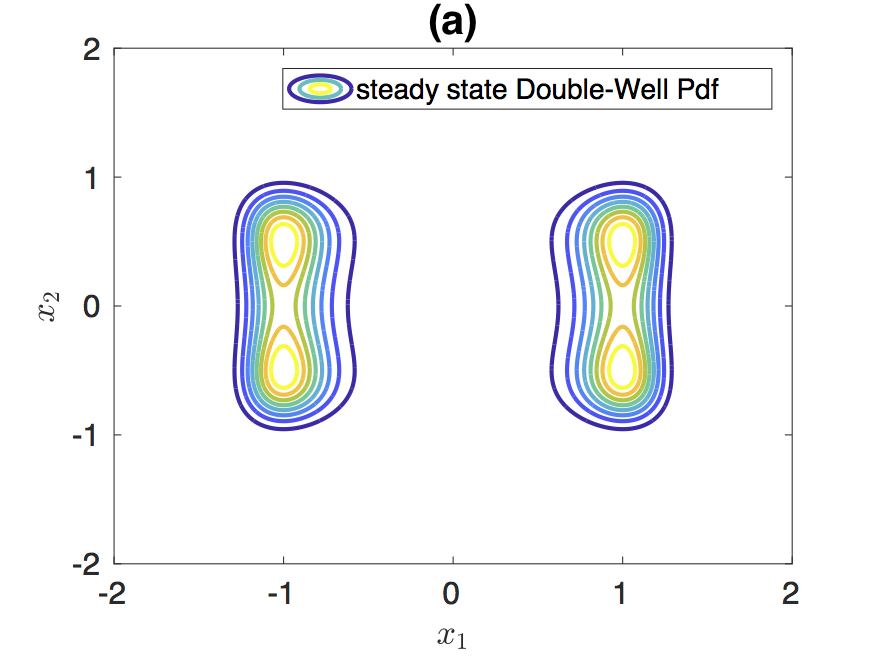}
 	\includegraphics[width=0.23\paperwidth]{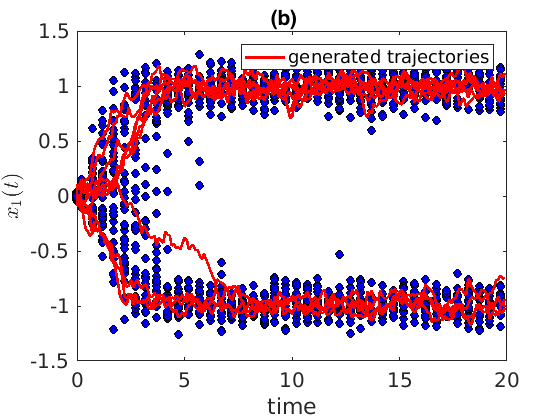}
    \includegraphics[width=0.23\paperwidth]{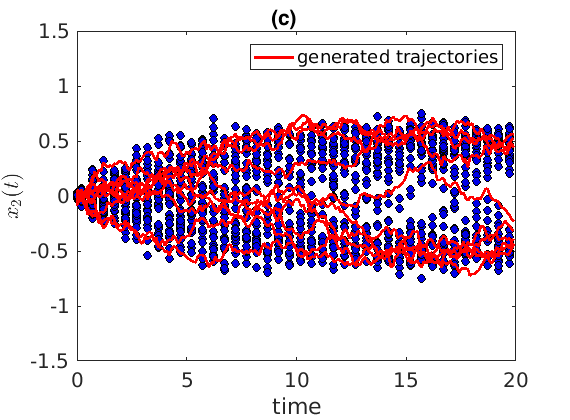}
\vspace{-3mm}
	\caption{ 
	$\mathbf{(a)}$ Contour plot of the level sets of the steady state joint pdf which is multimodal. 
	The quadruple-well potential restricts samples in four minima centered at $\{-1,-0.5\}, \{-1,0.5\}, \{1,-0.5\}, \{1,0.5\}$, so simple averaging of the data along each variable would result to samples around $\{0,0\}$ and inference is bound to fail.
	$\mathbf{(b)},\mathbf{(c)}$ Separately shown, the density evolution of bimodal stochastic processes $x_{1}, x_{2}$ of \Cref{eq:SDEs_N2} over time (blue dots).
	%$p(x_{1},t)$ starting from $P(x_1,t=0)=p_{0}(x_1)\sim \mathcal{N}(0,\sigma^{2}_{0}=0.1^2)$ 
	%and $P(x_2,t=0)=p_{0}(x_2)\sim \mathcal{N}(0,\sigma^{2}_{0}=0.1^2)$.
	The noise coefficients are $\{\sigma_1=0.2, \sigma_2=0.1\}$, resulting to larger variance of samples for process $x_1$.
	Upon inference of the system \cref{eq:SDEs_N2} based on time-course data, we generate stochastic trajectories (red solid lines) starting from the same distribution centered at $\{0,0\}$.
	For comparison, we used SINDy \cite{SINDy_2016} coupled with a cubic polynomial dictionary and input based on an average trajectory of the population data over each measurement timepoint. As expected, SINDy is not able to capture the bi-modality of population data, nor it is designed to be able to infer the stochastic noise coefficients $\sigma$. None of the terms in \Cref{eq:SDEs_N2} is correctly identified. 
	%The two modes are $\pm 1$ corresponding to the double well potential attaining minima at \{+1,-1\} for $x_1$ and \{+0.5,-0.5\} for $x_2$ respectively.
	} 
	\label{fig:two_sp_FP} 
\end{figure}

We sample from a pool of initial distributions $p_0(X;t=0)=\mathcal{N}(\mu_0,\sigma^{2}_{0})$
as shown in \cref{tab:p0}.

\begin{table} [ht]
\centering
\begin{tabular}{|c |c | c |}
\hline
$\mu_0(x_1)$ & $\mu_0(x_2)$ & $\sigma_0$  \\
\hline
\small
 0 &  0 &  0.1   \\
 -1 &  1 & 0.1         \\
 0.1 &  -0.1  & 0.2    \\
 -0.1 &  0.2 & 0.2  \\
 -0.2 &  0.1 & 0.1  \\
 -0.2 &  -0.2 & 0.15  \\
 -0.5 &  0.0 & 0.1  \\
 0 &  0.5 & 0.2  \\
 0.2 &  0.2 & 0.15  \\
\hline
\end{tabular}\caption{Parameter pool for the initial distribution $p_0(X;t=0)=\mathcal{N}(\mu_0,\sigma^{2}_{0})$ of the double-well system in the main text. When an activation intervention is performed, data are generated according to starting distribution $p_0$, for which parameters are randomly picked from this table.
The choice is arbitrary. %, though we chose means and st. dev. values for the system to be identifiable.
}
\label{tab:p0}
\end{table}

We further demonstrate the effect of the USDL-FP algorithm parameters $M_1, M_2$ and data paremters $NoS, dt$ on the relative error and the precision-recall metrics.
We set as the baseline example, 4 randomly picked interventions (activations) from \cref{tab:p0}, varying each parameter while keeping the others at their optimal value $(M_1=16, M_2=31, dt=0.1)$ for $NoS=200$ and $NoS=400$ samples per cloud. This is an identifiable problem for which we can retrieve the ground truth solution.
\begin{figure}[ht]
\centering
    \includegraphics[width=0.24\paperwidth]{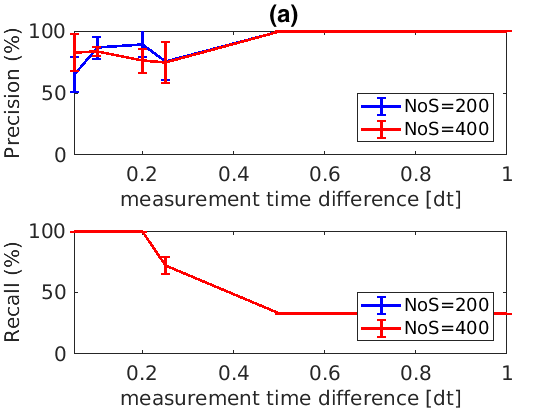}    
    \includegraphics[width=0.24\paperwidth]{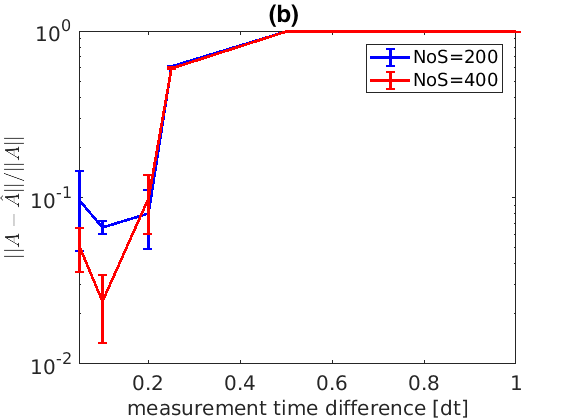}
    \includegraphics[width=0.24\paperwidth]{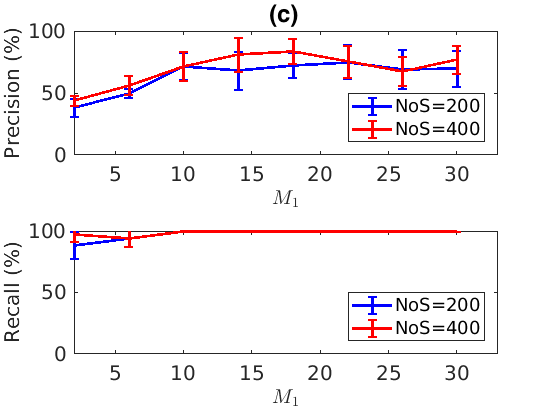}
    \includegraphics[width=0.24\paperwidth]{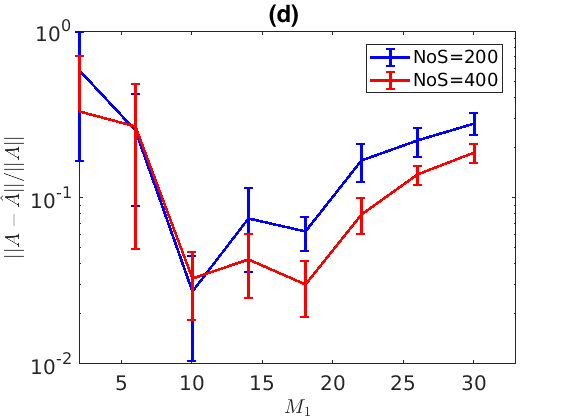}
    \includegraphics[width=0.24\paperwidth]{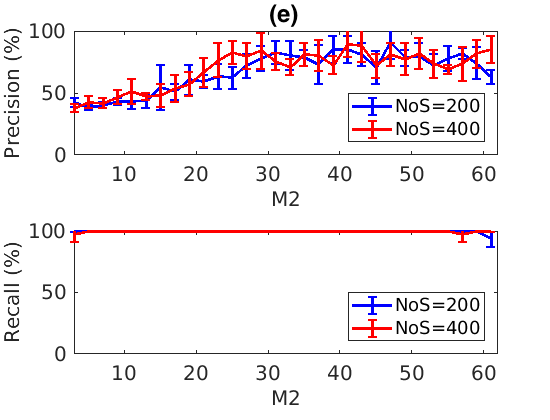}
    \includegraphics[width=0.24\paperwidth]{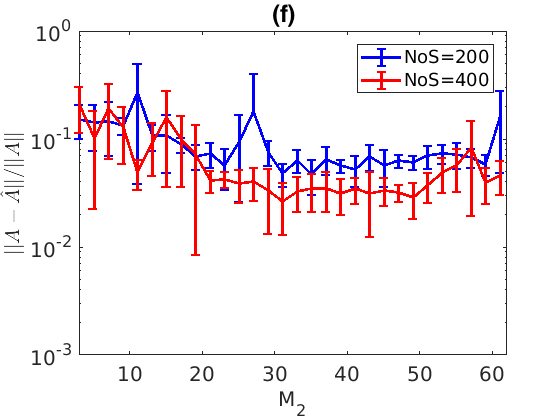}
    \caption{Parameter tuning of the quadruple-well example shown in the main text with respect to hyper-parameters $dt, M_1, M_2$, for low $NoS=200$ and high $NoS=400$ sample sizes. The results are based on 4 activations, randomly chosen from \cref{tab:p0}.
    The effect of every parameter variation is measured according to the precision, recall and relative error metrics.
    The optimal parameters being fixed while varying each one at a time are: $M_1=16$ B-splines, $M_2=31$ Fourier modes, measurement time interval $dt=0.1$
    %[{\red hard thresholding at 0.01}].
    In {\bf(a),(b)} we observe that inference breaks down for clouds of data being measured $dt>0.2$ time units. The cloud average variance is $0.4$ and $0.2$ for process $x_1$ and $x_2$ respectively. 
    As shown in {\bf(c),(d)} $M_1$ in the range of $[10:18]$ is favourable, meaning that the corresponding support of B-splines is the range $[0.22:0.4]$. Spatial test functions should be comparable to the cloud width, in order to encode their displacement over time.
    In {\bf(e),(f)} we investigate the effect of the number of Fourier modes used and a value in the range of $[25:50]$. }
    \label{fig:DW_supplementary}
\end{figure}

As shown in \cref{fig:DW_supplementary}(a)-(b), as the distance between sampling times $dt$ is increased (lower measurement frequency), inference is deteriorated until collapsing for values greater than 0.2.
This happens because we get less information from the dynamics of the stochastic processes.
As shown in the main text, the number of spatial test functions $M_1$ is related to the variance of each cloud, so the optimal value (close to) $M_1=10$ results to the support of each B-spline to be (circa) 0.4 which in turn is close to the cloud width of each mode of the SDE's. $M_1$ values significantly
diverging from this value, utilizing either more or less test functions over the same range, alters the B-splines support and thus are not able to capture the clouds evolution \cref{fig:DW_supplementary}(c)-(d).
Mathematically, this translates to  worse estimate of integral computations in the weak formulation, because of a small fraction of samples per B-spline.
Temporal test functions $M_2$ should be set over 31 in order to capture the frequency changes of the spatially projected data. For the whole range of values tested, we conclude that it  is not a very sensitive parameter with respect to the False Negatives though, a smaller value will decrease the precision by inclusion of false positive terms \cref{fig:DW_supplementary}(e)-(f).
It is evident that more samples per cloud result to better performance overall.

%[Next we get a thorough overview on the performance metrics by altering two parameters at the same time. \Cref{fig:DW_supplementary_thorough_M2_7} and \Cref{fig:DW_supplementary_thorough_M2_31} show how the relative error as well as precision-recall over varying the measurement frequency $dt$ and $M_1$ having $M_2$ fixed.
%General observations are that the number of samples per cloud $NoS$ either for 100 or 500 samples, shows the same underlying behavior over the variation of the metrics as in \Cref{fig:DW_supplementary}, though for $NoS=500$ (shown here) the plots have less variance and better clarity.

An important additional result over the previous figure, having set as baseline parameters: $dt=0.1, M_1=16, M_2=32$, precision, recall and relative error is maximized, thus inference is optimized.
%\begin{figure}[H]
%\centering
%    \includegraphics[width=0.7\paperwidth]{./figs/Double_Well/Err_vs_dt_VARIOUS_M1_NoS_500_M2_31.png}
%    \includegraphics[width=0.7\paperwidth]{./figs/Double_Well/prec_rec_vs_dt_VARIOUS_M1_NoS_500_M2_31.png}
%    \caption{Further experimentation on the Double-well hyper-parameter tuning. We set {\bf M2=31} over varying $dt$ and $M_1$ and monitor the effect on the same metrics. Longer measurement intervals $dt$ require less Fourier modes $M_2$.}
%    \label{fig:DW_supplementary_thorough_M2_31}
%\end{figure}

%--------------------------------------------------------
\section{Cascade: further experimental design}
\begin{figure}[ht]
\centering
    \includegraphics[width=0.3\paperwidth]{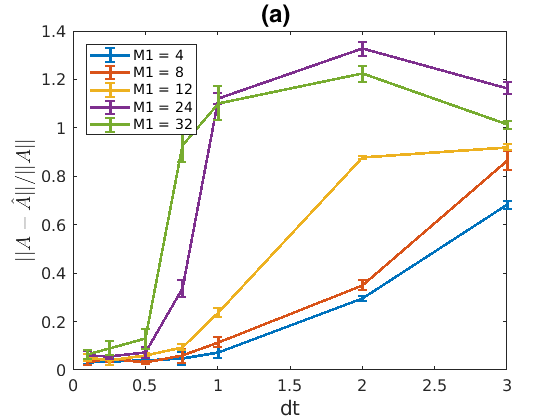}
    \includegraphics[width=0.3\paperwidth]{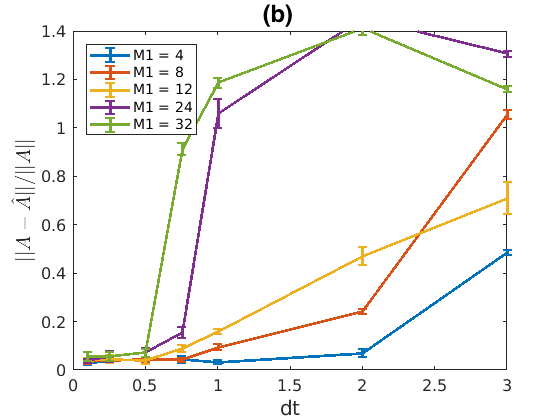}
    \includegraphics[width=0.3\paperwidth]{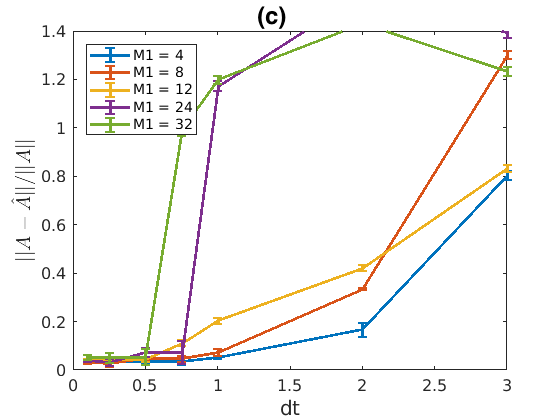}
    \includegraphics[width=0.3\paperwidth]{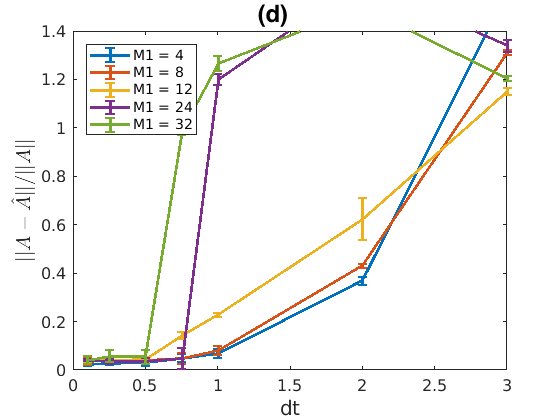}
    \caption{Further experimentation on the Cascade hyper-parameter tuning. We set {\bf M2=3,7,11,15} ${\bf(a)}$-${\bf(d)}$ over varying $dt$ and $M_1$ and monitor the effect on the same metric. Longer measurement intervals $dt$ require wider B-splines and less Fourier modes $M_2$, indicating interdependence of those parameters.}
    \label{fig:CASCADE_supplementary_thorough}
\end{figure}

In \Cref{fig:CASCADE_supplementary_thorough} we continue the experimental design of hyper-parameter tuning section of the main text. Instead of monitoring the relative error with respect to one parameter at a time, while setting the remaining ones to their optimal values, we look into the range of two parameters at a time have the third ($M_2$) fixed.
We deduce that in the case of sparse measurement time difference $dt=1$ we can infer a model with moderate error provided that we employ less B-splines i.e. B-splines with wider support. 
In addition we need to use less Fourier modes.
In this way we show the inter-dependence of the algorithm parameters (within a range for each one), that should be performed with respect to a metric. Multiple parameter sets conclude to similar $\hat{A}$ matrices.
Depending on the 
system under study, we might need to consider other metrics as well.

%[{\blue Put figures for Rel err vs dt over M1 here}]

%--------------------------------------------------------

%--------------------------------------------------------
\section{High sampling time difference: data re-simulation}
As mentioned in the main text, the available mass-cytometry data have a varying variance over time. This is a result of
the direct single-cell analysis and accompanied transformation of the raw data. In this current work, our proposed USDL framework coupled with a stochastic differential equation describing the data, assumes
that the variance of the random term is constant i.e. $\sigma(t)=\sigma$. In addition, the stochastic noise term is assumed to be Brownian motion, being a (multimodal in principle) Gaussian. On top of this data peculiarity we are faced with infrequent data measurement times. Considering the above limitations, we 
apply a constrained least squares optimization technique with regularization (smoothing penalty), termed collocation method \cite{USDL_bioinformatics, Ramsay_Book}, in order to export the mean
weighted trajectory through population data clouds. This is the cornerstone, on which re-simulated population data are generated and given to the USDL-FP algorithm. The collocated trajectory takes account of the variance and relative position of the population data. 
These re-simulated data are normally distributed and centered around the collocated trajectory.
We stress the fact that collocation is performed with caution to prevent loss of information by possible over-smoothing and the variance is close to the original. We remark that this methodology, concludes to a more identifiable system, whereas naive averaging would miss dynamical information.

As we showed in the mass cytometry data paradigm in the main text, where sampling time difference $dt$ is high, we encountered difficulties during inference. A way to circumvent is by data re-simulation via the generation of a constrained mean trajectory (collocation method \cite{Ramsay_Book})
with smoothing penalty \cite{Wabba_smoothing_NumAn_1978}, over the clouds of data. Following this, we populate the mean trajectory with clouds of samples at intermediate time intervals of
smaller time difference $dt$ than the original data set.
Care is taken such that the re-simulated clouds have the same variance as the original ones.

We demonstrate our proposed re-simulation based on the four protein Cascade by removing $85\%$ of the clouds of data. In this way, sampling time difference is increased to $dt>5$ where we
have already shown  %\Cref{fig:cascade_4sp_RelErr}(b) 
that inference collapses. The purpose of this test allows us to ``augment" (in some sense) infrequent temporal measurements as a re-sampling preprocessing stage, in a data set where there is not
enough dynamical information to deduce the underlying equations.

\begin{figure}[h]
    \includegraphics[width=0.7\paperwidth]{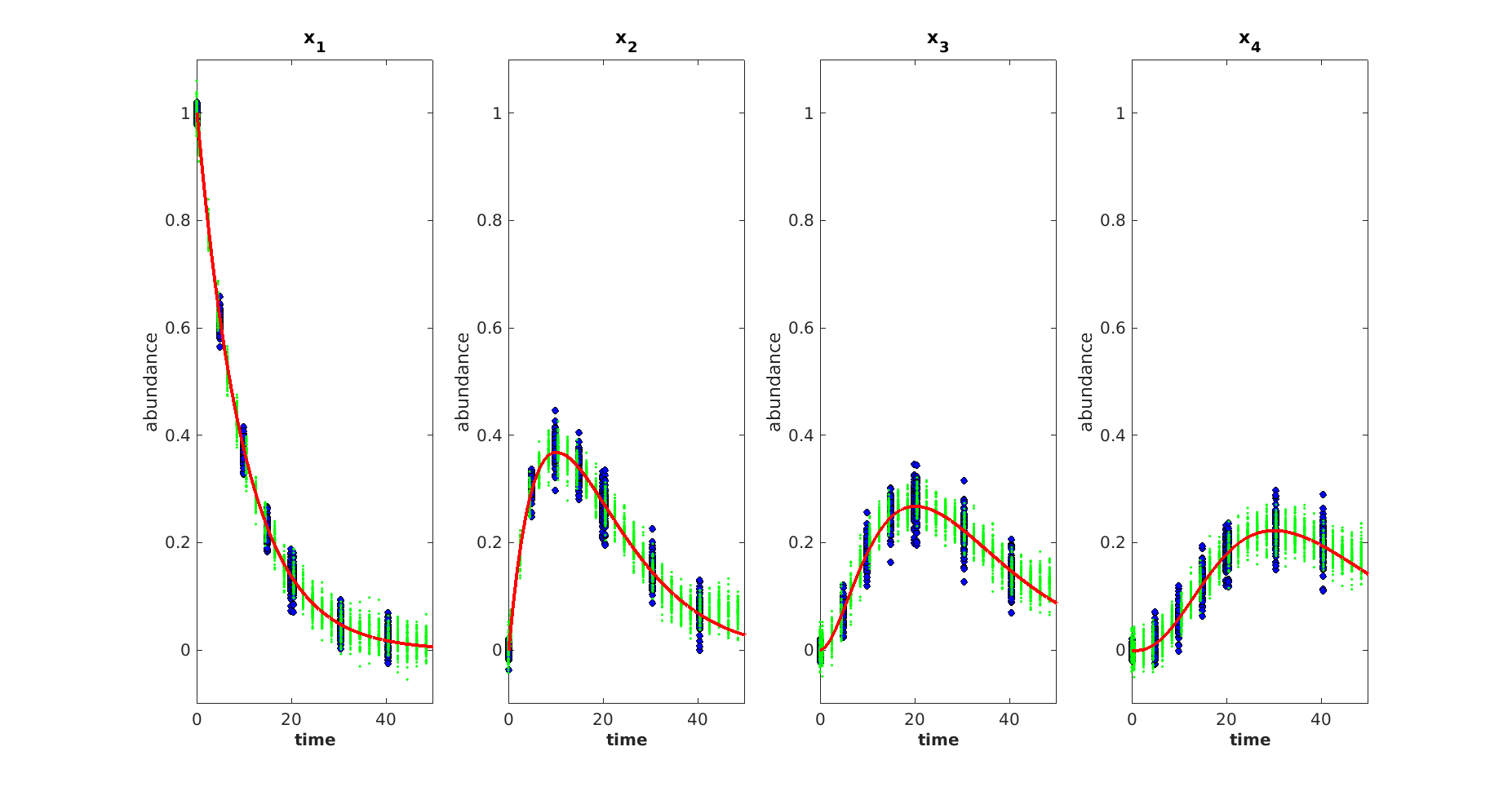}
	\caption{Cascade paradigm using re-simulation in the high sampling difference regime. We keep $15\%$ of the original data set (blue clouds) with respect to the sampling time points and try to infer the dynamics. The re-simulation methodology steps are: 
	i) apply the collocation method with smoothing penalty on the input population data, in order to generate a weighted trajectory  $X_{\text{col}}$,
    ii) determine the time intervals containing steep derivatives and chose intermediate time points $t_i^{\text{inter}}$ between measurement times (here $dt=0.5$ as the original data set), 
    iii) compute average variance $\bar{\sigma}_n$ (per variable) in those intervals, 
    iv) resimulate normally distributed clouds ($\mathcal{N}(0,\bar{\sigma}_n)+X_{\text{col}}$) at $t_i^{\text{inter}}$, here shown in green. Then the inference using the resimulated data concludes to $\hat{A}$, which we then use to generate trajectories here shown in red solid lines.
    The relative error is $0.05$ (instead of 0.75), similar to that of $dt=0.5$ of the original data set and the noise coefficient $\sigma$ is recovered.
}
	\label{fig:cascade_resimulation}
\end{figure}

A point to consider for this low dimensional example is that the solution $\hat{A}$ based on a linear dictionary resembles the least-squares solution, though for bigger  dictionary sizes we get sparser solutions. We note that this re-simulation
does not alter the structure of the inferred system and the noise coefficient is precisely recovered, provided that appropriate tuning is done, as shown in the previous section.

%===============================================================

\section{Inhibitions}

Formally, we impose interventions by additional ``input signal" (measurement data) and not by induction of hard constraints to the minimization
problem \Cref{eq:SSR_L0}.
\begin{equation}\label{eq:interv}
    dx^{(n)}_{t} = \sum_{q=1}^{Q} a_{nq}\psi_q(x) + {\blue b_n u_n} +  \sigma_n dW^{(n)}_{t}, \quad n=1,\dots, N
\end{equation}
where $b_n\in \mathbb{R}$ reflects the strength of the input signal $u_n$ to the $n$-th variable $x_n$.
For instance, an inhibition of the form $X_{n^{*}}\rightarrow 0$ is
modeled by setting $u_{n^{*}}=x_{n^{*}}$ and $u_n=0$ for $n\neq n^{*}$.

The weak space projected system for the $n$-th variable is given by:
\begin{equation}
    Z_n = \Psi_n a_{n} + {\blue b_n v_n}
\end{equation}
where $Z_n, \Psi_n$ as in main text \ref{sec:intervention}, where $a_n$ incorporates unknown diffusion coefficient $\sigma_n$ while $v_n$ is the \underline{projection} of the intervention input signal to the test functions $v_{nm}=\langle u_n, \phi_m \rangle$.

For the $r-$th inhibition, this system is written as:
\begin{equation}
    Z^{(r)}_n = \Psi^{(r)}_n a_{n} + {\blue b^{(r)}_n v^{(r)}_n}
\end{equation}

Upon merging all the $R$ interventions (for the $n$-th variable) we get in matrix form \cref{eq:MatrixFormInterv}:
\begin{figure}[H]\label{eq:MatrixFormInterv}
\centering
   $\begin{bmatrix} 
            Z^{(1)}_n   \\
            Z^{(2)}_n     \\
            \vdots  \\
            Z^{(R)}_n 
        
    \end{bmatrix}$=
    $\begin{bmatrix} 
            \Psi^{(1)}_n & { \blue v^{(1)}_n} & 0 & \dots & 0  \\
            \Psi^{(2)}_n & 0       & {\blue v^{(2)}_n} &\dots & 0    \\
            \vdots & \vdots  & \vdots &\dots & \vdots \\
            \Psi^{(R)}_n & 0 & 0 & \dots & {\blue v^{(R)}_n}
        
    \end{bmatrix}$
    $\begin{bmatrix} 
            a_n \\
            {\blue b^{(1)}_n}    \\
            \vdots  \\
           {\blue b^{(R)}_n}
        
    \end{bmatrix}$
    \caption{Interventions matrix form for the $n$-th variable}
\end{figure}

where the vector $a_n$ corresponds to the unknown dictionary coefficients and $b^{(r)}_{n}$ to the unknown inhibition coefficients, all to be inferred using OMP, as it remains a problem with a sparse solution. In total, the unknowns sum to $Q+R$ coefficients to be inferred per variable $n$.

As mentioned in the main text, an activation is
a predefined initial condition, so it is modeled by setting $u_n=0$ for the $n-$th corresponding variable, meaning no additional signal. In effect,
for that variable the corresponding $b_n=0$.

Solving for all possible interventions of the variables $n'\neq n$, we conclude to:

\begin{figure}[H]\label{eq:MatrixFormInterv_ALL}
\centering
   $\begin{bmatrix} 
            Z^{(1)}_n   \\
            Z^{(2)}_n     \\
            \vdots  \\
            Z^{(R)}_n 
        
    \end{bmatrix}$=
    $\begin{bmatrix} 
            \Psi^{(1)}_n & {\blue {v^{(1)}}} & 0 & \dots & 0  \\
            \Psi^{(2)}_n & 0       & {\blue v^{(2)}} &\dots & 0    \\
            \vdots & \vdots  & \vdots &\dots & \vdots \\
            \Psi^{(R)}_n & 0 & 0 & \dots & {\blue v^{(R)}}
        
    \end{bmatrix}$
    $\begin{bmatrix} 
            a_n \\
            {\blue b^{(1)}}    \\
            \vdots  \\
           {\blue b^{(R)}}
        
    \end{bmatrix}$
    \caption{Every possible intervention in matrix form for the $n$-th variable.}
\end{figure}

where $b^{(r)} \in \mathbb{R}^N, u^{(r)} \in \mathbb{R}^N$, $a_n \in \mathbb{R}^{Q+1}$.

%--------------------------------------------------------
\section{Three protein pathway}
We aim to infer the three protein pathway Slp76 $\rightarrow$ Erk $\rightarrow$ S6.
Despite the fact that the complete biochemical network is probably nonlinear with respect to 
the variables, the assumed model for inference is linear $\frac{dX}{dt}=AX$, similar to synthetic cascade example ($\Sigma=0$). The linear model is sufficient for this low dimensional\textbackslash single maximum paradigm and at the same time, the associated connectivity matrix $A$ encodes only the direct causal interactions between the variables.
Another inherent difficulty on the inference of this example is: a) the measurement error associated with machine errors, assumed to be additive and b) uncertainty error due to the 
fact that each measurement comes from a different cell and each cell has different concentrations of the measured quantities.

\begin{figure}[ht]
     \includegraphics[width=0.25\paperwidth]{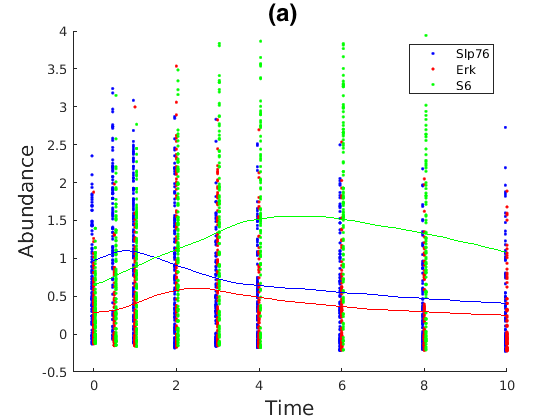}
    \includegraphics[width=0.15\paperwidth]{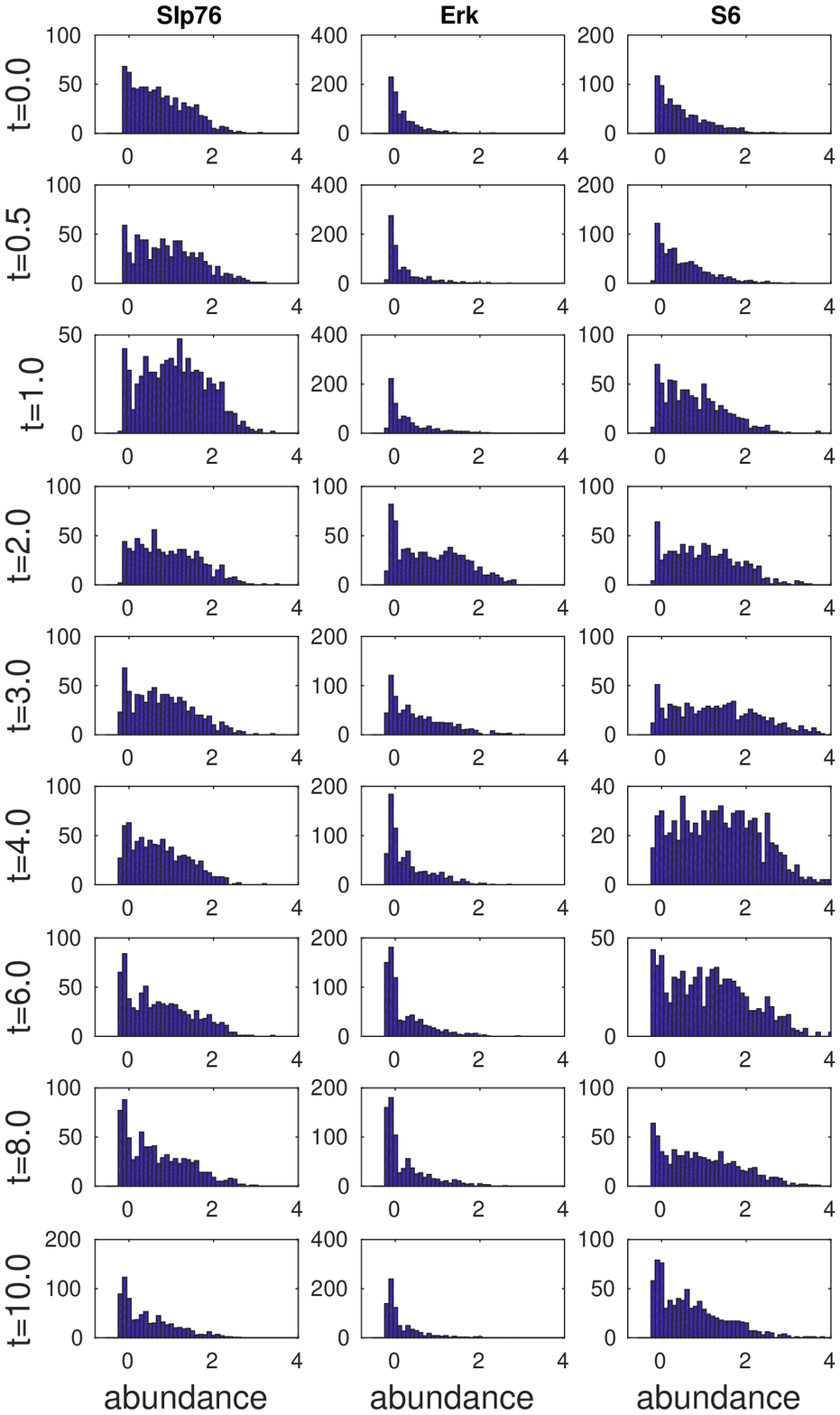}
    \includegraphics[width=0.25\paperwidth]{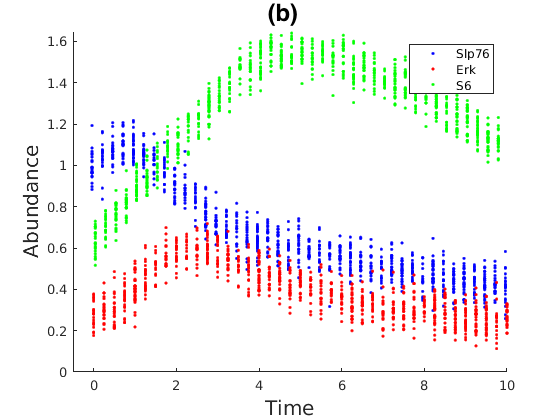} 
    \includegraphics[width=0.22\paperwidth]{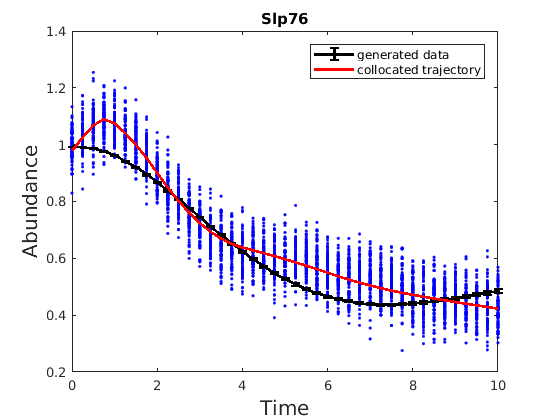}
    \includegraphics[width=0.22\paperwidth]{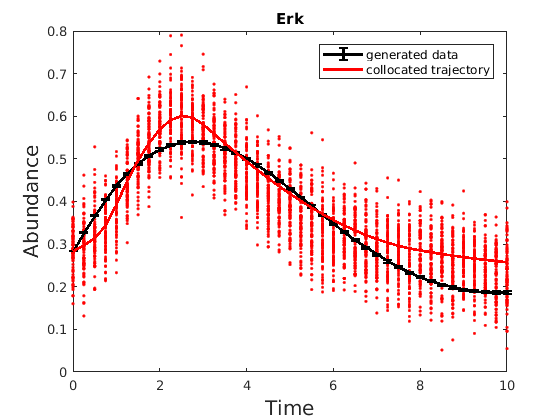}
    \includegraphics[width=0.22\paperwidth]{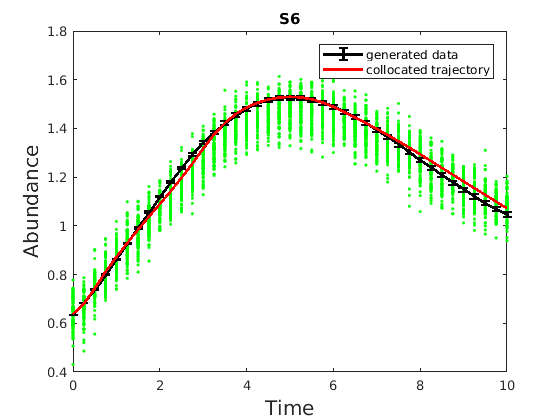}
    \begin{tabular}{|c ||c | c | c|}
\hline
reaction & $a_1$ & $a_2$ & $a_3$  \\
\hline
 SLp76 &  0.138 &  -0.593 &  0.065   \\
 Erk &  {\blue 0.306} & {\blue -0.192}       &  -0.103    \\
 S6 &  0.096  & {\blue 0.979} & {\blue -0.305}  \\
\hline
\end{tabular}
	\caption{Mass cytometry population data for 3 proteins SLp76, Erk and S6. ${\bf(a)}$ The solid lines show the collocated trajectories which are constrained averages of the abundance clouds of samples in $[0:10]$.
	The collocated trajectories characterize the re-simulated data.
	${\bf(b)}$  Re-simulation of the collocated trajectories: Gaussian clouds with measurement variance 0.24 and measurement time difference $dt=0.25$ seconds.
	(lower panel) Histogram of the protein abundances over measured times of ${\bf(a)}$, before re-simulation. Re-simulation is needed because the sampling times are sparse and the evolving distributions are non-Gaussian, which violates the modelling assumption of USDL FP.
	(lower panels) Population data from ${\bf(b)}$ against the average collocated trajectory (red) and average generated trajectory from $\hat{A}$ matrices over 100 runs of USDL FP with errorbars (black). 
	Each run is based on a random subsampling of the original mass-cytometry data set. Although the fitting is satisfactory with respect to correct identification of the maxima, it does not capture the collocated trajectories exactly. Upon inspection of a randomly picked $\hat{A}$, the model for the driving protein Slp76 is based on the remaining which is a modelling restriction of a closed system used. In the next example we rectify this by adding prior knowledge. [We used a linear dictionary without constant terms.]}
	\label{fig:Dremi_data_N_3}
\end{figure}

\Cref{fig:Dremi_data_N_3} shows the input population data before and after re-simulation. The original data are skewed though their variance in the re-simulated is retained up to the point of an interpretable system depicting clear variations [is this correct?]. 
The re-simulated clouds are closer and normally distributed in contrast to the unevenly distributed original ones, and this property meets the assumptions of the algorithm. 
\Cref{fig:Dremi_data_N_3} shows the generated trajectories based on each inferred connectivity matrix $\hat{A}$ (solid lines) by subsampling the original data-set . Although the overall fit is satisfactory, the first protein SLp76 maximum abundance at $t=1.5$ is not accurately captured. As we explain in the next section, this is because of the fact that we try to model a closed system, that is describe the behavior of SLp76 using the other two, which increase at later times.

%---------------------------------------
\section{4 Protein Network: further experiments}
\begin{figure}[ht]
\centering
	\includegraphics[width=0.17\paperwidth]{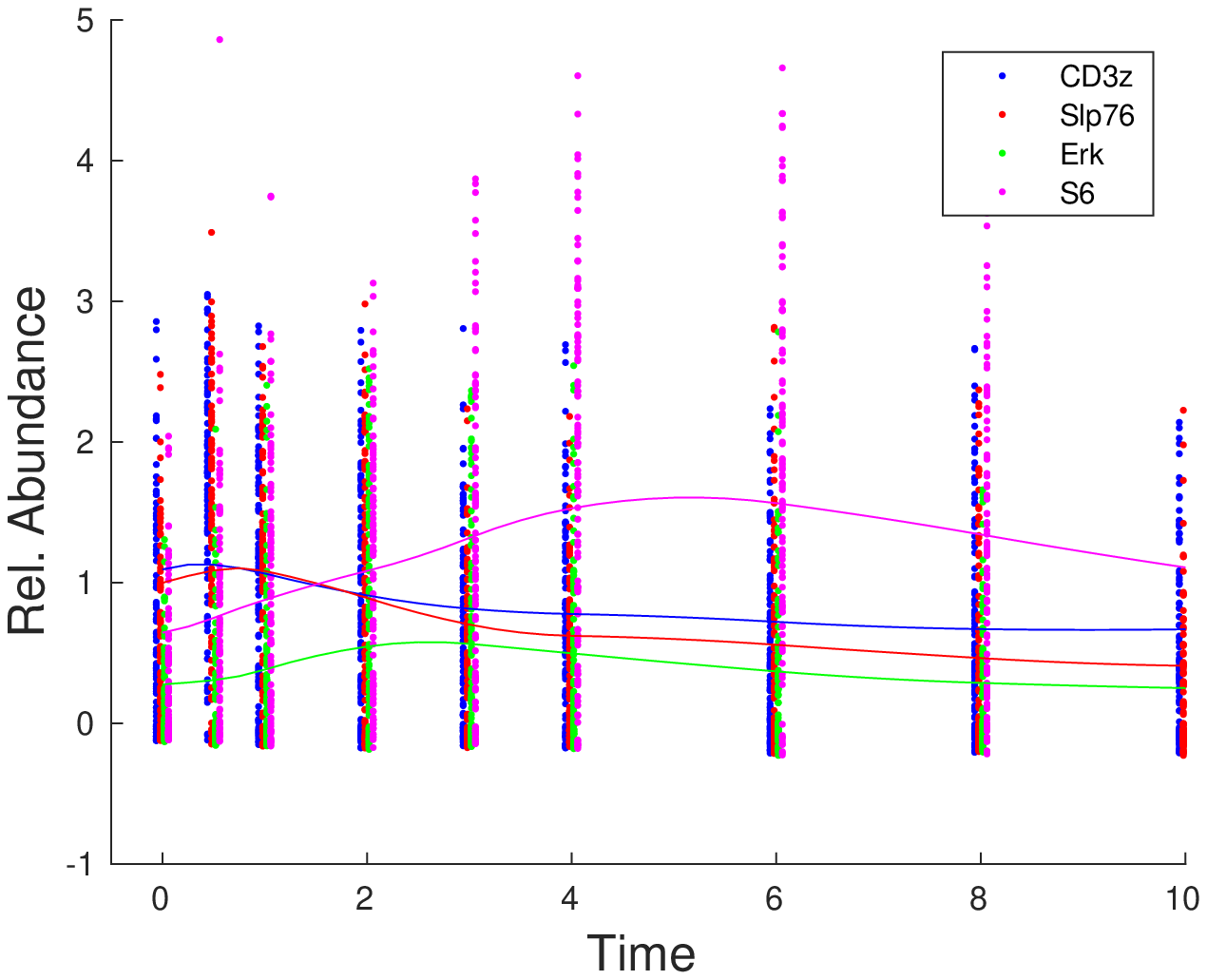}\\
	\includegraphics[width=0.17\paperwidth]{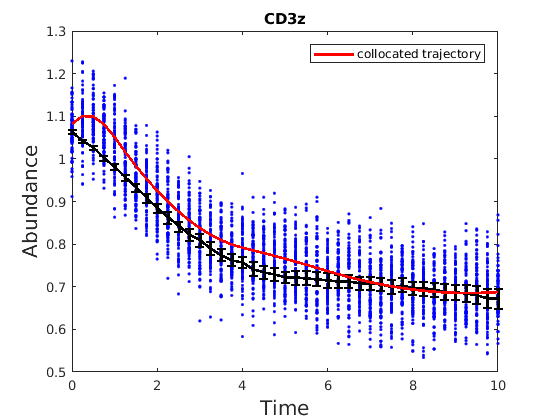}
	\includegraphics[width=0.17\paperwidth]{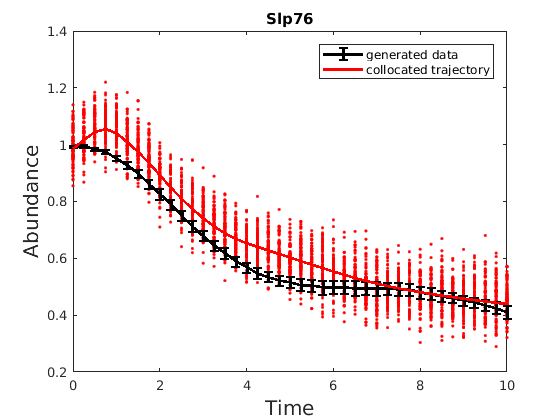}
	\includegraphics[width=0.17\paperwidth]{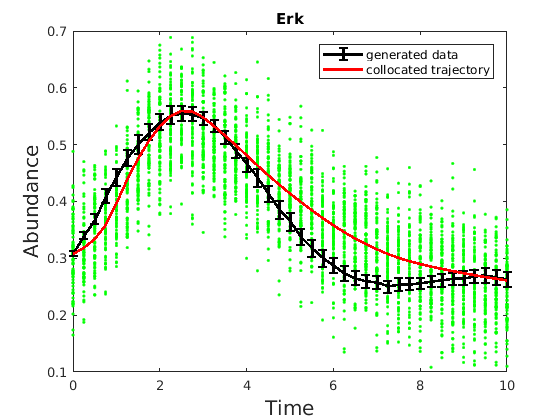}
	\includegraphics[width=0.17\paperwidth]{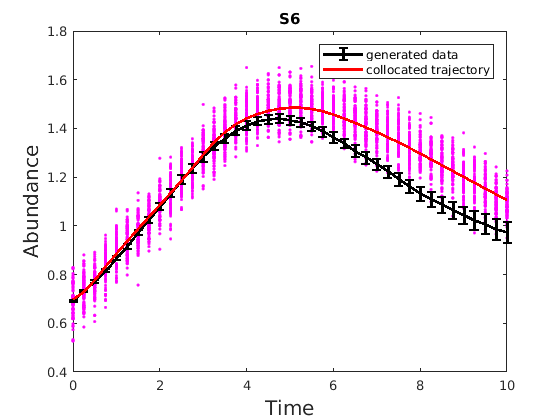}
	\includegraphics[width=0.01\paperwidth]{./figs/train.png}\\
	\includegraphics[width=0.17\paperwidth]{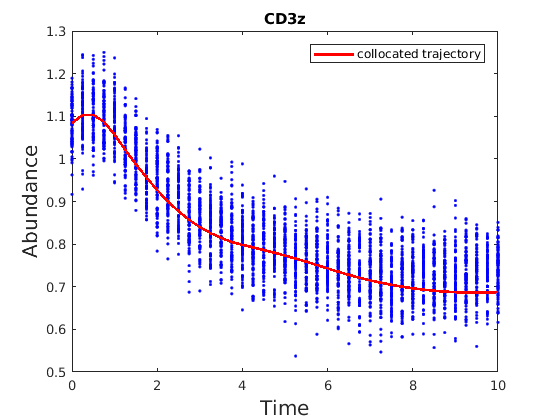}
	\includegraphics[width=0.17\paperwidth]{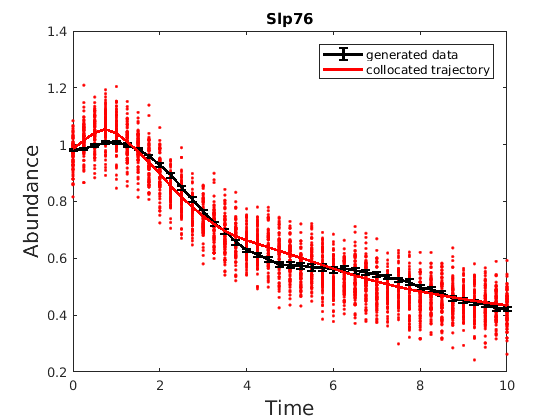}
	\includegraphics[width=0.17\paperwidth]{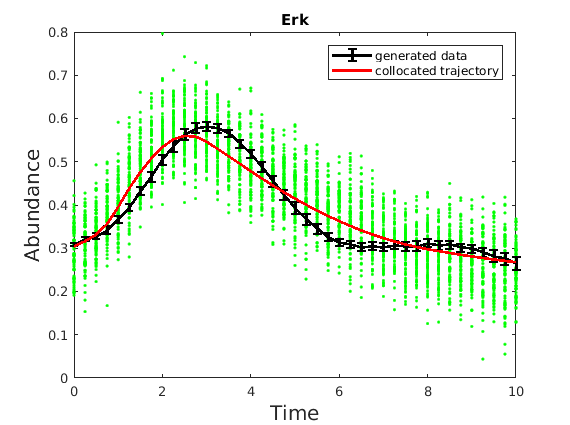}
	\includegraphics[width=0.17\paperwidth]{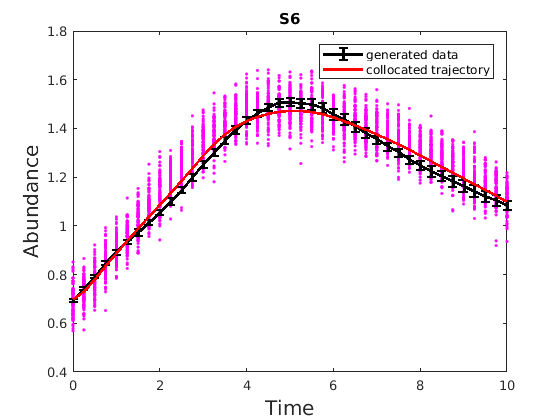}
	\includegraphics[width=0.01\paperwidth]{./figs/train.png}	 \begin{tabular}{|c ||c |c |}
    \hline
    species & $||X_{coll}-X_{gen}||_2$ & $||X_{coll}-X_{gen}||_2$  (driving force) \\
    \hline
     CD3z & 0.05 &  {\blue 0} \\
     SLp76 & 0.098 & 0.07    \\
     Erk & 0.11 & 0.08          \\
     S6 & 0.08 & 0.04      \\
    \hline
    \end{tabular}
    \caption{(Upper panel) Mass cytometry population data (dots) of 4 proteins at 9 measurement timepoints, along with their collocated trajectories (solid lines). 
    Inference using linear dictionary plus constant term.
    (middle panel row) Re-simulated population data based on collocated trajectories. The black solid line with errorbars represents, the averaged generated trajectory over the ones by using $25$ $\hat{A}$ matrices. Each matrix is computed on one of the $25$ subsampled sets of the training data set. %$M_2=15$. 
    The generated trajectories capture the minima of the dynamics with small discrepancies. 
    (lower panel row) Using CD3z as driving force instead of generating its dynamics/trajectories based on $\hat{A}$. We see that prior knowledge improves fitting due to the fact that unmeasured quantities affect the CD3z protein. Inference is slightly inferior to the one with additional quadratic terms dictionary, as shown in the main text.}
	\label{fig:DREMI4_t1}
\end{figure}
\Cref{fig:DREMI4_t1} shows the average generated trajectory based on a 25 inferred connectivity matrices $\hat{A}$ by using a linear dictionary (plus constant term).
As it is evident from the figure and supported by the $L_2$ distance, inference is slightly inferior for SLp76 that the one shown in the main text, using a quadratic dictionary (without cross terms). In \Cref{fig:DREMI4_t1}(lower row of panels), we get a better fit upon using the CD3z data as prior knowledge. This choice is based on known results from bibliography on this subnetwork, see \cref{fig:DREMI_net_suppl} for comparison. 
As stated in the main text, there might exist unmeasured/hidden mechanisms affecting this driving protein, that cannot be explained from the other proteins considered as input.

\begin{figure}[ht]
\centering
	\includegraphics[width=0.17\paperwidth]{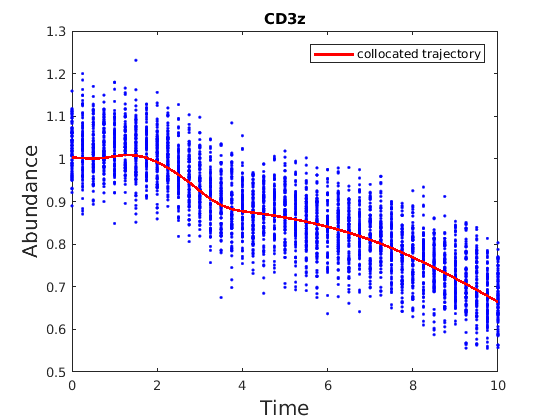}
	\includegraphics[width=0.17\paperwidth]{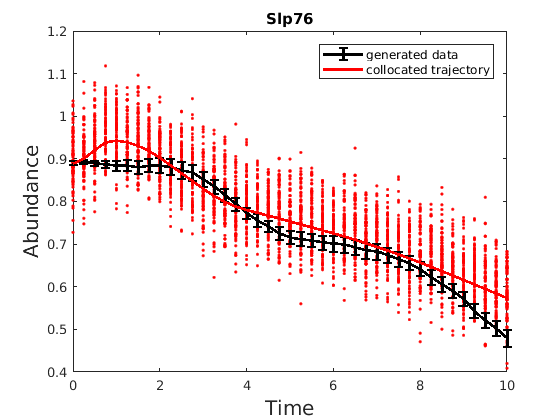}
	\includegraphics[width=0.17\paperwidth]{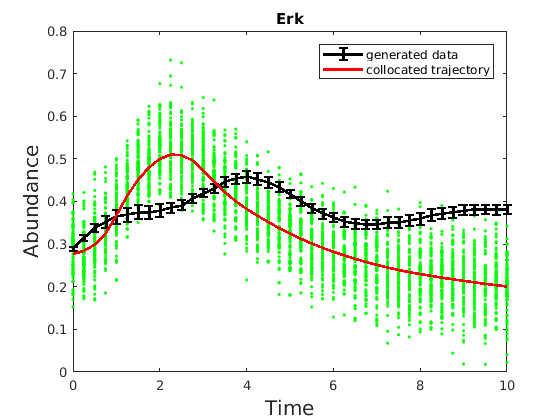}
	\includegraphics[width=0.17\paperwidth]{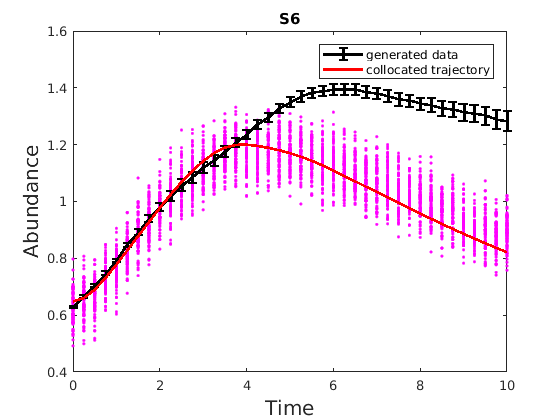}
	\includegraphics[width=0.01\paperwidth]{./figs/test_label.png}
	 \begin{tabular}{|c ||c |}
    \hline
    species & $||X_{coll}-X_{gen}||_2$   \\
    \hline
     CD3z &  {\blue 0} \\
     SLp76 &  0.065    \\
     Erk &  0.33          \\
     S6 &  0.26      \\
    \hline
    \end{tabular}
    \caption{Test set of protein inference using linear dictionary and $\hat{A}$ matrices based on training set of \cref{fig:DREMI4_t1}. Our assumption is that the underlying mechanism is the same in both data sets, though this might be obscured by confounding.}
	\label{fig:DREMI4_t1_forecast}
\end{figure}

We proceed and test the USDL-FP inferred $\hat{A}$ models on unseen test data of a different activation \cref{fig:DREMI4_t1_forecast}. In comparison to the quadratic dictionary results shown in the main text, we get a slightly inferior fit.

%===============================================
\section{Projections}
We use B-splines as spatial test functions and project data over each time point. 
\begin{figure}[ht]
\centering
	\includegraphics[height=14em]{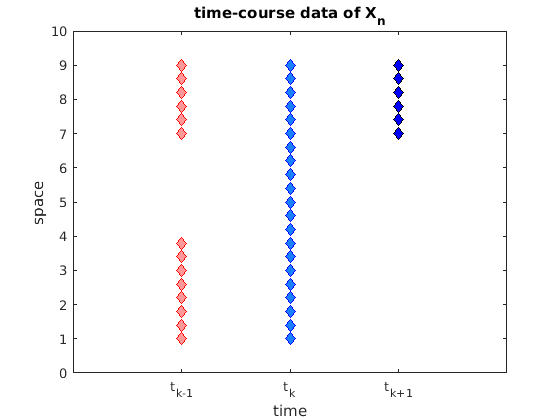}
	\includegraphics[height=15em]{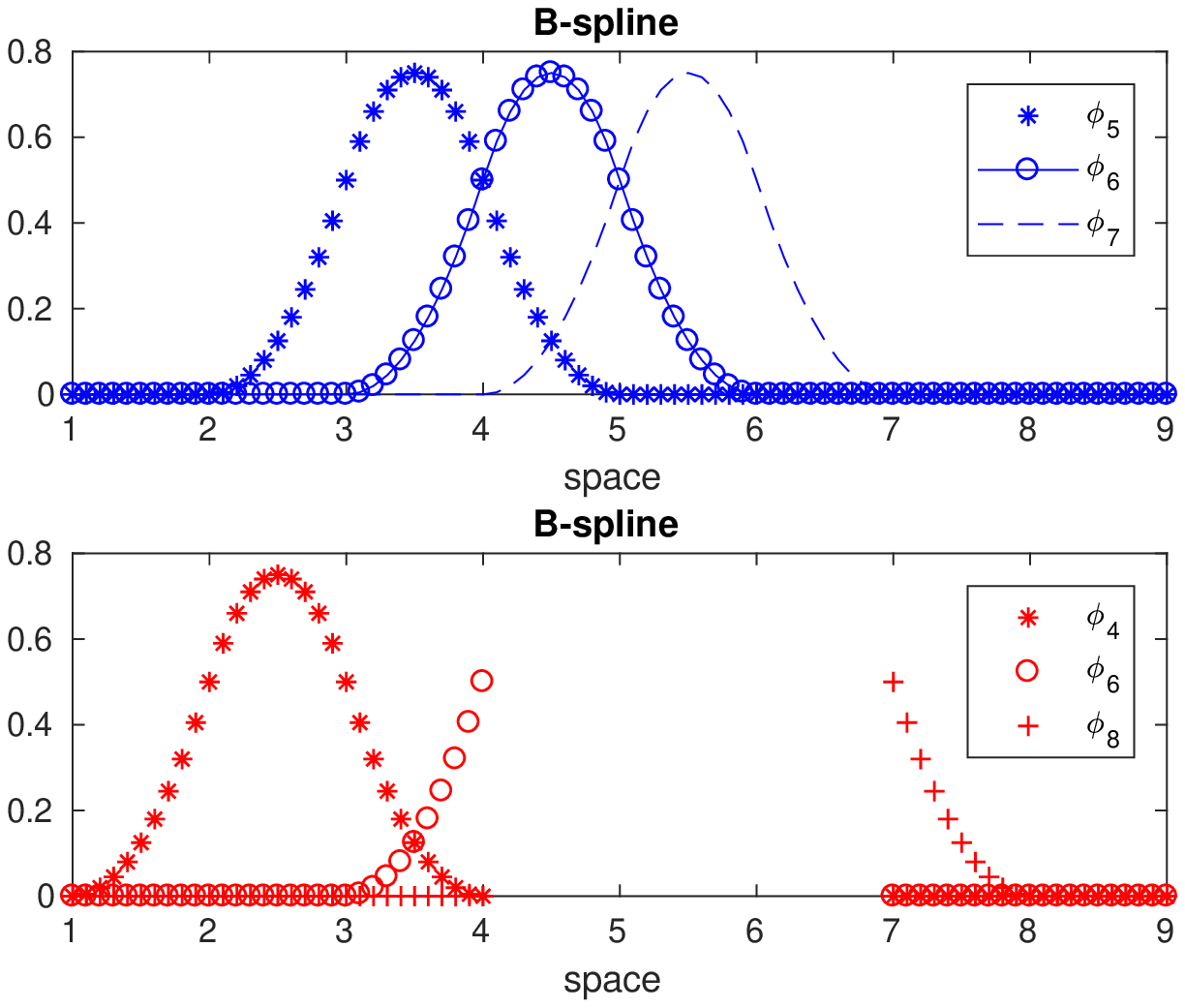}
	\caption{(left panel)Time-course data for variable $X_n$ (upper right panel) Weak space projection of three (out of $M_1=10$) B-splines are shown for the $X_n(t_k)$ time-point samples from population data of variable $X_n$ (lower right panel) three spatial B-spline test functions on the data of the same variable $X_n$ at time $t_{k-1}$.
     Their functional form is fixed irrespective of the data. In the range $[4:7]$, no data exist to be projected. Computations of integrals of the weak form with respect to $p(x,t)$, depend on the appropriate choice of spatial test-functions. Too many B-splines with narrow support, cannot track the displacement of the clouds of data.}
	\label{fig:Bsplines_func}
\end{figure}
\Cref{fig:Bsplines_func} shows the projections of time-course data, using equally-spaced, quadratic B-splines on the domain [0,10]. 
By employing additional B-splines, spatial information is encoded across more functions, thus small variability can be captured provided that the sample points are not too few so that the integral etsimations of the weak form using sampling are valid.

\begin{figure}[ht]
%\centering
	\includegraphics[height=10em]{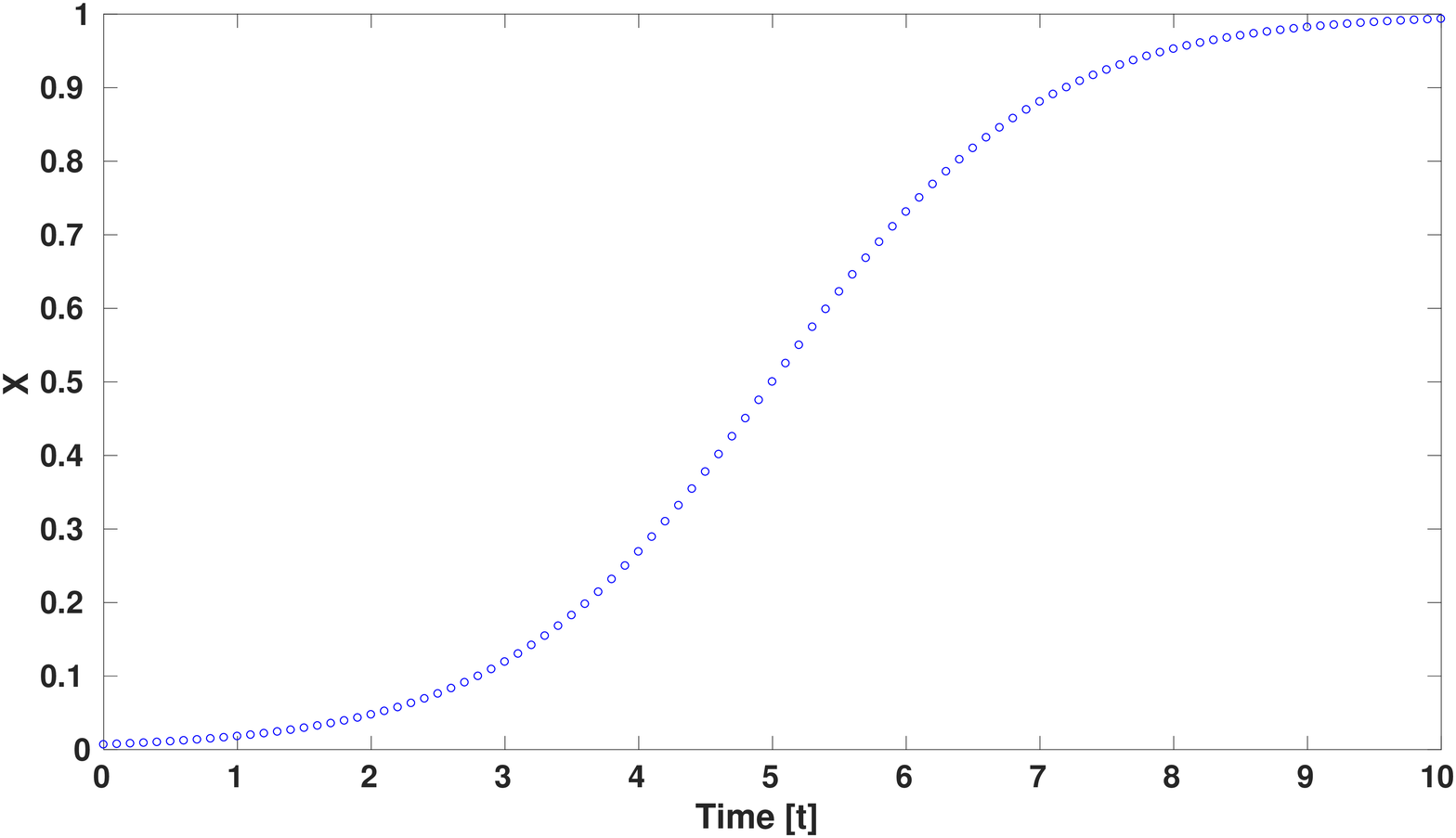}
	\includegraphics[height=10em]{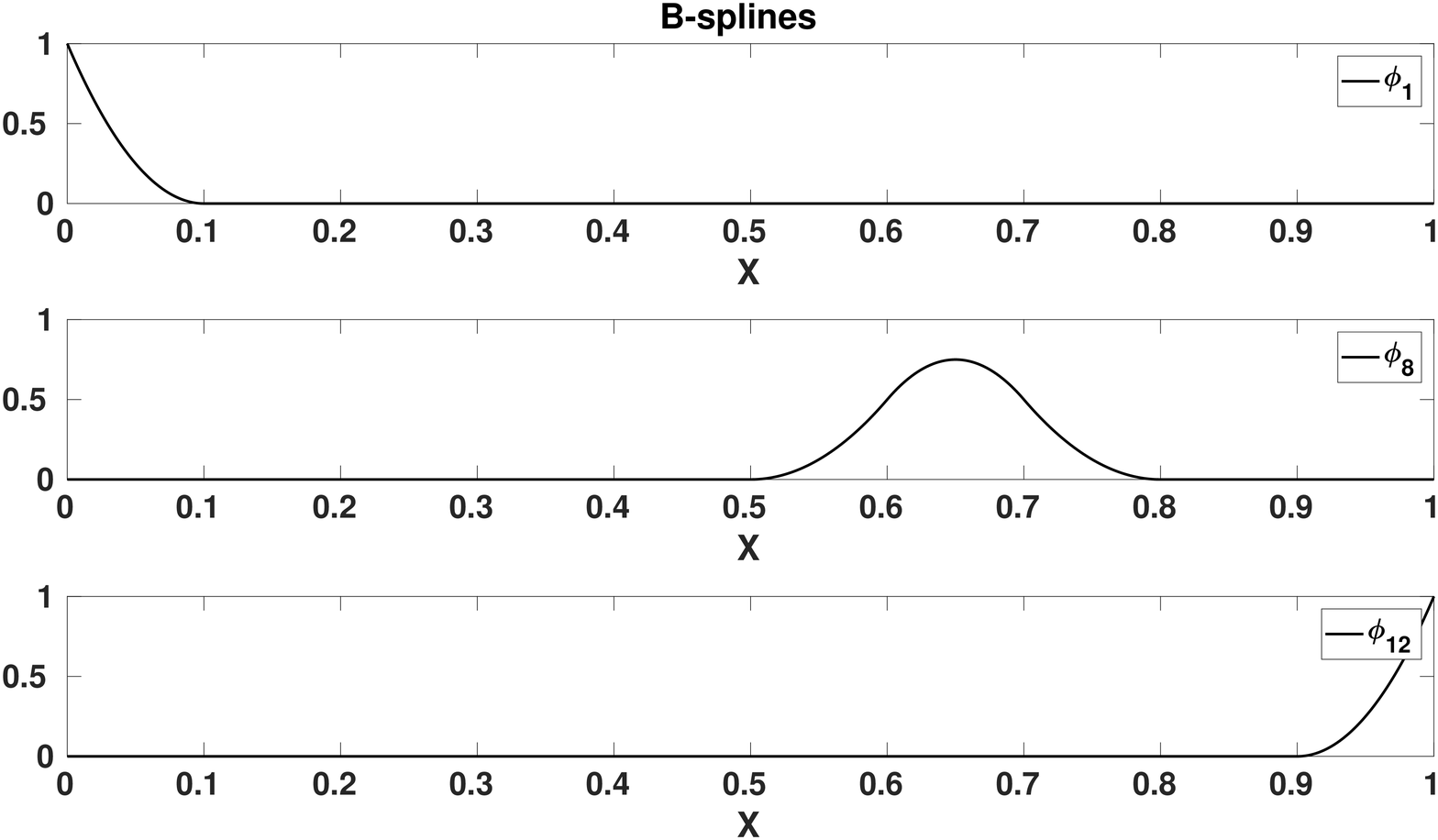}
	\includegraphics[height=10em]{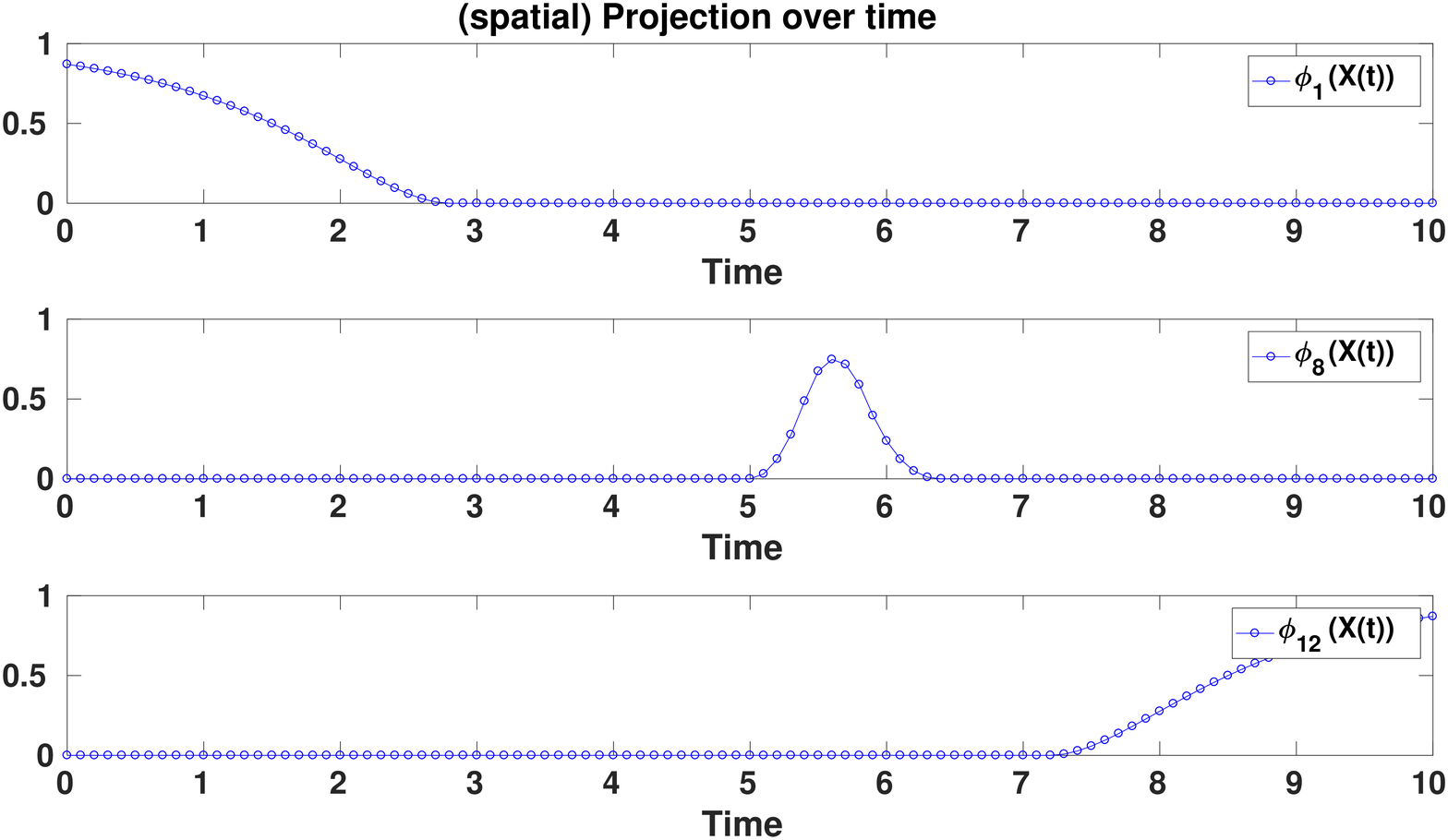}
	\includegraphics[height=10em]{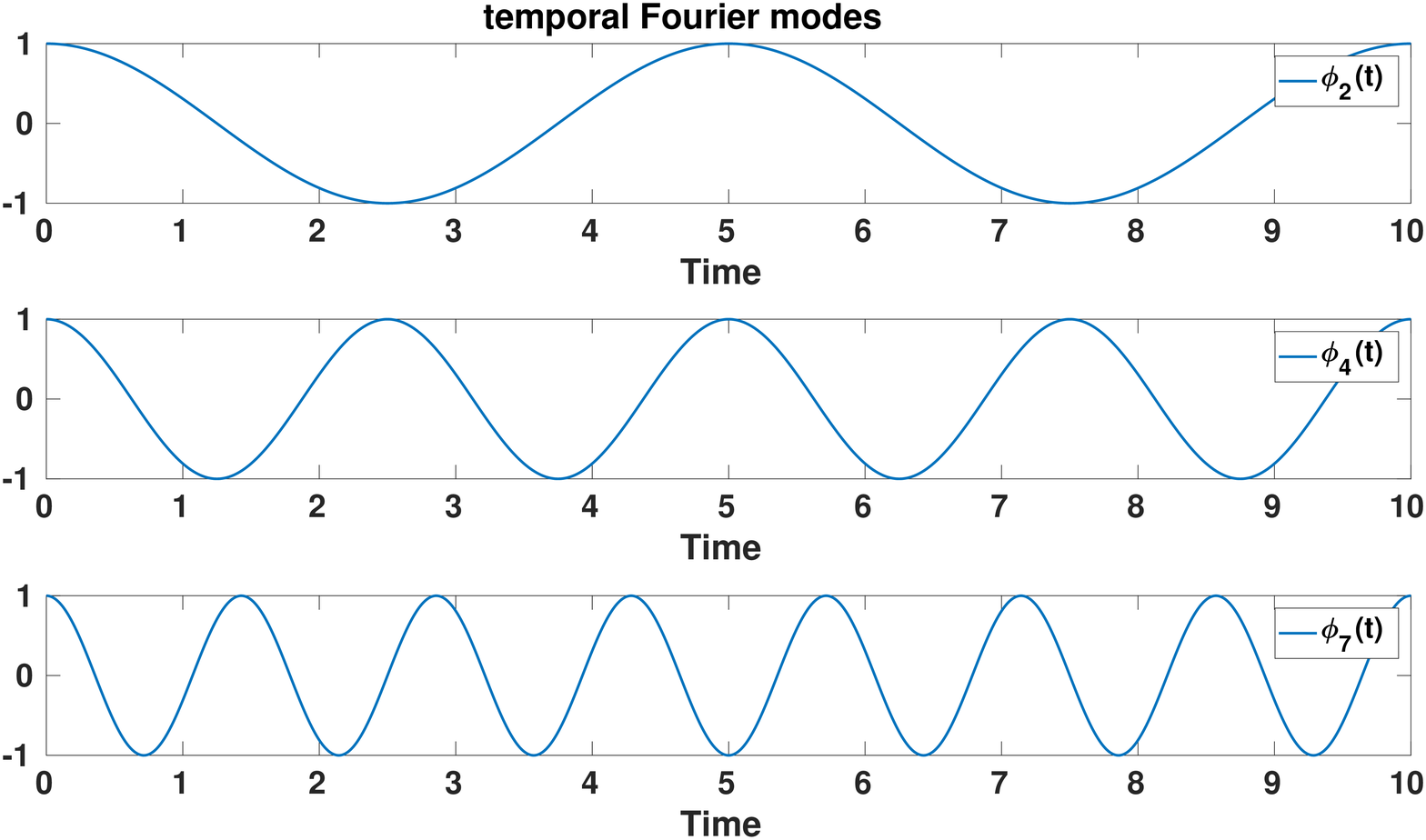}
	\includegraphics[height=10em]{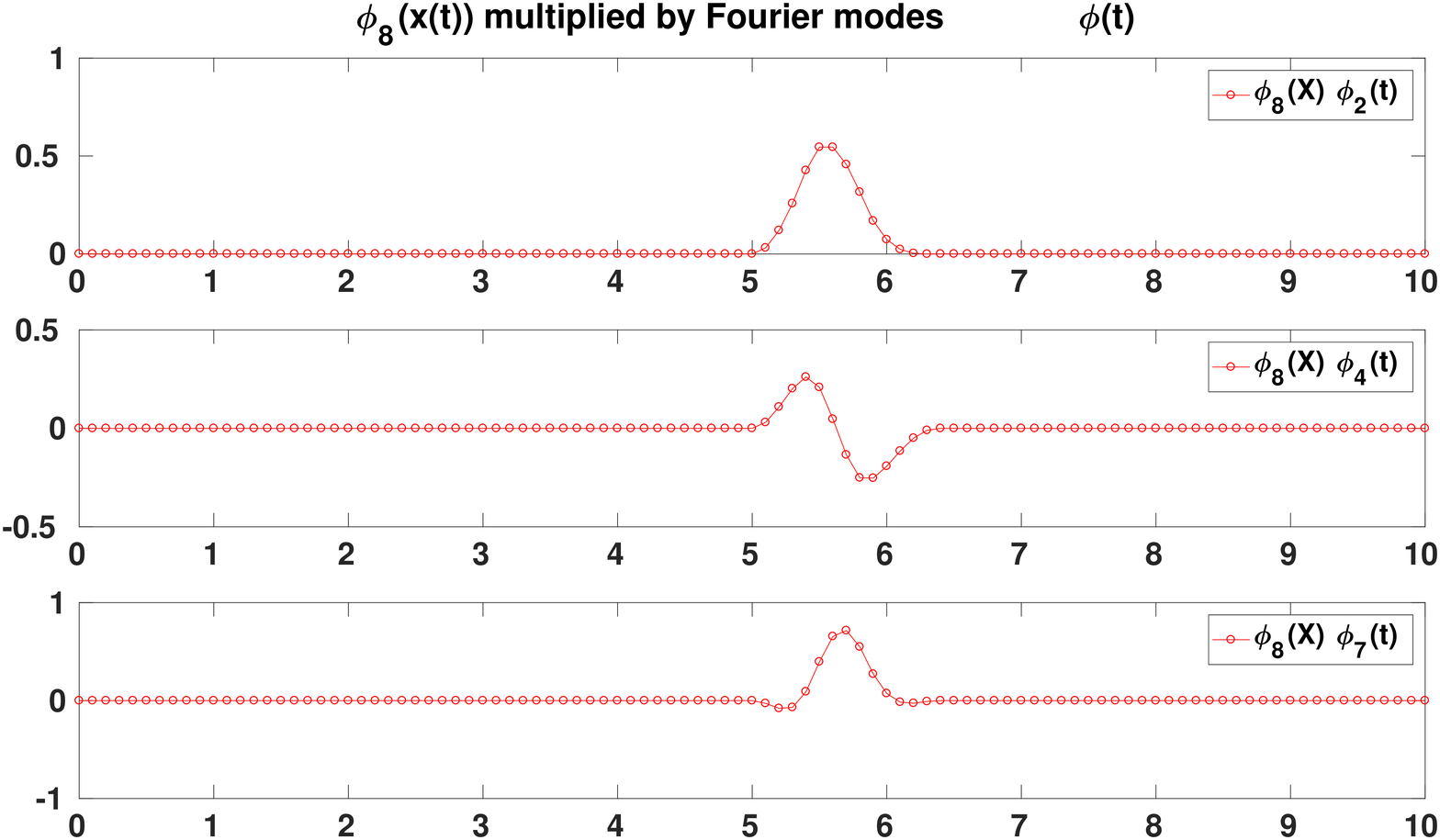}
	\caption{Graphical representation of $X$ data projection over spatial and temporal test functions. (upper left) 1-Dimensional input data, sigmoidal curve with 1 sample per time point. (upper right) 3 B-splines over the spatial domain [0:1] (middle left) projection of data over the chosen B-splines, $\bar{\phi}_1(x(t)), \bar{\phi}_8(x(t)), \bar{\phi}_{12}(x(t))$ (middle right)
	three Fourier modes $\tilde{\phi}_2(t), \tilde{\phi}_4(t), \tilde{\phi}_7(t)$. (lower left) subsequent projection in time: multiplication of (spatially projected) $\bar{\phi}_8(x(t))$ with $\tilde{\phi}_2(t), \tilde{\phi}_4(t), \tilde{\phi}_7(t)$.
	As shown inf Figure 1(c) of the main text, the resulting Fourier coefficients define a point in the weak space, if we only chose the 3 B-splines and 3 Fourier modes shown above. In a similar computation, $Z$ contains the derivatives of $X$ projected on the test functions. }
	\label{fig:projection_graphically}
\end{figure}

\Cref{fig:projection_graphically} shows in detail 3 spatial and 3 temporal projections of a simple 1-Dimensional sigmoidal trajectory.

\begin{figure}[ht]
%\centering
	\includegraphics[height=12em]{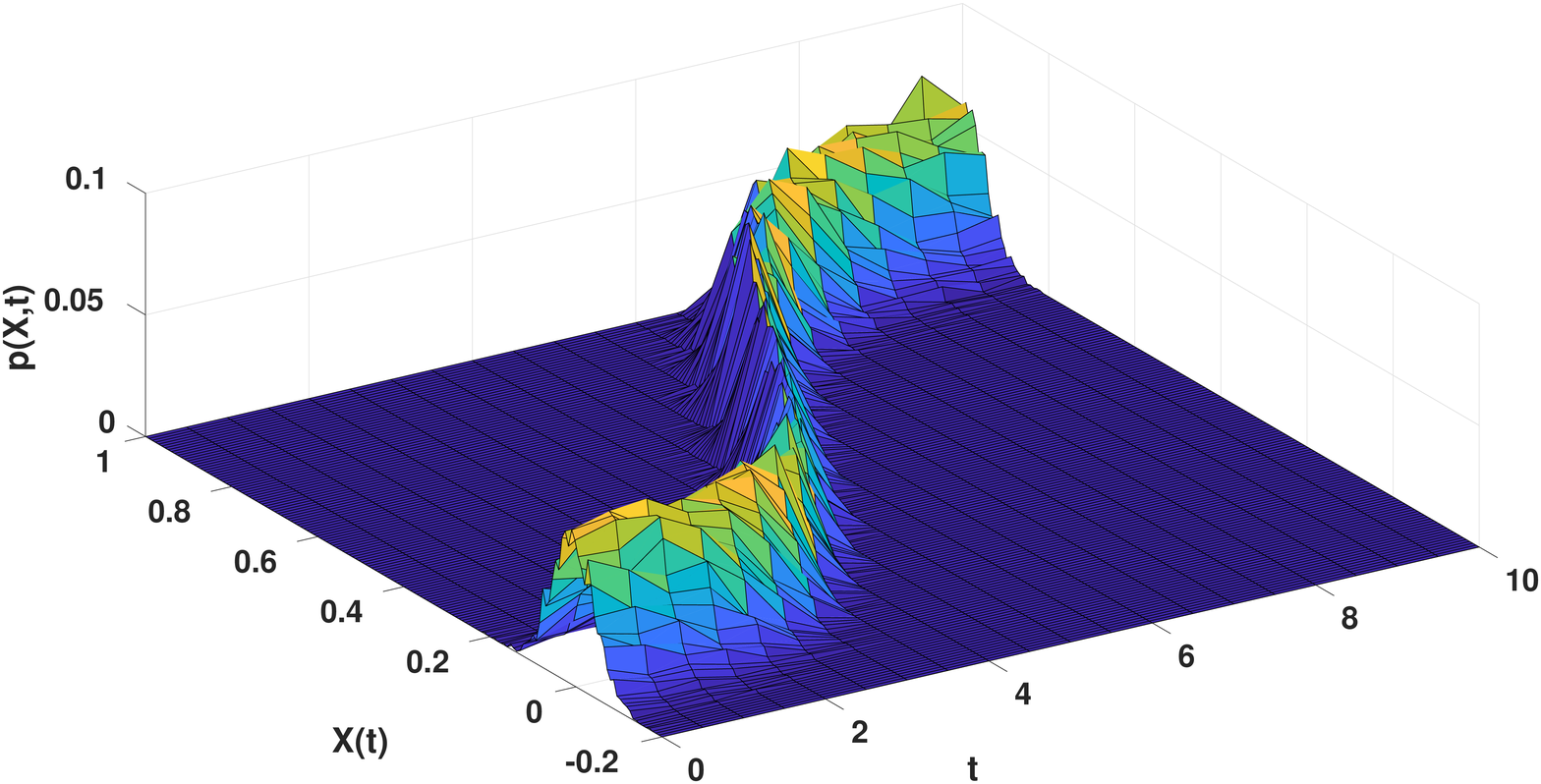}
	\includegraphics[height=12em]{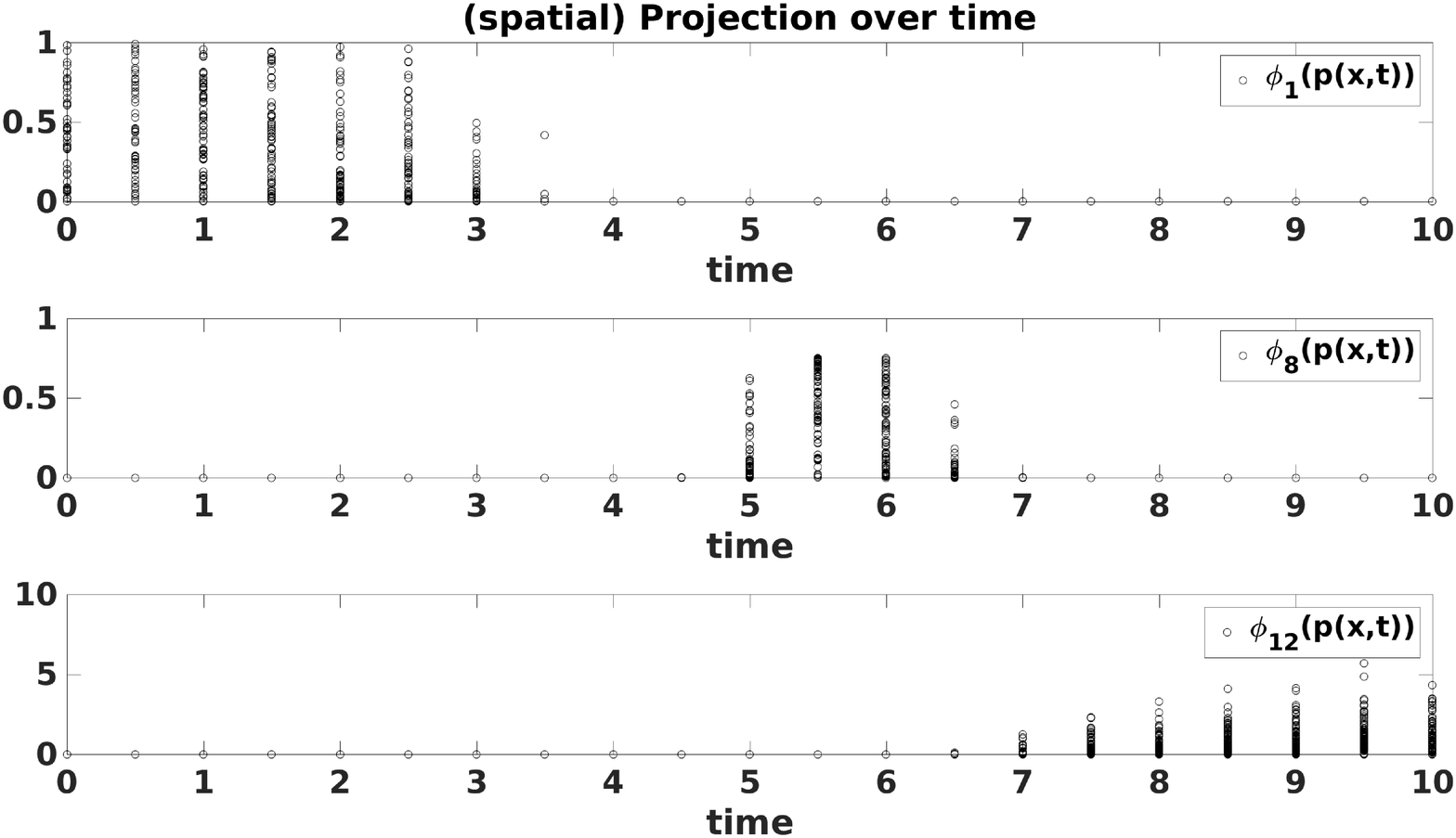}
	\includegraphics[height=12em]{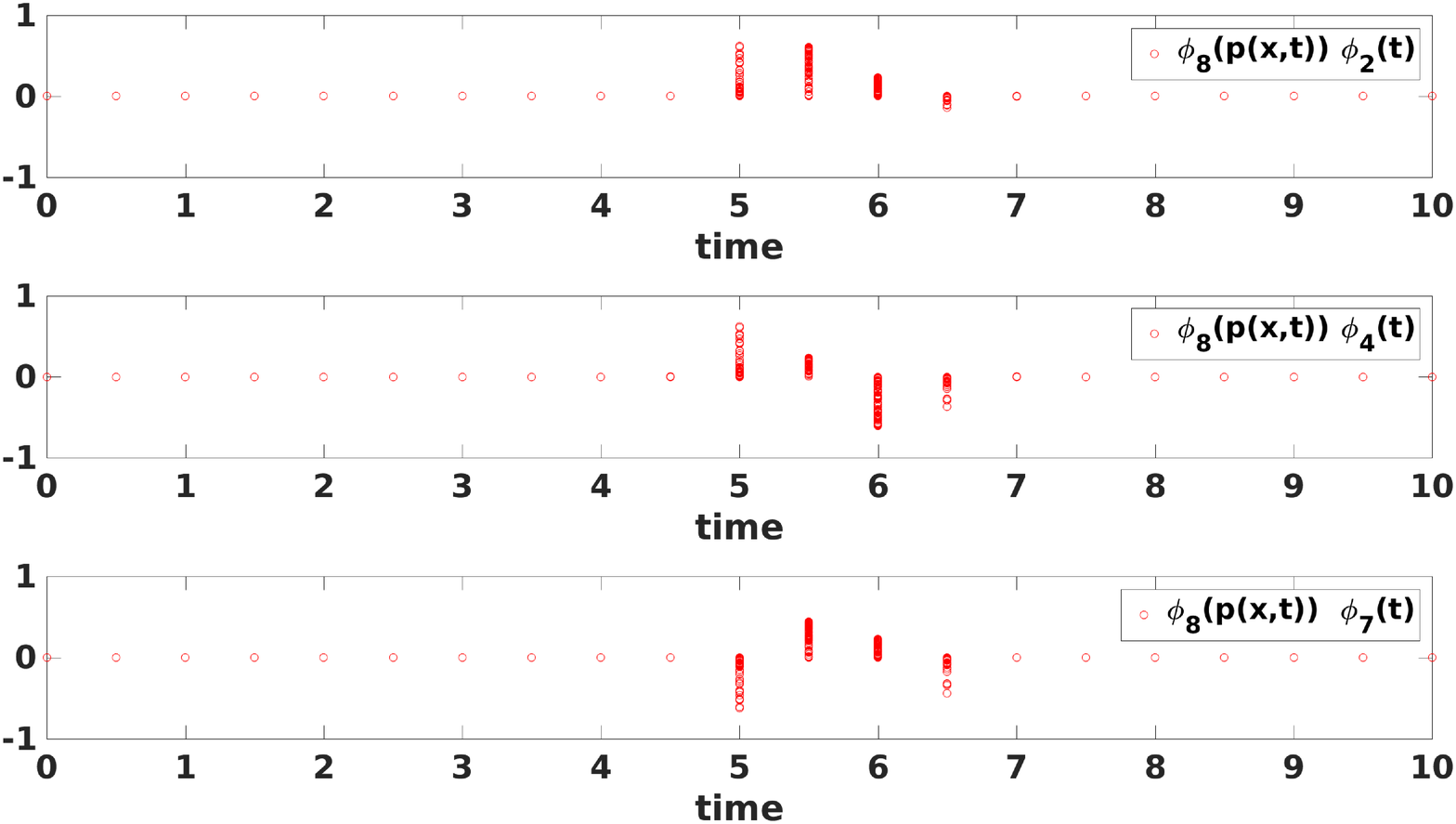}
	\caption{Graphical representation of evolving pdf $p(x,t)$, of single variable $X$ of NoS=100 samples per timepoint and subsequent projection over spatial and temporal test functions. (upper left) evolving pdf projected on 3 spatial test functions $\bar{\phi}_1(x), \bar{\phi}_8(x), \bar{\phi}_{12}(x)$. (lower left) $\bar{\phi}_8(p(x,t))$ projected on 3 Fourier modes $\tilde{\phi}_2(t), \tilde{\phi}_4(t), \tilde{\phi}_7(t) $, over time.}
	\label{fig:projection_3D}
\end{figure}

\Cref{fig:projection_3D} is the more general case of an evolving pdf of a sigmoidal, consisting of $NoS$ samples per time-point. Eventually we conclude with a linear system $Z=\Psi a$ for which we seek a sparse solution.

\begin{figure}[h]
\centering
	\includegraphics[height=14em]{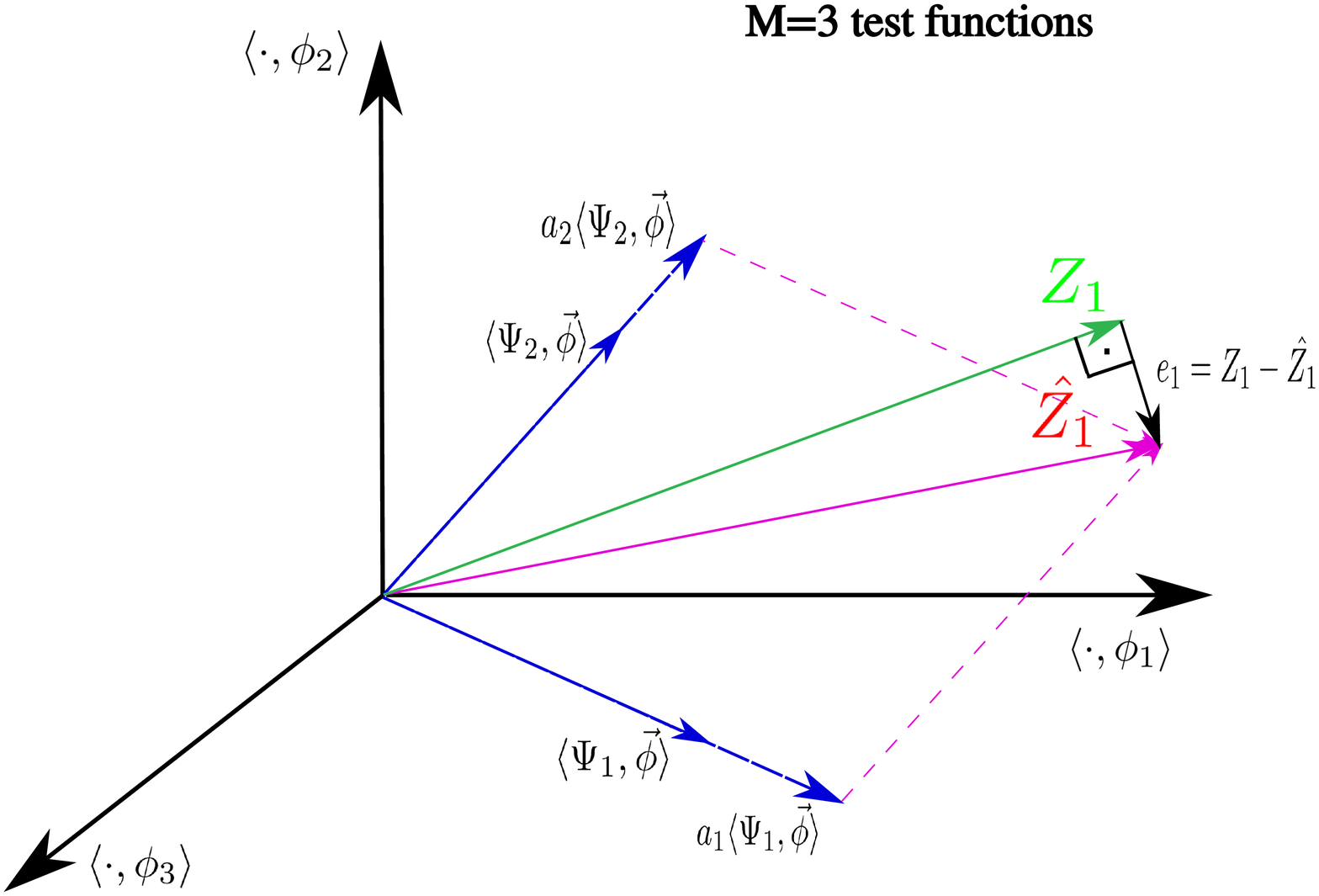}
	\caption{Geometric representation of the linear regression on $S=2$ selected (constructed) features $\langle \Psi_S, \Vec{\phi} \rangle$. In this simple example, for visualization purposes, $N=1$ variables, $M=3$ test functions: $\{\phi_1, \phi_2, \phi_3\}$, $Q=2$ dictionary atoms: $\{\psi_1, \psi_2\}$.
	$\hat{Z}_1$ is the linear combination of the constructed features projected on the weak space of $\{\phi_1, \phi_2, \phi_3\}$ or in other words, the approximation of the left-hand side of \Cref{eq:linearSLE}. 
	$e_1$ is the regression error upon selecting $S\leq Q$ features and it is being reduced when the set of correctly chosen explanatory features $S$ grows.  }
	\label{fig:geom_repr_regression}
\end{figure}

Figure \cref{fig:geom_repr_regression} is a graphical representation of
the orthogonal matching pursuit regression in the space spanned by weak space test functions $\phi$.

%---------------------------------------------------

\section{Feature selection algorithm: parameter tuning}\label{sec:App_autotuning}
Every optimization algorithm (here OMP for feature selection) is accompanied with hyper-parameters acting as "sparsification knob" which
need tuning according to the data-set. Iterative hard thresholding \cite{SINDy_2016} predefines a parameter $\lambda$ below which terms are discarded, in an iterative manner. 
Moreover, iterative soft thresholding using Cross-Validation discards one term at a time, based on $k$ splits of the data-set. In a similar mindset, we split the full data-set on training, testing and
validation disjoint subsets and then used the Bayesian Information Criterion (BIC) scoring on the test set. We use BIC in order to avoid optimistically biased trained models (i.e. $\theta$'s) \cite{Tsamard_JMLR_18}. 
The summary of the automated tuning of sparse learning algorithm parameters is shown in \cref{alg:auto_ML} for OMP.

\begin{algorithm}
\caption{Auto-tune OMP parameters}
\label{alg:auto_ML}
\hspace*{\algorithmicindent} \textbf{Input:} population data: \{data\_train, data\_test, data\_validation\},\\
\hfill discretization (range) of parameters $\theta=\{thres, iter\_trhes\}$ \\
\hspace*{\algorithmicindent} \textbf{Output:} optimal parameter $\theta^*$ along with performance plots\\
\begin{algorithmic}[1]
\For{i=1,\dots, $range\_{\theta}$}
\State $A^{\theta_i}_{train}$ = {\tt OMP}($\theta_{i};X_{train}$) \Comment{$X_{train}=$data\_train}
\State I(i) = BIC($A^{\theta_i}_{train}, X_{test}$ ) \Comment{BIC is used as information criterion $I$}
\EndFor
\State{$\theta^* =\underset{\theta}{\min}  I$} \Comment {find optimal parameter set} 
\State{set: $A\equiv A^{\theta^*}_{train}$=  {\tt OMP}($\theta^*;X_{train})$ } 
\State Compute Performance curves using $X_{valid}$ and $A$. \Comment{$X_{validation}=$data\_validation}
%\For{r=1,\dots, R}
%\State $Z^{(r)} \gets [z_1, \dots, z_N]$ \Comment{$\in M\times N$}
%\EndFor
%\Return $\hat{A}$, metrics MIP and ERC

\end{algorithmic}
\end{algorithm}

Every algorithm enforcing sparsity must be associated with
a criterion indicating if over-fitting or under-fitting of the data is
attained. For the OMP algorithm and the examples considered, we use the Bayesian Information Criterion (BIC) in order to ``score" the goodness of fit of the suggested model, associated with sparsity, on the test data.
The lowest BIC-scored model is preferred over all available (sparse) regression models, where the scoring is based on the maximization of the log likelihood function and penalization of the number of parameters $K$.

\begin{figure}[h]
\centering
    \includegraphics[height=17em]{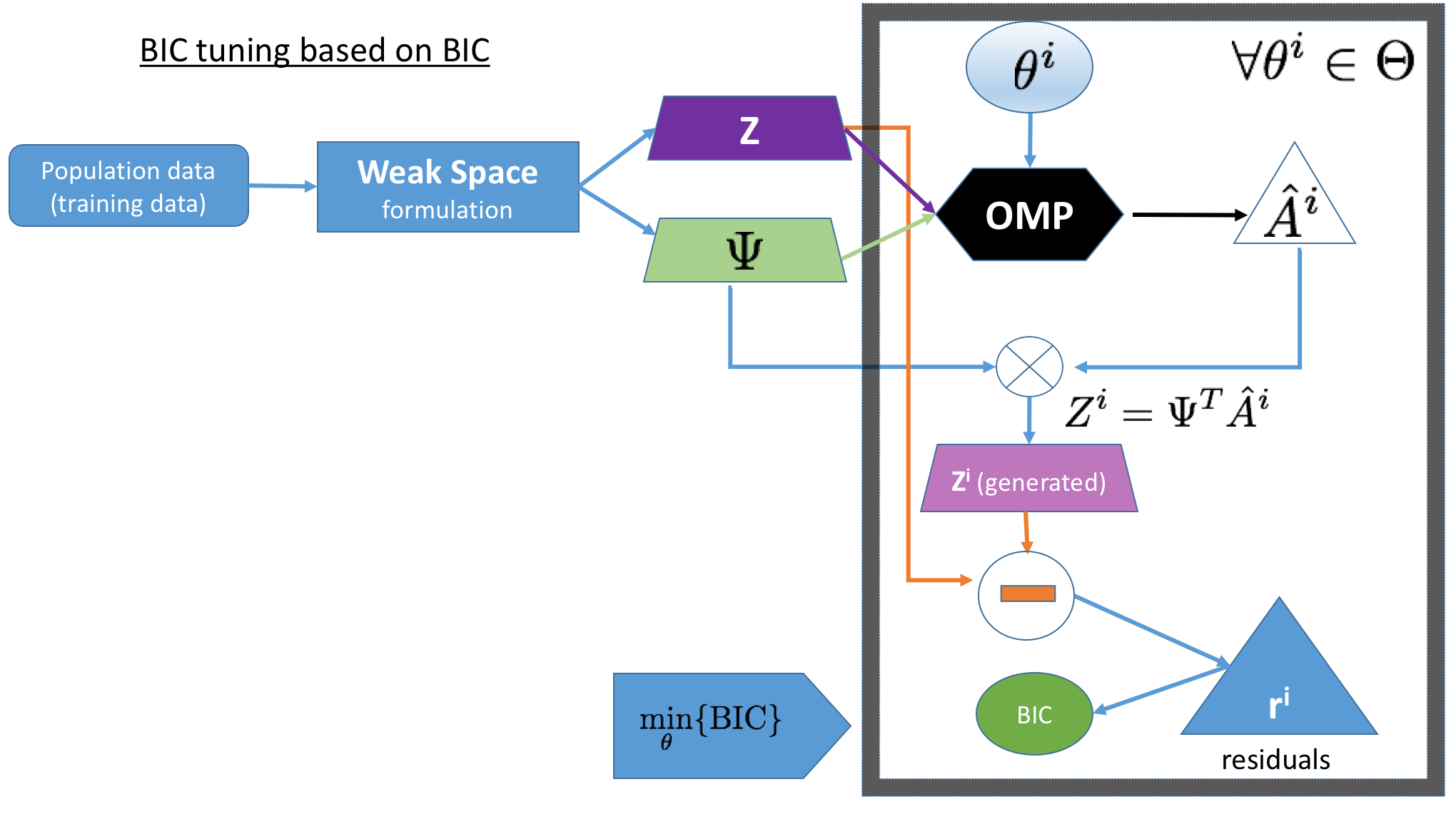}
	\caption{OMP hyper-parameter tuning according BIC, based on the test data subset.
	This is a graphical representation of \cref{alg:auto_ML}.
	Every chosen $\theta_i$ results in another connectivity matrix $\hat{A}^i$ with a different number of features, so the optimal one is chosen by the associated BIC scoring on the unseen test data set. [$Z_{test}$,$\Psi_{test}$ indices are not shown in the rectangular area...]
	Performance estimation is later computed on the validation data subset.
	}
	\label{fig:BIC_tuning}
\end{figure}

In more detail, the hyper-parameters to be tuned in OMP are the maximum selected features $K$ (dictionary atoms) and threshold $\theta \in \Theta$ ($\Theta$ being a set of scalar values of $\mathcal{O}(10^{-2})$) on the relative reconstruction error. 
Available data is partitioned in training, test and validation subsets. \Cref{fig:BIC_tuning} shows the layout of the feature selection tuning algorithm.

\bibliographystyle{siamplain}
\bibliography{references}

%% file: ex_shared.tex
\usepackage{lipsum}
\usepackage{amsfonts}
\usepackage{graphicx}
\usepackage{epstopdf}
\usepackage{algpseudocode}
\ifpdf
  \DeclareGraphicsExtensions{.eps,.pdf,.png,.jpg}
\else
  \DeclareGraphicsExtensions{.eps}
\fi

\newcommand{\blue}{\color{blue}}

\usepackage{xr}

\usepackage{enumitem}
\setlist[enumerate]{leftmargin=.5in}
\setlist[itemize]{leftmargin=.5in}

\newsiamremark{remark}{Remark}
\newsiamremark{hypothesis}{Hypothesis}
\crefname{hypothesis}{Hypothesis}{Hypotheses}
\newsiamthm{claim}{Claim}

\headers{Population Dynamics Learning}{A. Tsourtis, Y. Pantazis, and I. Tsamardinos}

\title{Inference of Stochastic Dynamical Systems from Cross-Sectional Population Data\thanks{Submitted to the editors \today. % this is footnotemark[1]!!!
\funding{The research leading to these results has received funding from the European Research Council under the European Union's Seventh Framework Programme (FP/2007‐2013)/ERC Grant Agreement no. 617393; CAUSALPATH – Next Generation Causal Analysis project.}}}

\author{Anastasios Tsourtis\footnote[4]{Computer Science Deptartment, University of Crete, Heraklion, Greece 
  (\email{tsourtis@uoc.gr}, \email{tsamard.it@gmail.com}).}
\and Yannis Pantazis\footnote[5]{Institute of Applied and Computational Mathematics, Foundation for Research and Technology - Hellas, Heraklion, Greece 
  (\email{pantazis@iacm.forth.gr}).}
\and Ioannis Tsamardinos\footnotemark[4]\,\,\footnotemark[5]
}

\usepackage{amsopn}
\DeclareMathOperator{\diag}{diag}